\RequirePackage{fix-cm}
\documentclass{article} 
\usepackage{iclr2027_conference,times}
\usepackage{microtype}

\usepackage{amsmath,amsfonts,bm}

\def\eqref#1{equation~\ref{#1}}

\def\1{\bm{1}}

\DeclareMathAlphabet{\mathsfit}{\encodingdefault}{\sfdefault}{m}{sl}
\SetMathAlphabet{\mathsfit}{bold}{\encodingdefault}{\sfdefault}{bx}{n}

\usepackage{hyperref}
\usepackage{url}
\usepackage{graphicx}
\usepackage{bbm}
\usepackage{caption}
\usepackage{placeins}
\usepackage{booktabs}
\usepackage{multirow}
\usepackage[table]{xcolor}
\usepackage{wrapfig}
\usepackage{makecell}
\usepackage{array}
\usepackage{siunitx}
\usepackage{amssymb}
\usepackage{caption}
\usepackage{enumitem}
\usepackage{subcaption}
\usepackage{xcolor}
\usepackage{adjustbox}
\usepackage{float}
\usepackage{amsmath,amssymb,booktabs}
\usepackage{algorithm}
\usepackage[noend]{algpseudocode}
\newcommand{\bestcell}[1]{\cellcolor{green!15}\textbf{#1}}
\newcommand{\secondcell}[1]{\cellcolor{green!15}\underline{#1}}

\newcommand{\costsave}[1]{%
    \textcolor{purple!80!black}{
    $\downarrow$\,#1\%~(\$)}%
}

\newcommand{\specevo}{\textsc{SpecEvo}}
\usepackage{booktabs}
\usepackage{tabularx}
\usepackage{array}
\usepackage{xcolor}
\usepackage{refcount}
\usepackage{xparse}

\definecolor{apptocblue}{RGB}{0,0,180}

\newlength{\AppTOCPageWidth}
\newcommand{\AppTOCLine}[5]{%
    \par
    \vspace{#1}%
    \begingroup
    \setlength{\parskip}{0pt}%
    \setlength{\parindent}{0pt}%
    \leftskip=#2\relax
    \rightskip=0pt
    \parfillskip=0pt
    \hangindent=#3\relax
    \hangafter=1
    \color{apptocblue}%
    #4%
    \noindent
    \makebox[#3][l]{%
        \hyperref[#5]{\ref*{#5}}%
    }%
    \hyperref[#5]{\nameref*{#5}}%
    \nobreak
    \leaders\hbox{%
        \normalfont
        \kern0.16em.\kern0.16em
    }\hfill
    \nobreak
    \makebox[\AppTOCPageWidth][r]{%
        \hyperref[#5]{\pageref*{#5}}%
    }%
    \par
    \endgroup
}

\newcommand{\AppTOCSectionLine}[1]{%
    \AppTOCLine
        {0.45em}    
        {0em}       
        {2.2em}     
        {\bfseries} 
        {#1}%
}

\newcommand{\AppTOCSubsectionLine}[1]{%
    \AppTOCLine
        {0.08em}    
        {1.7em}     
        {3.1em}     
        {\normalfont}
        {#1}%
}

\newcommand{\AppTOCSubsubsectionLine}[1]{%
    \AppTOCLine
        {0em}
        {4.0em}     
        {4.1em}
        {\normalfont}
        {#1}%
}

\ExplSyntaxOn

\cs_generate_variant:Nn \regex_count:nnN { nVN }

\cs_new_protected:Npn \specevo_appendix_toc_entry:n #1
{
    \tl_set:Nx \l_tmpa_tl { \getrefnumber{#1} }

    \regex_count:nVN { \. } \l_tmpa_tl \l_tmpa_int

    \int_case:nnF { \l_tmpa_int }
    {
        {0}{\AppTOCSectionLine{#1}}
        {1}{\AppTOCSubsectionLine{#1}}
    }
    {
        \AppTOCSubsubsectionLine{#1}
    }
}

\NewDocumentCommand{\AppendixTOCEntry}{m}
{
    \specevo_appendix_toc_entry:n {#1}
}

\ExplSyntaxOff

\NewDocumentCommand{\PrintAppendixContents}{m +m}{%
    \clearpage
    \thispagestyle{plain}
    \phantomsection

    \begin{center}
        \vspace*{-1.4em}

        {\LARGE\bfseries
        #1\par}
    \end{center}

    \vspace{0.7em}

    \begingroup
    \small
    \setlength{\parskip}{0pt}
    \setlength{\parindent}{0pt}
    \noindent
    This appendix provides supplementary material supporting the methodology,
    evaluation, and analysis presented in the main paper.
    Appendix~\ref{app:limit} discusses the limitations of
    \textsc{SpecEvo}.
    Appendices~\ref{app:related-work} and~\ref{app:baseline-details} extend the
    related-work discussion and document the baselines and common evaluation
    protocol. Appendix~\ref{app:algorithm-details} provides the complete
    algorithmic specification of \textsc{SpecEvo}, including the Speculator
    loop and the Navigator and Advisor interventions, while
    Appendix~\ref{app:metric-details} defines the metrics used to analyze
    search dynamics. Appendices~\ref{app:benchmark-details}
    and~\ref{app:add_result} describe the benchmark suites and provide detailed
    experimental results, including API cost and token usage breakdowns.
    Appendix~\ref{app:prompt-templates} presents the complete prompt templates
    for all three roles, and Appendix~\ref{app:qualitative} analyzes generated
    programs, recurring failure modes, algorithmic mechanisms, and the guidance
    consolidated by the Advisor on Circle Packing Rectangle and EPLB.
    \endgroup

    \vspace{1.0em}

    {\Large\scshape Contents\par}

    \vspace{0.35em}

    \begingroup
    \setlength{\parskip}{0pt}
    \setlength{\parindent}{0pt}
    \normalsize
    #2
    \endgroup

    \clearpage
}

\usepackage{listings}
\usepackage{xcolor}

\definecolor{codebg}{RGB}{248,249,251}
\definecolor{codeframe}{RGB}{205,211,220}
\definecolor{codekeyword}{RGB}{31,78,121}
\definecolor{codestring}{RGB}{163,21,21}
\definecolor{codecomment}{RGB}{72,118,72}
\definecolor{codenumber}{RGB}{120,120,120}

\lstdefinestyle{specevoPython}{
    language=Python,
    basicstyle=\ttfamily\scriptsize,
    keywordstyle=\color{codekeyword}\bfseries,
    stringstyle=\color{codestring},
    commentstyle=\color{codecomment}\itshape,
    numbers=left,
    numberstyle=\tiny\color{codenumber},
    numbersep=9pt,
    stepnumber=1,
    backgroundcolor=\color{codebg},
    frame=single,
    rulecolor=\color{codeframe},
    framerule=0.5pt,
    framesep=6pt,
    xleftmargin=1.8em,
    framexleftmargin=1.4em,
    xrightmargin=0.3em,
    breaklines=true,
    breakatwhitespace=false,
    postbreak=\mbox{\textcolor{codenumber}{$\hookrightarrow$}\space},
    showstringspaces=false,
    keepspaces=true,
    columns=fullflexible,
    tabsize=4,
    captionpos=t,
    aboveskip=0.8em,
    belowskip=0.8em
}

\usepackage{xcolor}
\usepackage{caption}
\usepackage{minted}
\usepackage[most]{tcolorbox}

\tcbuselibrary{minted,breakable,skins}

\definecolor{codebg}{HTML}{F4FBFD}
\definecolor{codeframe}{HTML}{4E7D87}
\definecolor{codelinenum}{HTML}{9AA3A6}

\newsavebox{\LSRTableBox}
\newsavebox{\AblationTableBox}
\newlength{\AblationRowGap}
\newlength{\AblationLastGap}

\usepackage{xcolor, soul,todonotes} 
\definecolor{kmycolor}{rgb}{0.858, 0.188, 0.478}

\definecolor{tycolor}{rgb}{0.0, 0.45, 0.75}

\title{\raggedright SpecEvo: Speculative Evolution with Large Language Models for Cost-Efficient Scientific Discovery}

\author{
\hfill
Ha Minh Hieu\textsuperscript{1,*} \quad
Tongyao Zhu\textsuperscript{2,\textdagger} \quad
Do Xuan Long\textsuperscript{2,\textdagger} \quad
Min-Yen Kan\textsuperscript{2,\textdagger}
\hfill
\\
\hfill
\textsuperscript{1}Hanoi University of Science and Technology, 
\textsuperscript{2}National University of Singapore
\hfill
\\
\hfill
\texttt{haminhhieu1005@gmail.com} \quad
\texttt{\{tongyao.zhu,xuanlong.do\}@u.nus.edu}
\hfill
\\
\hfill
\texttt{knmnyn@nus.edu.sg}
\hfill
}

\iclrfinalcopy 
\begin{document}
\raggedbottom

\maketitle
\begingroup
\renewcommand{\thefootnote}{\fnsymbol{footnote}}
\footnotetext[1]{Work done during the internship at WING, NUS.}
\footnotetext[2]{Corresponding author.}
\endgroup

\begin{abstract}
LLM-driven evolutionary search is a promising paradigm for scientific discovery, but existing methods remain costly and slow, often applying expensive reasoning uniformly while underusing evidence from failed, redundant, and saturated trials. We introduce \emph{Speculative Evolution} (\textsc{SpecEvo}), a multi-LLM framework that coordinates lightweight and frontier models through continual consolidation of search evidence. Lightweight models power both parallel Speculators, which generate and refine candidate programs, and an Advisor, which distills accumulated successes, failures, and saturated directions into persistent guidance for subsequent exploration. At sparse checkpoints, a frontier-model Navigator inspects the evolving population and intervenes through synthesis, surgical refinement, or reframing according to search stagnation. This \emph{speculate-then-consolidate} workflow delegates routine exploration and evidence consolidation to inexpensive models while reserving frontier reasoning for strategic interventions whose benefits extend across many subsequent low-cost evaluations. Across 176 tasks spanning four scientific and algorithmic discovery domains, \textsc{SpecEvo} matches or outperforms frontier-model baselines while reducing API cost by up to 80\%, achieves up to \(1.5\times\) greater evaluation efficiency than cost-efficient baselines on equation discovery, and improves over the strongest CO-Bench baseline by 3.8\%. These results support trajectory-conditioned model coordination as an effective approach to cost-efficient scientific discovery. \footnote[3]{Our code is available at
\href{https://github.com/langkhachhoha/SpecEvo}{\texttt{https://github.com/langkhachhoha/SpecEvo}}}
\end{abstract}

\section{Introduction}
Recent advances in large language models have enabled a new paradigm of evolutionary search for automated scientific discovery \citep{yamada2025ai,liu2026evox}. LLMs act as semantic variation operators, proposing meaningful modifications to candidate programs that are refined through execution-based evaluation and selection \citep{liu2026systematic, wu2024evolutionary}. Such approaches have achieved strong results across mathematical discovery, combinatorial optimization, and systems engineering \citep{hieu2026pareto, cemri2026adaevolve, novikov2025alphaevolve,liu2024evolution}.

\begin{wrapfigure}{r}{0.53\textwidth}
    \centering
    \includegraphics[width=\linewidth]{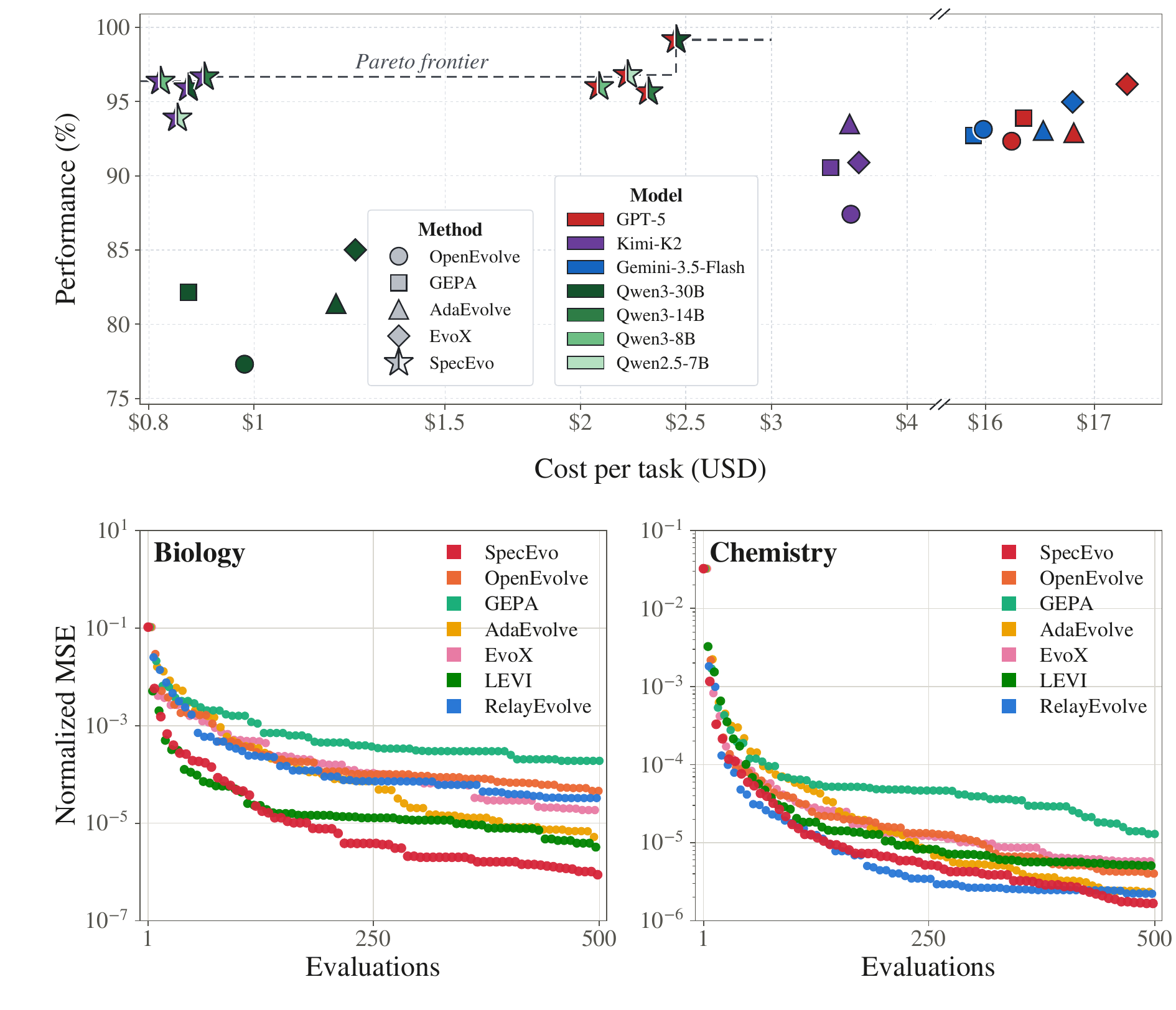}
    \caption{\textbf{Top:} Performance vs.\ API cost on mathematical discovery tasks; \specevo{} forms the Pareto frontier. \textbf{Bottom:} Best-so-far NMSE on LSR-Synth Biology and Chemistry over 500 evaluations.}
    \label{fig:trade-off-per-cost}
\end{wrapfigure}

Despite their strong results, many systems nevertheless use a single backbone for candidate generation and higher-level search decisions \citep{agrawal2025gepa,cemri2026adaevolve,liu2026evox}. When that backbone is a frontier model, routine proposals repeatedly incur expensive inference. Figure~\ref{fig:trade-off-per-cost} illustrates the resulting performance--cost trade-off: using lightweight backbones lowers API expenditure but can substantially reduce solution quality in the evaluated baselines. Lightweight models offer inexpensive exploration, but without reusable guidance they can repeatedly revisit similar strategies or reproduce implementation errors \citep{dat2025hsevo,liu2024exploring}. Merely generating more affordable proposals therefore does not ensure sustained search progress.
Heterogeneous-model approaches address part of this problem by assigning different search operations to different models or transferring a population from lightweight exploration to frontier-model refinement \citep{tanveer2026levi,luo2026relay,10.1145/3770855.3818067}. These approaches motivate a complementary question: \emph{How can accumulated successes, failures, and signs of saturation continually inform both subsequent lightweight proposals and the scope of frontier-model intervention?}

To address these gaps, we present \textsc{SpecEvo}, a multi-LLM evolutionary framework organized around a workflow we call \emph{speculate-then-consolidate} (StC). StC repeatedly converts accumulated search evidence into guidance and strategic interventions that inform subsequent low-cost exploration; \textsc{SpecEvo} provides its concrete realization through specialized model roles and shared search state. Consolidation occurs periodically while lightweight exploration continues. A central design choice is to distinguish the programs retained for further search from the evidence retained for learning. An adaptive MAP-Elites-style \citep{mouret2015illuminating} archive maintains one elite per structural--semantic niche, allowing programs outside the global top ranks to remain available when they improve their own niches. A separate history records all evaluated attempts, including rejected programs and execution failures. Together, these structures expose which strategies remain competitive, which directions consume evaluations without progress, and which implementation errors recur. This evidence supports both revisable guidance for lightweight models and search-state-dependent choices about frontier intervention.
Across model configurations, \textsc{SpecEvo} improves the performance--cost trade-off over methods built around a single frontier model. Its advantages over cost-efficient allocation strategies persist under matched model-call and evaluation budgets, supporting the value of coordinating model use through accumulated search evidence. Ablations and search-trajectory analyses further support the importance of preserving diverse candidates, adapting intervention scope to stagnation, and revising guidance as evidence accumulates. Our main contributions are:

\begin{itemize}
    \item We introduce \textsc{SpecEvo}, an asynchronous realization of StC with parallel lightweight Speculators, a lightweight Advisor, and a sparsely invoked frontier-model Navigator.

    \item We develop coordinated feedback mechanisms that combine failure- and niche-progress-aware guidance with a stagnation signal controlling parent sampling and selecting among synthesis, surgical refinement, and reframing at fixed Navigator checkpoints.

    \item We provide a systematic evaluation on 176 tasks spanning mathematical discovery, systems optimization, combinatorial optimization, and scientific equation discovery, combining comparisons under matched budgets with controlled model-allocation studies, component ablations, and analyses of search dynamics.
\end{itemize}

\FloatBarrier

\section{Related Work}
\label{related-work}

\textbf{LLM-Guided Evolutionary Search.}
FunSearch and AlphaEvolve demonstrate how execution-based selection can turn LLM-generated program variations into mathematical and systems discoveries \citep{romera2024mathematical,novikov2025alphaevolve}. Subsequent work addresses two complementary challenges within this loop. The first is preserving useful alternatives: OpenEvolve combines island evolution with quality-diversity archiving, while SeaEvo organizes search around persistent strategy-level memory \citep{openevolve,luo2026seaevo}. These designs retain alternatives that can support exploration beyond the current best program. The second is using evaluation feedback to improve search decisions: GEPA derives reflective updates and maintains complementary candidates through Pareto selection, AdaEvolve adapts exploration and resource allocation to observed progress, and EvoX evolves the search policy itself \citep{agrawal2025gepa,cemri2026adaevolve,liu2026evox}. \textsc{SpecEvo} builds on these principles, using archive- and trajectory-level evidence to coordinate lightweight exploration, guidance revision, and frontier intervention.

\textbf{Cost-Efficient LLM Allocation.}
Cost-efficient LLM systems commonly coordinate cheap and strong models through cascades or adaptive routing~\citep{chen2023frugalgpt,aggarwal2024automix,ding2024hybrid,ong2025routellm, zhang2026budget}. SCOPE uses query-level partial evaluations to select a fixed model assignment for each module of a compound system under a quality constraint~\citep{10.1145/3770855.3818067}. While reducing configuration-search cost, it does not adapt allocation to an evolving search state. Within evolutionary search, LEVI combines diversity preservation and proxy evaluation with role-aware routing, assigning local refinements to cheap models and paradigm shifts to a frontier model~\citep{tanveer2026levi}. However, the frontier model remains tied to a predefined role, while its error-oriented advice does not use broader progress and saturation evidence to adapt interventions. RelayEvolve develops multiple cheap-model trajectories and performs a one-time handoff of a quality-diverse population to strong-model refinement~\citep{luo2026relay}. Afterward, lightweight exploration ends, and evidence from discarded or exhausted trajectories is not consolidated into reusable guidance. \textsc{SpecEvo} instead continually consolidates accumulated evidence to guide both lightweight exploration and state-dependent frontier-model intervention.

\section{Methodology}
\label{method}

Figure~\ref{fig:overall} summarizes how \textsc{SpecEvo} initializes its population and coordinates the subsequent search, while Appendices~\ref{app:algorithm-details} and~\ref{app:prompt-templates} provide the complete algorithms and prompt templates, respectively.

\subsection{Motivation: From Speculative Decoding to Speculative Evolution} 
\label{sec:lab_motivation} 

Speculative decoding uses lightweight drafts and parallel verification by a stronger model to amortize inference while preserving the target distribution \citep{leviathan2023fast,chen2023accelerating}. Related evolutionary methods likewise reserve stronger models for difficult mutations or structural changes \citep{ray2026adaptevolve,tanveer2026levi}. In \textsc{SpecEvo}, candidate quality is established by an executable evaluator, while accumulated outcomes inform future proposals. The \emph{speculate-then-consolidate} (StC) workflow therefore transfers the principle of asymmetric computation to search trajectories: inexpensive exploration supplies evidence, and periodic consolidation turns that evidence into guidance and strategic proposals for subsequent exploration.

\begin{figure}[H]
    \centering
    \includegraphics[width=0.85\linewidth]{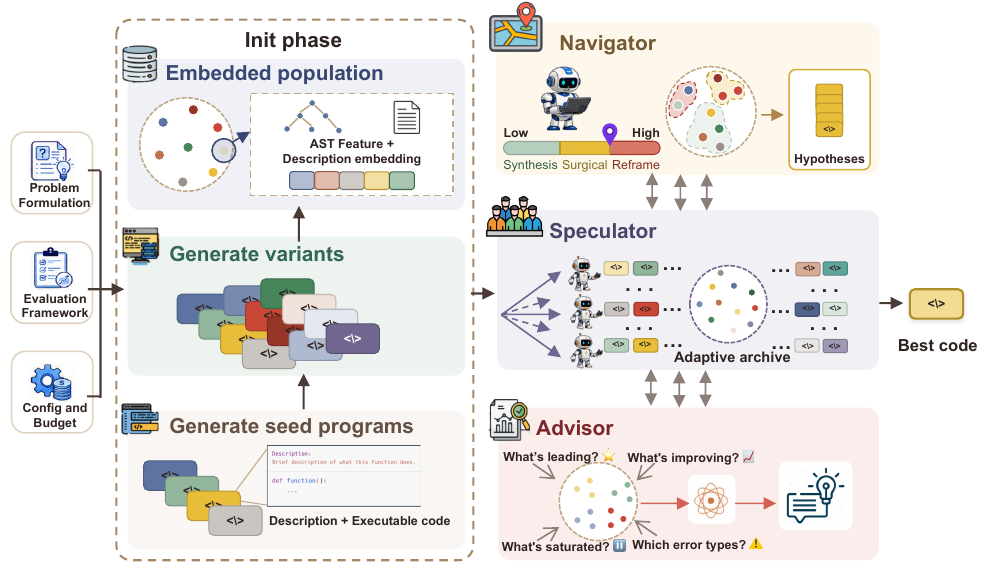}
\caption{\small
Overview of \textsc{SpecEvo}. Initialization (left) generates $M$ seed programs, expands each executable seed into $V$ lightweight variants, and embeds the evaluated population to initialize the archive. During evolution (right), $W$ lightweight-model Speculators continuously explore the archive. The frontier-model Navigator (top) intervenes every $\Delta_N$ completed evaluations, while the lightweight-model Advisor (bottom) updates reusable guidance every $\Delta_A$ evaluations. 
}
    \label{fig:overall}
\end{figure}

Figure~\ref{fig:overall} separates population initialization on the left from the ongoing StC loop on the right. Initialization proceeds from seed generation through lightweight diversification to population embedding. The resulting archive supports continuous Speculator exploration, while the Navigator above and the Advisor below provide strategic proposals and reusable guidance at slower cadences. Once the prescribed budget is exhausted, the search terminates and the highest-scoring program in the final archive is returned. We next describe the shared search state and each functional role.

\paragraph{Problem setup.}
Given a task $\mathcal{T}$ with a space $\mathcal{P}$ of executable candidate programs and an evaluator $F:\mathcal{P}\to\mathbb{R}$, we aim to discover $p^\star=\arg\max_{p\in\mathcal{P}}F(p)$ under a prescribed wall-clock, API-cost, or evaluation budget. At evaluation step $t$, we maintain two shared structures: a behavioral archive $\mathcal{A}_t$ retaining a bounded set of competitive programs, and a search history $\mathcal{H}_t$ recording every evaluated attempt, including rejected candidates and execution failures. Each stored program $p$ is paired with a natural-language description $d(p)$ of its strategy, supplied by the generating model.

\paragraph{Initialization.}
The \textbf{Generate seed programs} block in Figure~\ref{fig:overall} provides diverse, verified starting hypotheses. The Navigator uses the frontier model $\mathcal{M}_{\mathrm{N}}$ to generate $M{=}5$ seed programs sequentially, conditioning each call on the strategies and evaluation outcomes of earlier seeds. The prompt asks the model to identify their algorithmic paradigms and propose a different modeling, decomposition, construction, or search strategy. In the subsequent \textbf{Generate variants} block, the Speculator pool uses the lightweight model $\mathcal{M}_{\mathrm{S}}$ to expand each executable seed into $V{=}20$ inexpensive variants in parallel, drawing on $n_{\mathrm{insp}}\in\{1,2\}$ other seeds as inspiration. We initialize the behavioral archive from these evaluated programs using the representation and admission rule described next.

\paragraph{Behavioral archive.}
Fitness alone does not distinguish the strategies implemented by candidate programs. A population selected solely by global fitness can become dominated by variants of one strategy, discarding lower-scoring alternatives that may provide complementary mechanisms or promising starting points for further search. We therefore maintain $\mathcal{A}_t$ as a MAP-Elites-style archive \citep{mouret2015illuminating} holding at most $K{=}50$ incumbents, one per \emph{niche}: a region of structural--semantic descriptor space within which candidates compete on fitness. Competition within niches preserves alternatives across descriptor regions, keeping them available for Speculator mutation and crossover and Navigator synthesis even when they fall outside the global top ranks.

The \textbf{Embedded population} block constructs the descriptors used by the \textbf{Adaptive archive}. Let $r=1,2,\ldots$ index successive descriptor-fitting rounds. At each round, we estimate the PCA basis and standardization parameters from the current archive and hold them fixed until the next refit. Under the descriptor map fitted at round $r$, each program is represented as
\begin{equation}
\mathbf{z}_r(p)=
\big[
\mathcal{Z}^{\mathrm{ast}}_r\!\big(\phi(p)\big)
\;\big\|\;
\mathcal{Z}^{\mathrm{sem}}_r\!\big(
\mathbf{U}_r^{\top}
(\mathbf{e}(p)-\bar{\mathbf{e}}_r)
\big)
\big]
\in \mathbb{R}^{16},
\label{eq:descriptor}
\end{equation}
where $\|$ denotes concatenation.
The structural vector $\phi(p)\in\mathbb{R}^{8}$ contains eight statistics extracted from the code's abstract syntax tree (AST), each transformed by $\log(1+x)$ to reduce sensitivity to unusually large programs.
The semantic block captures the strategy description $d(p)$ through its embedding $\mathbf{e}(p)$. We subtract the archive embedding mean $\bar{\mathbf{e}}_r$ and project onto the eight leading PCA directions stored in $\mathbf{U}_r$, yielding eight semantic features. The operators $\mathcal{Z}^{\mathrm{ast}}_r$ and $\mathcal{Z}^{\mathrm{sem}}_r$ standardize each feature separately by subtracting its archive mean and dividing by its archive standard deviation, both estimated at fitting round $r$; features with zero variance are mapped to zero. This makes the structural and semantic features comparable in scale for the Euclidean distances used in niche assignment (Appendix~\ref{app:behavioral-descriptor}).

\subsection{The Speculate-then-Consolidate Workflow}
\label{sec:workflow}

While fewer than $K$ slots are occupied, valid candidates are admitted into new slots. At capacity, the first descriptor fit defines sites $\mathbf{c}_i=\mathbf{z}_r(p_i)$ whose Voronoi cells form the niches. A candidate is then assigned to its nearest site and admitted only if it strictly improves that niche's incumbent. Replacement leaves the site and descriptor map fixed until the next refit, stabilizing candidate assignment. After initialization, every $\Delta_C{=}50$ completed evaluations, we refit PCA and normalization statistics on the current archive, recompute incumbent descriptors, and re-anchor the sites to follow the evolving population. Slot indices support progress tracking, although niche boundaries can change (Appendix~\ref{app:algorithm-details}). Table~\ref{ablation} shows the effect of refitting on overall search performance.


\paragraph{Stagnation signal.}
The low-to-high scale in the Navigator block represents search stagnation: $s_t$ increases toward $1$ when either global-best improvements or archive admissions become infrequent. We use this signal to adjust both parent selection and the scope of Navigator intervention:
\begin{equation}
s_t=
\max\!\Big\{
\min\!\big(1,\tfrac{t-\tau^{\mathrm{best}}_t}{H_g}\big),
\;
\min\!\big(1,\tfrac{t-\tau^{\mathrm{adm}}_t}{H_l}\big)
\Big\},
\label{eq:stagnation}
\end{equation}
where $\tau^{\mathrm{best}}_t$ and $\tau^{\mathrm{adm}}_t$ denote the evaluation steps of the most recent global-best improvement and archive admission, respectively, with normalization horizons $H_g{=}100$ and $H_l{=}20$. Each term measures a normalized waiting time, capped at one. Taking their maximum makes $s_t=1$ after 100 evaluations without a global-best improvement or 20 without an archive admission.


\paragraph{Speculators: continuous exploration.}
We use $W$ asynchronous Speculators ($W{=}4$ in our experiments), which continuously generate candidates using $\mathcal{M}_{\mathrm{S}}$. Sampling concentrates on leading elites when the search is progressing, while increasing stagnation broadens exploration toward lower-ranked archive regions. Whenever a worker becomes idle, it selects a parent according to its fitness rank $r_i$ under a Zipf distribution whose exponent adapts to stagnation \citep{piantadosi2014zipf}:
\begin{equation}
\Pr(p_i\mid\mathcal{A}_t)=
\frac{r_i^{-\beta_t}}
{\sum_{j=1}^{|\mathcal{A}_t|}r_j^{-\beta_t}},
\qquad
\beta_t=
\beta_{\min}+(1-s_t)(\beta_{\max}-\beta_{\min}),
\label{eq:zipf}
\end{equation}
with $(\beta_{\min},\beta_{\max})=(0.3,2.0)$. Thus, $\beta_t$ decreases from $2.0$ to $0.3$ as stagnation increases, flattening the distribution while retaining a preference for higher ranks. We use rank-based selection to avoid dependence on task-specific score scales. The worker then inserts the selected parent into one of six uniformly sampled variation prompts: general mutation, focused single-weakness repair, mechanism replacement, analysis-targeted mutation, structural crossover, and component-level crossover. Figure~\ref{fig:cdi} and Table~\ref{ablation} show that using these complementary prompts sustains higher code diversity and yields better performance than a single general mutation prompt. For crossover, the worker draws the second parent from a different niche whenever possible. Every proposal is executed, checked against the archive admission rule, and recorded in $\mathcal{H}_t$ whether or not it is admitted. 

\paragraph{Navigator: sparse, state-dependent intervention.}
At each Navigator checkpoint, the thresholds $\gamma_1$ and $\gamma_2$ partition the stagnation signal into three levels of increasing intervention scope:
\begin{equation}
m_t=
\begin{cases}
\textsc{Synthesis}, & s_t\le\gamma_1,\\[1pt]
\textsc{Surgical}, & \gamma_1<s_t\le\gamma_2,\\[1pt]
\textsc{Reframe}, & s_t>\gamma_2,
\end{cases}
\qquad (\gamma_1,\gamma_2)=(0.4,0.7).
\label{eq:modes}
\end{equation}
When progress remains steady, the Navigator uses \textsc{Synthesis} to combine complementary mechanisms from competitive niches into a coherent program. Under moderate stagnation, it uses \textsc{Surgical} for a targeted structural correction to the current champion, while under stronger stagnation it uses \textsc{Reframe} to introduce a strategy family absent from the current population (Appendix~\ref{app:prompt-navigator}).

The Navigator receives a two-resolution view rather than the full population: complete code for a few high-ranking niche incumbents and summaries of descriptions and scores from five additional niches. We use three anchors for \textsc{Synthesis}, the champion alone for \textsc{Surgical}, and two anchors for \textsc{Reframe}. This view bounds context while exposing behaviorally distinct alternatives without encouraging verbatim imitation (see Appendix~\ref{app:navigator-details} for details). The Navigator outputs a complete candidate program and its strategy description, and the program is evaluated immediately. If it is valid, the Speculators generate $J{=}5$ nearby variants, allowing a single expensive intervention to seed several low-cost follow-up evaluations.

\paragraph{Advisor: consolidating evidence into guidance.}
Every $\Delta_A{=}50$ evaluations, the Advisor converts recent trajectory statistics and accumulated failure information into a compact prescriptive note. Using $\mathcal{H}_t$ and the archive, it identifies \emph{working} niches whose incumbents improved during the latest interval and uses \emph{saturated} to denote niches with high evaluation activity but no incumbent improvement over the same interval. It also draws on a persistent memory of recurring execution-failure signatures with representative evaluator messages (Appendix~\ref{app:advisor-details}). Based on these signals and its previous note, the Advisor rewrites guidance of at most 500 words specifying which mechanisms to preserve, which directions to de-emphasize, and which recurring errors to avoid. The note thus acts as a rolling, revisable memory rather than an ever-growing transcript. To reduce feedback-induced herding, the note is independently included in each Speculator prompt with probability $\rho_A{=}0.35$, leaving most proposals to explore without explicit Advisor guidance.

\section{Experiments}
\label{sec:experiments}

We address three research questions: \textbf{RQ1:} Does \textsc{SpecEvo} outperform strong LLM-driven evolutionary methods under matched wall-clock, API-cost, and evaluation budgets? \textbf{RQ2:} Does \emph{speculate-then-consolidate} outperform cost-efficient strategies exploiting the asymmetry between lightweight and frontier models? \textbf{RQ3:} How do the Speculator, Navigator, and Advisor individually contribute to exploration diversity, strategic search progress, and failure reduction?

\textbf{Benchmarks.}
We evaluate \textsc{SpecEvo} on 176 real-world scientific and algorithmic discovery tasks spanning four categories: mathematics (7 problems from AlphaEvolve Math~\citep{novikov2025alphaevolve}), systems optimization (4 problems from ADRS~\citep{cheng2025barbarians}), combinatorial optimization (36 problems from CO-Bench~\citep{sun2026co}), and scientific equation discovery
(129 problems from LSR-Synth~\citep{shojaee2025llm}). Detailed benchmark descriptions are provided in
Appendix~\ref{app:benchmark-details}.

\textbf{Baselines and setup.}
Our baselines comprise single-LLM configurations of OpenEvolve~\citep{openevolve}, GEPA~\citep{agrawal2025gepa}, AdaEvolve~\citep{cemri2026adaevolve}, and EvoX~\citep{liu2026evox}, alongside the cost-efficient heterogeneous-model approaches LEVI~\citep{tanveer2026levi} and RelayEvolve~\citep{luo2026relay}. Following RelayEvolve's evaluation design, we further examine four controlled model-allocation strategies: \emph{All-cheap}, which uses only the lightweight model; \emph{Fixed-switch}, which uses the lightweight model for the first half of the calls and the frontier model thereafter; \emph{Random}, which selects either model with equal probability at each call; and \emph{Bandit}, which adaptively selects between the two models using improvements in the best-so-far score as rewards. Appendix~\ref{app:baseline-details} provides their implementation details. We evaluate frontier proprietary and open-weight models: GPT-5~\citep{openai2025gpt5} and Kimi-K2-Thinking (hereafter Kimi-K2;~\citealp{team2025kimi}), together with four Qwen models ranging from 7B to 30B parameters~\citep{qwen2025qwen3}, all accessed through OpenRouter (see Table~\ref{tab:openrouter_model_pricing} for prices). Unless otherwise stated, \textsc{SpecEvo} uses GPT-5 for the Navigator and Qwen3-30B-A3B-Instruct-2507 for the Speculators and Advisor. Experiments run on a dual-socket server with two 16-core Intel Xeon Silver 4314 CPUs, with results averaged over three independent runs. To ensure a fair comparison under the three-hour wall-clock budget, we configure baselines with native support for parallel search to use $W=4$ workers, while retaining their original sequential execution schedules otherwise. Section~\ref{sec:ablation} reports component ablations and hyperparameter sensitivity analyses under the same three-hour wall-clock budget. Full hyperparameters and metric definitions are provided in Appendices~\ref{app:algorithm-details} and~\ref{app:metric-details}, respectively.

\subsection{Main Results}
\label{sec:fixed-time}

\begin{figure}[t]
    \centering
    \includegraphics[width=\linewidth]{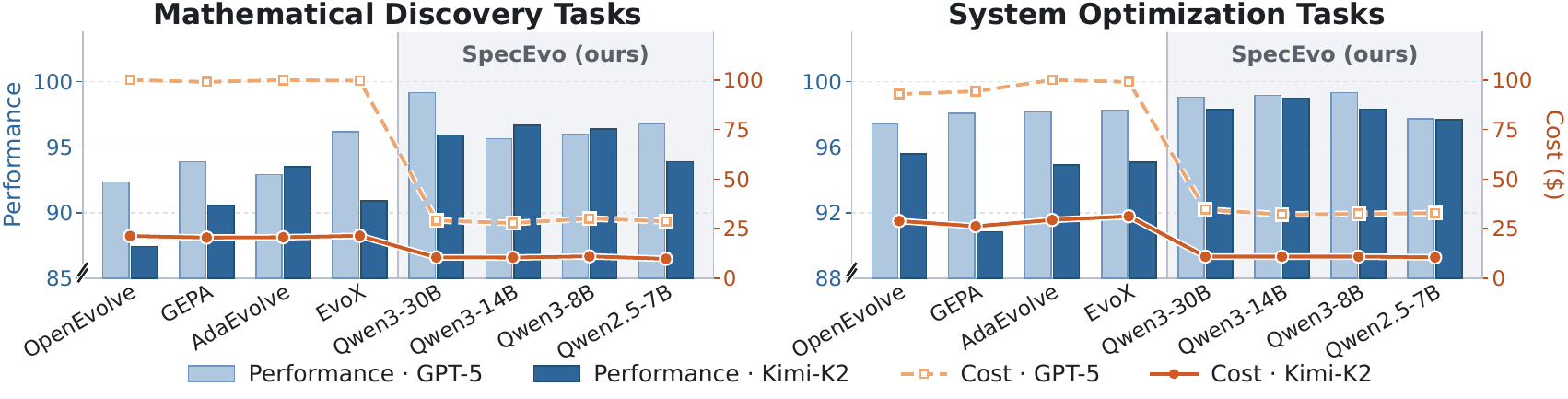}
    \caption{\small Performance under a fixed three-hour wall-clock budget. \textsc{SpecEvo} attains the highest performance in both families while reducing API cost by roughly 50\%--80\%, remaining stable across model configurations.}
    \label{fig:rq1}
\end{figure}

\begin{figure}[t]
    \centering

    \begin{minipage}[t]{0.30\textwidth}
        \centering
        \includegraphics[width=\linewidth]{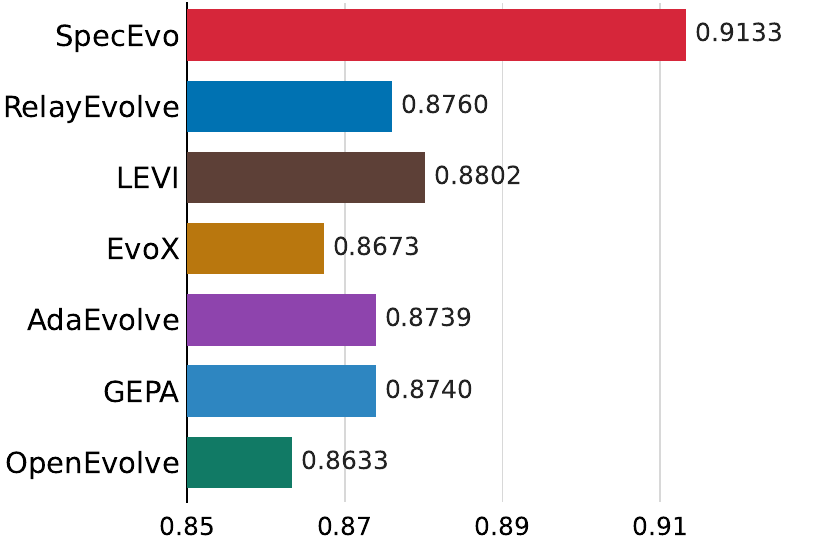}
        \captionof{figure}{\small
        Performance comparison on CO-Bench.
        }
        \label{fig:co-bench-average}
    \end{minipage}
    \hfill
    \begin{minipage}[t]{0.68\textwidth}
        \centering
        \includegraphics[width=\linewidth]{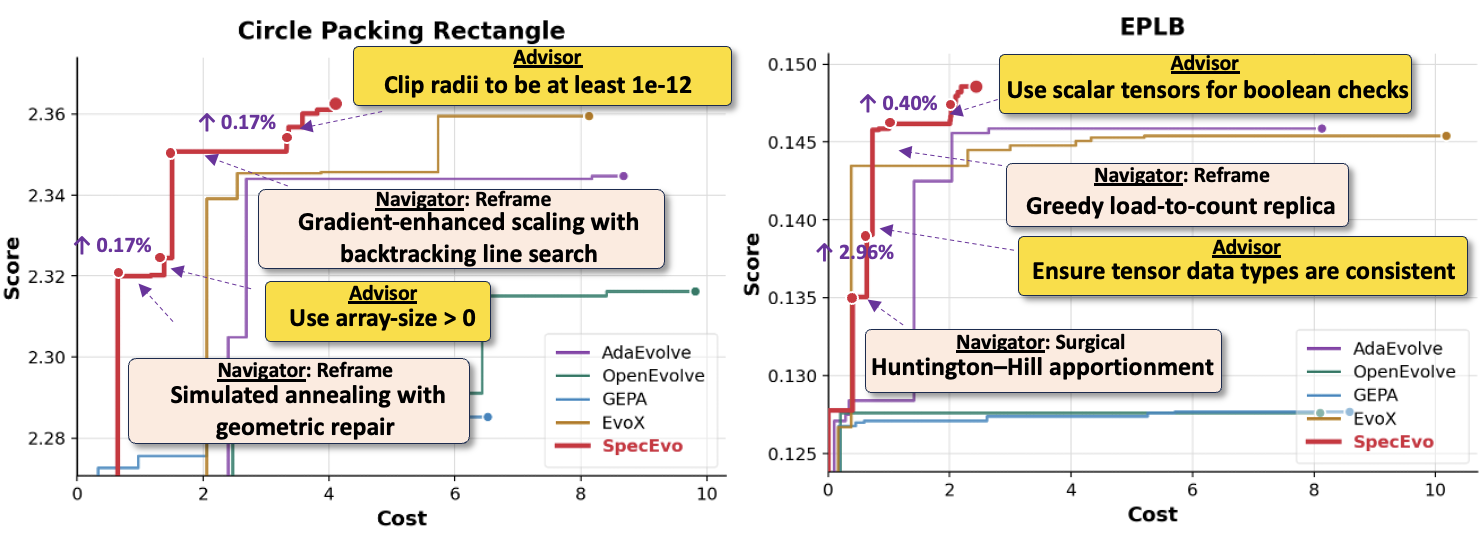}
        \captionof{figure}{\small
        Cost--quality trade-off with GPT-5.
        Navigator and Advisor accelerate search and help escape local optima.
        }
        \label{fig:cost_tradeoff}
    \end{minipage}

\end{figure}

\paragraph{RQ1: Overall effectiveness.}
\textsc{SpecEvo} remains consistently strong across domains, budgets, and model configurations, supporting the contribution of its search organization across the tested backbones. Figures~\ref{fig:rq1} and~\ref{fig:co-bench-average} summarize performance under the three-hour wall-clock budget. We find that \textsc{SpecEvo} leads most mathematical settings and all systems settings, while achieving the highest aggregate CO-Bench score, 3.8\% higher than the strongest competitor, and ranking first on over 50\% of the problems. Across these domains, its API cost is approximately 50\%--80\% lower than that of frontier-heavy baselines under the API pricing used in our experiments. Figure~\ref{fig:trade-off-per-cost} further shows that \textsc{SpecEvo} is \textbf{robust across the tested frontier and lightweight backbones}, whereas competing methods exhibit performance drops of up to \(20\%\) with Qwen3-30B. With GPT-5 as the frontier model, all small-model variants attain the best score of 0.240 on MinMax 14-3 and the maximum score of 26.256 on PRISM. Moreover, while several baselines degrade noticeably when the frontier backbone is changed from GPT-5 to Kimi-K2, \textsc{SpecEvo} remains competitive across configurations (Appendix~\ref{app:breakdown_math}). See Appendices~\ref{app:breakdown_cost}--\ref{app:breakdown_cobench} for detailed results and cost breakdowns. Figure~\ref{fig:cost_tradeoff} further shows that \textsc{SpecEvo} reaches stronger solutions earlier and at substantially lower API cost, whereas other baselines improve more slowly and plateau at lower quality despite greater expenditure.

\begin{figure}[t]
    \centering
\includegraphics[width=\linewidth]{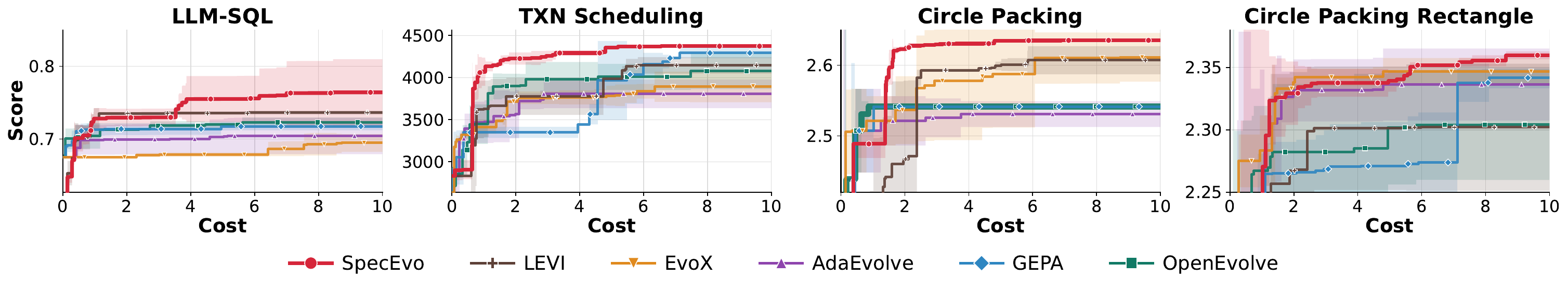}
    \caption{\small Performance comparison under \$10 API cost. }
    \label{fix_dollar}
\end{figure}

Under the \$10 API budget, Figure~\ref{fix_dollar} further shows that \textsc{SpecEvo} outperforms the strongest baseline by shifting routine exploration to inexpensive Speculators and reserving sparse, history-informed frontier calls for strategic intervention, with the largest improvement observed on LLM-SQL (\(5.12\%\) gain). Importantly, Table~\ref{tab:llm_srbench} and Figure~\ref{fig:trade-off-per-cost} (bottom) show that the advantage persists under an identical budget of 500 evaluations on LSR-Synth. In a separate sample-efficiency analysis, we continue the baseline trajectories beyond 500 evaluations and measure
when they first match \textsc{SpecEvo}'s performance at evaluation 500. \textsc{SpecEvo} achieves the lowest NMSE across all eight domain--split settings, while competing methods require approximately 43\%--65\% more evaluations to reach comparable quality. The consistent advantage on OOD data and high symbolic accuracy further suggest that \textsc{SpecEvo} discovers more generalizable functional structure rather than merely fitting observed samples. Further breakdowns and statistical significance analyses are provided in Appendix~\ref{app:breakdown_lsrsynth}.

\paragraph{RQ2: Heterogeneous model coordination.}
\textsc{SpecEvo}'s advantage stems not merely from combining inexpensive and frontier models, but from coordinating their roles using accumulated search evidence. Table~\ref{tab:cost_efficient_math} compares coordination strategies under a common budget of 500 model calls. We find that \textsc{SpecEvo} ranks first on most tasks and never below second, while using the frontier model for only 2.4\% of calls, compared with 89.0\% for RelayEvolve. All-cheap omits frontier reasoning, while Fixed-switch, Random, and Bandit vary model use without the same trajectory-conditioned coordination. RelayEvolve ends lightweight exploration after a one-way handoff, whereas LEVI assigns the frontier model primarily to predefined paradigm shifts. \textsc{SpecEvo} instead combines fixed Navigator checkpoints with stagnation-dependent intervention modes, inexpensive follow-up variants, and Advisor guidance distilled from search history. Since LEVI also uses a small frontier-call share, these results suggest that effective coordination depends on how scarce frontier reasoning informs subsequent search, beyond simply reducing its frequency. 
\begin{table}[t]
\centering
\footnotesize
\setlength{\tabcolsep}{2.4pt}
\renewcommand{\arraystretch}{1.15}
\caption{\small Performance under a budget of 500 model calls. \textsc{SpecEvo} achieves the best performance on five of six tasks, while using the frontier model for around 2.4\% of calls.
}
\label{tab:cost_efficient_math}
\begin{adjustbox}{max width=\textwidth}
\begin{tabular}{@{}lccccccccccccc@{}}
\toprule
\multirow{2}{*}{\textbf{Strategy}}
& \multicolumn{2}{c}{\textbf{Circle Packing}}
& \multicolumn{2}{c}{\textbf{Circle Rect.}}
& \multicolumn{2}{c}{\textbf{Heil. Conv.}}
& \multicolumn{2}{c}{\textbf{Heil. Tri.}}
& \multicolumn{2}{c}{\textbf{MinMax 14-3}}
& \multicolumn{2}{c}{\textbf{Signal}}
& \multirow{2}{*}{\shortstack{\textbf{Frontier Model}\\\textbf{Calls (\%)}}} \\
\cmidrule(lr){2-3}
\cmidrule(lr){4-5}
\cmidrule(lr){6-7}
\cmidrule(lr){8-9}
\cmidrule(lr){10-11}
\cmidrule(lr){12-13}
& \textbf{Avg} & \textbf{Best}
& \textbf{Avg} & \textbf{Best}
& \textbf{Avg} & \textbf{Best}
& \textbf{Avg} & \textbf{Best}
& \textbf{Avg} & \textbf{Best}
& \textbf{Avg} & \textbf{Best}
& \\
\midrule

RelayEvolve
& 2.521 $\pm$ 0.018 & 2.531
& 2.238 $\pm$ 0.020 & 2.249
& 0.022 $\pm$ 0.000 & 0.022
& 0.029 $\pm$ 0.001 & 0.030
& 0.221 $\pm$ 0.005 & 0.224
& 0.677 $\pm$ 0.012 & 0.684
& 89.0 \\

All-cheap
& 2.265 $\pm$ 0.164 & 2.360
& 2.198 $\pm$ 0.018 & 2.219
& 0.020 $\pm$ 0.002 & 0.022
& 0.015 $\pm$ 0.007 & 0.019
& 0.218 $\pm$ 0.006 & 0.222
& 0.673 $\pm$ 0.005 & 0.678
& 0.0 \\

Fixed-switch
& 2.500 $\pm$ 0.000 & 2.500
& \secondcell{2.264} $\pm$ 0.142 & 2.346
& 0.021 $\pm$ 0.001 & 0.021
& 0.028 $\pm$ 0.004 & 0.032
& 0.229 $\pm$ 0.016 & \secondcell{0.238}
& \bestcell{0.725} $\pm$ 0.075 & \bestcell{0.769}
& 50.0 \\

Random
& 2.500 $\pm$ 0.000 & 2.500
& 1.762 $\pm$ 0.505 & 2.345
& \secondcell{0.024} $\pm$ 0.001 & 0.024
& 0.023 $\pm$ 0.004 & 0.027
& 0.224 $\pm$ 0.000 & 0.224
& 0.659 $\pm$ 0.060 & 0.702
& 49.9 \\

Bandit
& 2.471 $\pm$ 0.025 & 2.500
& 2.201 $\pm$ 0.087 & 2.251
& \secondcell{0.024} $\pm$ 0.002 & \secondcell{0.026}
& 0.027 $\pm$ 0.002 & 0.029
& \secondcell{0.233} $\pm$ 0.004 & 0.236
& 0.658 $\pm$ 0.032 & 0.695
& 50.2 \\

LEVI
& \secondcell{2.544} $\pm$ 0.032 & \secondcell{2.607}
& 2.174 $\pm$ 0.109 & \secondcell{2.352}
& 0.023 $\pm$ 0.001 & 0.025
& \secondcell{0.031} $\pm$ 0.002 & \secondcell{0.033}
& 0.231 $\pm$ 0.005 & \secondcell{0.238}
& 0.562 $\pm$ 0.041 & 0.619
& 3.2 \\

\textbf{\textsc{SpecEvo} (Ours)}
& \bestcell{2.602} $\pm$ 0.034 & \bestcell{2.635}
& \bestcell{2.342} $\pm$ 0.011 & \bestcell{2.362}
& \bestcell{0.026} $\pm$ 0.001 & \bestcell{0.027}
& \bestcell{0.033} $\pm$ 0.000 & \bestcell{0.034}
& \bestcell{0.239} $\pm$ 0.001 & \bestcell{0.240}
& \secondcell{0.687} $\pm$ 0.015 & \secondcell{0.725}
& \textbf{2.4} \\

\bottomrule
\end{tabular}
\end{adjustbox}
\end{table}

\begin{figure}[t]
    \centering

    \begin{minipage}[t]{0.32\textwidth}
        \centering
        \includegraphics[width=\linewidth]{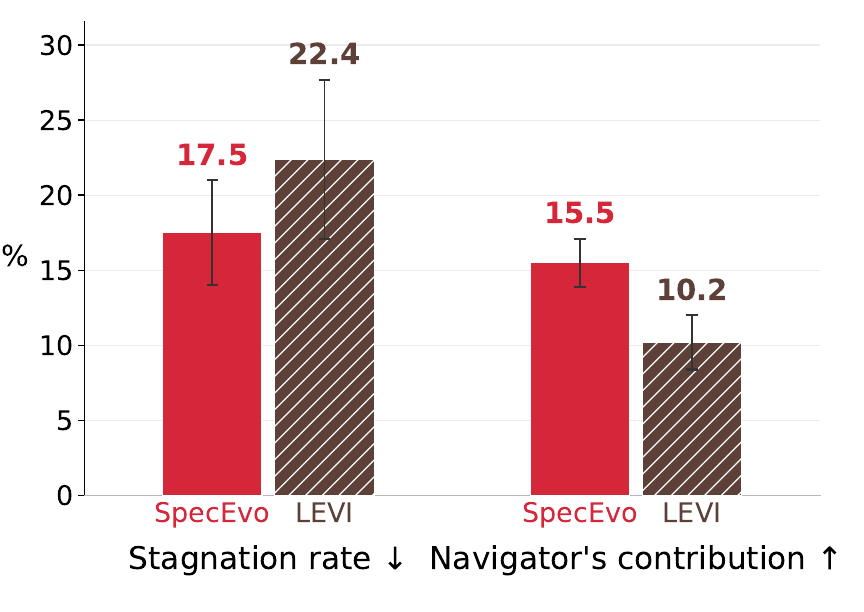}
        \captionof{figure}{\small
        Stagnation rates and frontier-model contributions.
        }
        \label{levi}
    \end{minipage}
    \hfill
    \begin{minipage}[t]{0.32\textwidth}
        \centering
        \includegraphics[width=\linewidth]{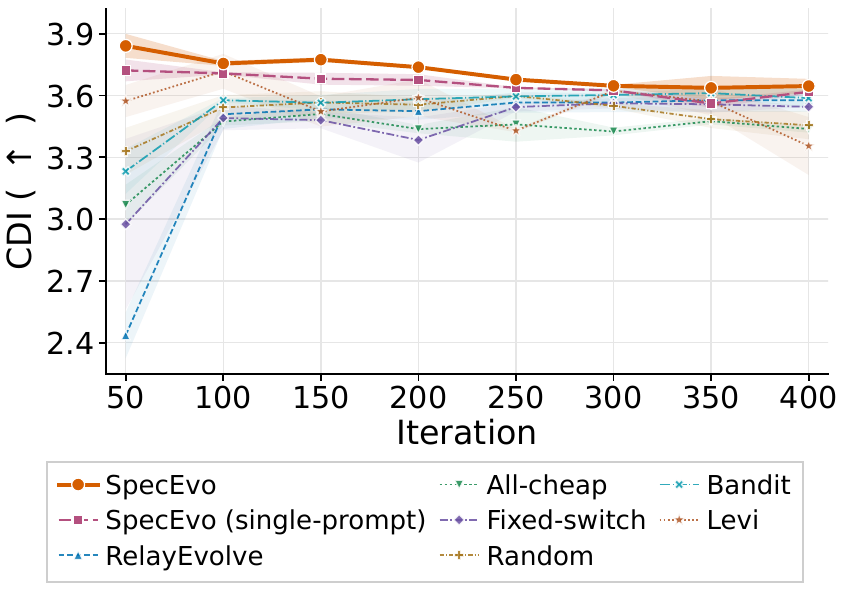}
        \captionof{figure}{\small
        Diversity analysis of population. 
        }
        \label{fig:cdi}
    \end{minipage}
    \hfill
    \begin{minipage}[t]{0.32\textwidth}
        \centering
        \includegraphics[width=\linewidth]{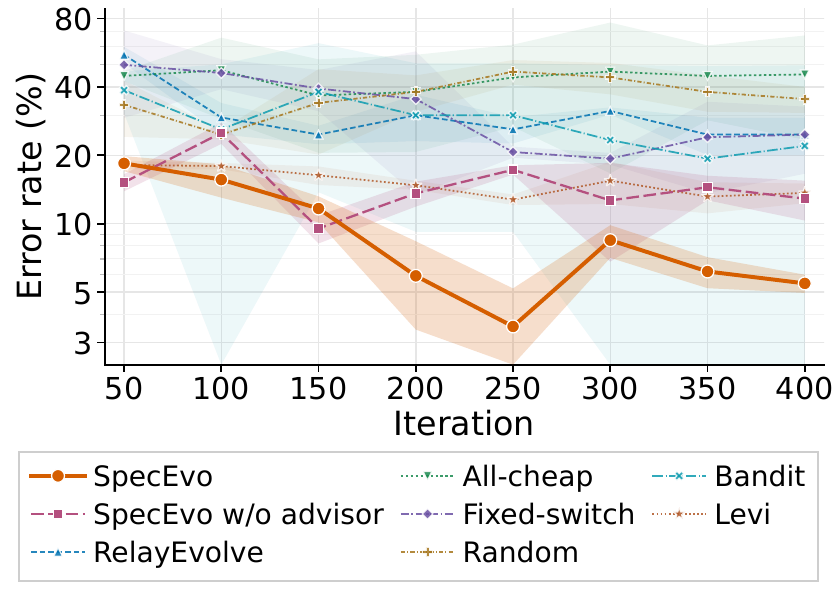}
        \captionof{figure}{\small
        Execution error rates with and without the Advisor.
        }
        \label{fig:error}
    \end{minipage}

\end{figure}

\paragraph{RQ3: Search dynamics.}
The search traces on mathematical discovery tasks reveal how the three roles form a complementary feedback loop.  Figure~\ref{fig:cdi} compares population diversity using the \emph{Code Diversity Index} (CDI) \citep{dat2025hsevo}. CDI is the entropy of normalized positive edge lengths in a minimum spanning tree connecting the programs' code embeddings. Higher values indicate that separation is distributed more evenly across the population. Because CDI also depends on population size, we compare methods at checkpoints with comparable archive sizes. \textsc{SpecEvo} maintains higher CDI than its single-prompt ablation and LEVI as search progresses. Speculators combine stagnation-adaptive parent sampling with specialized variation prompts to counter lightweight models' tendency to repeat familiar structures \citep{dat2025hsevo}. The archive retains distinct strategy families, including analytic template-based constructions, penalty-annealed search, and neighbor-query-accelerated local refinement (examples in Appendix~\ref{app:qualitative}).

Figure~\ref{levi} summarizes stagnation and frontier-model contribution using statistics computed from the search traces. We define stagnation rate as the longest consecutive span of evaluations without a new global best, divided by the total number of evaluations. The reported rate is $17.5\%$ for \textsc{SpecEvo}, compared with $22.4\%$ for LEVI. Frontier-model contribution measures the fraction of new-best events produced either directly by the frontier model or by lightweight offspring with a frontier-generated parent. This fraction is $15.5\%$ for \textsc{SpecEvo}, compared with $10.2\%$ for LEVI, capturing improvements from both Navigator proposals and their subsequent lightweight refinement. At fixed checkpoints, the Navigator selects among synthesis, surgical refinement, and reframing according to stagnation. The annotated examples in Figure~\ref{fig:cost_tradeoff} illustrate how these interventions change the search strategy and are followed by further score improvements.
It also provides concrete examples of the Advisor turning execution failures into reusable instructions. On Circle Packing Rectangle, it recommends explicit nonempty-array checks and clipping radii to at least $10^{-12}$. On EPLB, it recommends scalar Boolean checks and consistent tensor data types. These instructions address specific causes of rejected candidates, allowing subsequent proposals to retain useful algorithmic ideas while correcting their implementation. Consistent with this role, Figure~\ref{fig:error} shows that \textsc{SpecEvo} reaches execution error rates of approximately $5\%$--$8\%$, compared with $13\%$--$25\%$ for LEVI. The higher error rates of the variant without the Advisor further support the benefit of carrying failure-related guidance across attempts. Advice is included in only a subset of Speculator prompts, leaving the remaining proposals free to explore without this guidance.

\subsection{Ablation Studies}
\label{sec:ablation}
\begin{figure}[t]
    \centering
    \includegraphics[width=0.85\linewidth]{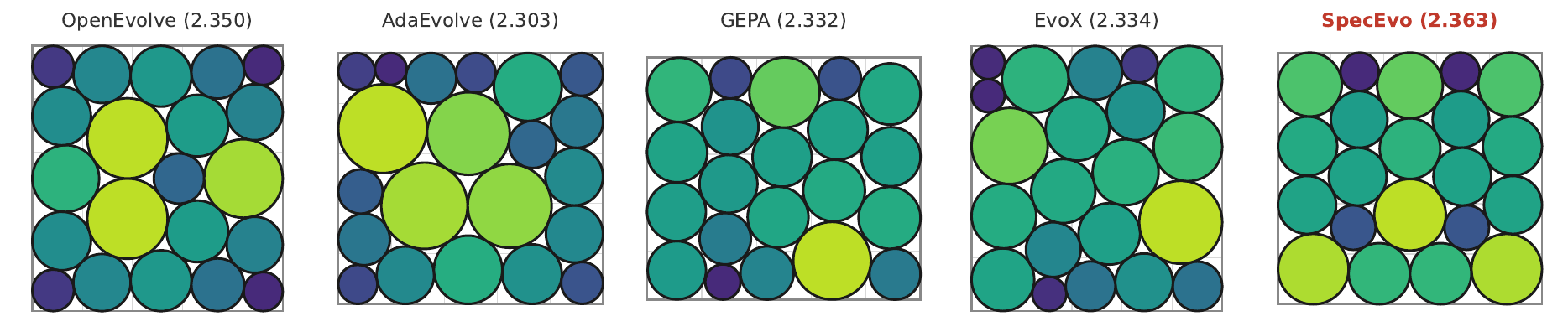}
\caption{\small Best solutions generated by different methods for the Circle Packing Rectangle task.
}
\label{fig:circle_packing_rectangle_solutions}
\end{figure}
\begin{figure}[t]
    \centering

    \begin{minipage}[t]{0.53\textwidth}
        \centering
        \includegraphics[width=\linewidth]{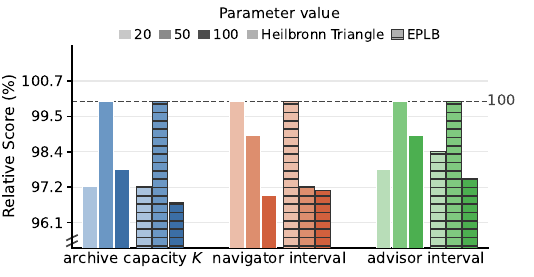}
        \captionof{figure}{\small
        Sensitivity to archive capacity $K$ and
        Navigator/Advisor intervals.
        }
        \label{fig:hyperparam}
    \end{minipage}
    \hfill
    \begin{minipage}[t]{0.45\textwidth}
        \centering
        \includegraphics[width=\linewidth]{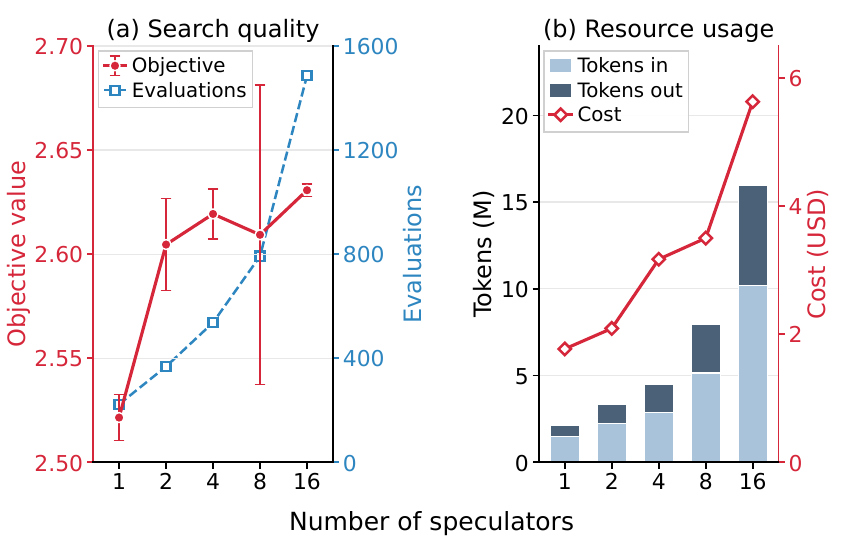}
        \captionof{figure}{\small
        Effect of the number of Speculators on search quality
        and resource usage.
        }
        \label{fig:scale}
    \end{minipage}

\end{figure}

We ablate the speculation and consolidation components of \textsc{SpecEvo}'s StC workflow (Table~\ref{ablation}). For speculation, we remove AST features or description embeddings from the archive descriptor, disable periodic refits, or replace the six specialized prompts with a single general mutation operator. For consolidation, we remove the Navigator or replace its adaptive mode selection with random selection or a fixed mode (Synthesis, Surgical, or Reframe). We also remove the Advisor or replace its evolving guidance with fixed advice. We further assess sensitivity to the archive capacity $K$, Navigator and Advisor intervals, and Speculator count (Figures~\ref{fig:hyperparam}, \ref{fig:scale}).

\begin{table*}[t]
\centering
\captionsetup{font=small, skip=4pt}

\scriptsize
\setlength{\tabcolsep}{1.8pt}
\renewcommand{\arraystretch}{1.08}

\sbox{\LSRTableBox}{%
\resizebox{0.56\textwidth}{!}{%
\begin{tabular}{@{}llccccccc@{}}
\toprule
\textbf{Domain}
& \textbf{Split}
& \textbf{OE}
& \textbf{GEPA}
& \textbf{AdaEvo}
& \textbf{EvoX}
& \textbf{RelayEvo}
& \textbf{LEVI}
& \textbf{\textsc{SpecEvo}} \\
\midrule

\multirow{2}{*}{Chemistry}
& ID
& 5.27E-06 & 1.72E-05 & 2.48E-06 & 6.87E-06
& 2.59E-06 & 5.80E-06 & \bestcell{4.57E-07} \\
& OOD
& 4.99E-02 & 2.69E-01 & 6.05E-02 & 8.68E-02
& 1.30E-01 & 1.17E-01 & \bestcell{1.68E-02} \\
\midrule

\multirow{2}{*}{Biology}
& ID
& 6.33E-05 & 8.79E-05 & 9.92E-06 & 3.79E-05
& 1.01E-05 & 3.87E-06 & \bestcell{3.34E-07} \\
& OOD
& 7.48E-01 & 1.92E+00 & 1.28E-01 & 5.83E-01
& 1.56E-01 & 1.96E-02 & \bestcell{3.74E-03} \\
\midrule

\multirow{2}{*}{Materials}
& ID
& 5.92E-08 & 1.86E-07 & 6.32E-07 & 9.05E-08
& 5.79E-08 & 5.73E-07 & \bestcell{4.84E-08} \\
& OOD
& 2.44E-06 & 7.57E-05 & 1.18E-04 & 1.63E-05
& 6.72E-06 & 8.36E-06 & \bestcell{1.68E-06} \\
\midrule

\multirow{2}{*}{Physics}
& ID
& 3.29E-04 & 1.53E-03 & 2.79E-04 & 6.66E-04
& 1.81E-04 & 1.38E-05 & \bestcell{2.92E-06} \\
& OOD
& 6.93E-03 & 2.82E-02 & 1.32E-02 & 5.32E-02
& 3.30E-03 & 2.23E-03 & \bestcell{4.75E-04} \\
\midrule

\multicolumn{2}{l}{\textbf{Evals. to Beat \textsc{SpecEvo}}}
& \makecell[c]{764}
& \makecell[c]{824}
& \makecell[c]{715}
& \makecell[c]{780}
& \makecell[c]{768}
& \makecell[c]{752}
& -- \\
\bottomrule
\end{tabular}%
}%
}

\fontsize{6}{7}\selectfont
\setlength{\tabcolsep}{3pt}
\renewcommand{\arraystretch}{1.00}
\setlength{\aboverulesep}{0.6pt}
\setlength{\belowrulesep}{1.0pt}

\setlength{\AblationRowGap}{0pt}
\setlength{\AblationLastGap}{0pt}
\newcommand{\AblationHeat}[2]{\cellcolor{red!#1}$\downarrow$\,#2\%}

\newcommand{\AblationTableContent}{%
\begin{tabular*}{0.40\textwidth}
{@{\extracolsep{\fill}}lcc@{}}
\toprule
\textbf{Ablation setting}
& \makecell[c]{\textbf{Circle Packing}\\\textbf{Rectangle}}
& \makecell[c]{\textbf{Heilbronn}\\\textbf{Triangle}} \\
\midrule

w/o AST features
& \AblationHeat{4}{0.38}
& \AblationHeat{24}{7.76} \\[\AblationRowGap]

w/o description embedding
& \AblationHeat{1}{0.13}
& \AblationHeat{20}{6.37} \\[\AblationRowGap]

w/o refit
& \AblationHeat{17}{1.65}
& \AblationHeat{16}{4.99} \\[\AblationRowGap]
\midrule

Single operator prompt
& \AblationHeat{23}{2.17}
& \AblationHeat{40}{12.74} \\[\AblationRowGap]
\midrule

w/o Navigator
& \AblationHeat{24}{2.25}
& \AblationHeat{30}{9.42} \\[\AblationRowGap]

Random mode
& \AblationHeat{27}{2.60}
& \AblationHeat{23}{7.47} \\[\AblationRowGap]

Synthesis mode only
& \AblationHeat{28}{2.71}
& \AblationHeat{3}{0.83} \\[\AblationRowGap]

Surgical mode only
& \AblationHeat{5}{0.49}
& \AblationHeat{39}{12.47} \\[\AblationRowGap]

Reframe mode only
& \AblationHeat{21}{1.97}
& \AblationHeat{24}{7.76} \\[\AblationRowGap]
\midrule

w/o Advisor
& \AblationHeat{40}{3.82}
& \AblationHeat{19}{6.09} \\[\AblationRowGap]

Fixed guidance
& \AblationHeat{36}{3.46}
& \AblationHeat{26}{8.31} \\[\AblationLastGap]

\bottomrule
\end{tabular*}%
}
\sbox{\AblationTableBox}{\AblationTableContent}

\setlength{\AblationRowGap}{%
  \dimexpr(
    \ht\LSRTableBox+\dp\LSRTableBox
    -\ht\AblationTableBox-\dp\AblationTableBox
  )/11\relax
}

\setlength{\AblationLastGap}{%
  \dimexpr
    \ht\LSRTableBox+\dp\LSRTableBox
    -\ht\AblationTableBox-\dp\AblationTableBox
    -10\AblationRowGap
  \relax
}

\sbox{\AblationTableBox}{\AblationTableContent}

\renewcommand{\arraystretch}{1}

\begin{tabular}{@{}c@{\hspace{0.02\textwidth}}c@{}}

\begin{minipage}[t]{0.56\textwidth}
\vspace{0pt}
\captionof{table}{
Performance on LSR-Synth under 500 evaluations.
The final row reports the total evaluations required
for each baseline to match \textsc{SpecEvo}@500.}
\label{tab:llm_srbench}
\end{minipage}
&
\begin{minipage}[t]{0.40\textwidth}
\vspace{0pt}
\captionof{table}{
Ablation study on representative tasks. Relative decreases in mean score
from the full \textsc{SpecEvo} configuration.}
\label{ablation}
\end{minipage}
\\[2pt]

\raisebox{\depth}{\usebox{\LSRTableBox}}
&
\raisebox{\depth}{\usebox{\AblationTableBox}}

\end{tabular}
\end{table*}

\textbf{Component contributions.} Table~\ref{ablation} supports two complementary sources of improvement: diverse exploration and consolidation that adapts to the search trajectory. Structural and semantic descriptors preserve distinct implementations and strategies, while periodic refits keep niches aligned with the evolving population. Specialized Speculator prompts sustain complementary search moves; collapsing them into a single operator produces the largest exploration-side degradation, particularly on Heilbronn Triangle. The consolidation ablations highlight the importance of conditioning feedback on search state. Random Navigator modes underperform on both tasks, while the strongest fixed mode differs across tasks, supporting the value of matching intervention scope to stagnation. Fixed guidance likewise underperforms adaptive Advisor updates and even yields lower mean performance than removing the Advisor on Heilbronn Triangle. This suggests that static advice can become counterproductive as the search evolves, motivating continual revision based on productive mechanisms, saturated directions, and recurring failures.

\textbf{Case studies.}
Figure~\ref{fig:cost_tradeoff} illustrates this division of labor, while Figure~\ref{fig:circle_packing_rectangle_solutions} shows the final packing solutions. On Circle Packing Rectangle, routine refinement initially concentrates on similar geometric constructions. The Navigator then uses \textsc{Reframe} to introduce simulated annealing with geometric repair and, later, gradient-enhanced scaling with backtracking line search. The Advisor complements these shifts with safeguards for empty arrays and near-zero radii, allowing subsequent Speculator variants to explore the new neighborhoods without repeatedly failing on numerical edge cases. A similar pattern appears on EPLB: \textsc{Surgical} first strengthens the incumbent through Huntington--Hill apportionment and load-aware heap scheduling, while \textsc{Reframe} later replaces the saturated placement logic with load-to-count allocation and rotated round-robin assignment. Between these interventions, the Advisor helps prevent recurring dtype and scalar-boolean errors. These interventions provide reusable starting points for continued low-cost Speculator refinement.

\textbf{Hyperparameter sensitivity.}
In Figure~\ref{fig:hyperparam}, we find that $K=50$ yields the strongest performance, suggesting a balance between strategy coverage and sufficient refinement within each niche. The preferred Navigator and Advisor intervals are \(20\) and \(50\) evaluations, respectively. This difference is consistent with their informational needs: strategic intervention benefits from timely responses to stagnation, whereas reusable guidance requires enough observations to distinguish recurring patterns from isolated outcomes. Figure~\ref{fig:scale} shows that scaling from one to four Speculators improves both solution quality and evaluation throughput on Circle Packing. Beyond four, throughput continues to increase, but quality gains become marginal while token usage and API cost rise sharply. We therefore use four Speculators to capture most of the benefit of parallel exploration at moderate cost.

\section{Conclusion}


We introduced \textsc{SpecEvo}, a multi-LLM framework that coordinates low-cost exploration and sparse frontier-model reasoning through continual consolidation of search evidence. Its \emph{speculate-then-consolidate} workflow combines parallel lightweight Speculators, persistent Advisor guidance distilled from successes, failures, and saturated directions, and stagnation-dependent Navigator interventions at fixed checkpoints. Across 176 scientific and algorithmic discovery tasks, \textsc{SpecEvo} matches or outperforms strong baselines under matched wall-clock, API-cost, and evaluation budgets, with 50\%--80\% lower API expenditure under the three-hour wall-clock budget. Its gains across backbone configurations and under matched evaluation budgets support the value of using accumulated search evidence to guide sparse frontier interventions and subsequent lightweight refinement.

\section*{AI Use Statement}

\paragraph{LLMs in the method and evaluation.}
LLMs are a core component of the method studied in this work. \textsc{SpecEvo} and all baselines call the language models listed in Table~\ref{tab:openrouter_model_pricing} to generate, refine, and analyze candidate programs (Section~\ref{sec:experiments} and Appendix~\ref{app:baseline-details}), and we use \texttt{text-embedding-3-small} for \textsc{SpecEvo}'s archive descriptors and for the diversity metric (Appendices~\ref{app:algorithm-details} and~\ref{app:metric-details}). To score symbolic accuracy on LSR-Synth, we follow the LLM-SRBench protocol~\citep{shojaee2025llm} and use GPT-4o to judge whether a discovered equation is mathematically equivalent to the ground truth (Appendix~\ref{app:breakdown_lsrsynth}). These uses are part of the experimental method and evaluation protocol described in the paper.

\paragraph{AI assistance in research and writing.}
We used generative AI tools to assist with writing, including language editing, improving clarity and readability, and condensing or rephrasing parts of the manuscript. The authors reviewed and revised all AI-assisted text. We take responsibility for the final content of this work, including text, claims, or artifacts produced with the aid of generative AI.


\bibliography{iclr2027_conference}
\bibliographystyle{iclr2027_conference}

\newpage
\appendix

\PrintAppendixContents
{Appendix}
{
    \AppendixTOCEntry{app:limit}
    \AppendixTOCEntry{app:related-work}
    \AppendixTOCEntry{app:baseline-details}
    \AppendixTOCEntry{app:algorithm-details}
    \AppendixTOCEntry{app:behavioral-descriptor}
    \AppendixTOCEntry{app:specevo-main}
    \AppendixTOCEntry{app:navigator-details}
    \AppendixTOCEntry{app:advisor-details}
    \AppendixTOCEntry{app:metric-details}
    \AppendixTOCEntry{app:fig3-metrics}
    \AppendixTOCEntry{app:cdi}
    \AppendixTOCEntry{app:stagnation-rate}
    \AppendixTOCEntry{app:frontier-new-bests}

    \AppendixTOCEntry{app:benchmark-details}
    \AppendixTOCEntry{app:math-benchmarks}
    \AppendixTOCEntry{app:system-benchmarks}
   
    \AppendixTOCEntry{app:cobench}
    \AppendixTOCEntry{app:llmsrbench}
    \AppendixTOCEntry{app:add_result}
    \AppendixTOCEntry{app:breakdown_cost}

     \AppendixTOCEntry{app:breakdown_math}
    \AppendixTOCEntry{app:breakdown_cobench}
    \AppendixTOCEntry{app:breakdown_lsrsynth}
    \AppendixTOCEntry{app:prompt-templates}

    \AppendixTOCEntry{app:prompt-speculator}

    \AppendixTOCEntry{app:prompt-navigator}

    \AppendixTOCEntry{app:prompt-advisor}
    \AppendixTOCEntry{app:qualitative}
    \AppendixTOCEntry{app:qualitative:cpr}
    \AppendixTOCEntry{app:qualitative:eplb}
}

\section{Limitations}
\label{app:limit}

\paragraph{Dependence on parallel computing infrastructure.}
The wall-clock efficiency of \textsc{SpecEvo} partly depends on the amount of parallelism available to the Speculator pool. Our experiments use \(W=4\) concurrent Speculators on a dual-socket server with two 16-core Intel Xeon Silver 4314 processors. Consequently, the number of candidates explored within a fixed wall-clock budget may change with the worker count, CPU capacity, memory bandwidth, network latency, and API rate limits. Although our fixed API-cost and fixed evaluation-budget experiments provide complementary controls, they do not make the wall-clock results hardware-independent. The reported cost reductions should therefore be interpreted primarily as savings in LLM API expenditure under the specified computing environment, rather than reductions in total computational cost.

\paragraph{Compute-intensive candidate evaluation.}
The benchmarks considered in this work use relatively lightweight evaluators: candidate programs can generally be executed and scored on CPUs without repeatedly training large neural networks, running long simulations, or conducting physical experiments. This setting allows multiple Speculators to evaluate proposals concurrently at modest local-compute cost. For tasks whose evaluation requires GPU training, large-scale inference, expensive simulation, or real-world experimentation, increasing the number of parallel workers may instead multiply accelerator usage, memory requirements, and the cost of unsuccessful trials. Moreover, a fixed number of evaluations does not represent a fixed compute budget when candidate evaluations have heterogeneous costs. Evaluating \textsc{SpecEvo} in such settings is an important direction for future work. A more comprehensive study should jointly control API expenditure, CPU- or GPU-hours, peak memory, energy consumption, wall-clock time, and the number of evaluated candidates. It would also be valuable to adapt the worker count and the frequencies of Navigator and Advisor intervention dynamically according to evaluator cost, available hardware, and observed search progress.

\paragraph{Reliability of model-generated descriptions.}
The semantic component of our archive descriptor relies on descriptions produced by the proposing model, predominantly lightweight Speculators.
These descriptions may omit important changes, remain generic across distinct programs, or claim mechanisms absent from the code. Such inaccuracies can distort semantic distances, grouping different strategies together or separating similar implementations, thereby affecting niche
assignment, parent selection, and the population summaries provided to the Navigator and Advisor. 
We have not systematically quantified sensitivity to description quality. Future work could evaluate this sensitivity through controlled description perturbations and explore code-grounded verification or descriptors derived from execution behavior.

\section{Extended Related Work}
\label{app:related-work}

\paragraph{LLM-Guided Evolutionary Search.} LLM-guided evolutionary search combines language models as semantic variation operators with executable evaluators and population-based optimization. FunSearch established this paradigm by evolving programs for open mathematical problems~\citep{romera2024mathematical}, while ELM combined pretrained code models with quality-diversity search~\citep{lehman2023evolution}. Subsequent methods specialize the evolutionary loop in different ways: EoH jointly evolves natural-language ideas and executable heuristics~\citep{liu2024evolution}; ReEvo derives reflective guidance from performance comparisons~\citep{ye2024reevo}; LLaMEA evolves optimization algorithms through iterative evaluation~\citep{van2024llamea}; HSEvo explicitly balances diversity and convergence~\citep{dat2025hsevo}; and MPaGE extends heuristic evolution to multiple objectives~\citep{hieu2026pareto}. Related systems apply LLM-guided search to reward design~\citep{ma2024eureka}, preference optimization~\citep{lu2024discovering}, and scientific equation discovery~\citep{li2026evosr, shojaee2025llm,   shojaee2025llma, grayeli2024symbolic, ma2024llm}. More general evolutionary coding agents include AlphaEvolve~\citep{novikov2025alphaevolve}, OpenEvolve~\citep{openevolve}, and ShinkaEvolve~\citep{lange2026shinkaevolve}, while GEPA evolves prompts and natural-language rules through reflection and Pareto selection~\citep{agrawal2025gepa}. Recent work further improves the representation and control of search: SeaEvo organizes candidates into persistent semantic strategy families~\citep{luo2026seaevo}; EvoX evolves the search strategy itself~\citep{liu2026evox}; AdaEvolve adapts exploration intensity, resource allocation across islands, and meta-level guidance~\citep{cemri2026adaevolve}; and CORAL delegates search decisions to autonomous peer agents communicating through shared memory~\citep{qu2026coral}. Unlike these systems, \textsc{SpecEvo} assigns routine exploration, experience consolidation, and strategic intervention to models of different cost, analogous to the division of labor in a research laboratory.


\paragraph{Cost-Aware LLM Routing and Evolutionary Search Allocation.}
Cost-aware LLM systems reduce inference expense by selecting among models with different capabilities. FrugalGPT constructs cascades that progressively escalate difficult queries~\citep{chen2023frugalgpt}, AutoMix uses self-verification to decide whether a small-model response requires augmentation by a stronger model~\citep{aggarwal2024automix}, and RouteLLM learns routing policies from human preference data~\citep{ong2025routellm}. These methods focus on independent inference requests, whereas evolutionary search must allocate resources across a sequence of interdependent population updates. ShinkaEvolve addresses this setting with a bandit that selects mutation models according to their observed performance~\citep{lange2026shinkaevolve}. AdaptEvolve instead runs a small model first and uses token-level confidence statistics with an online decision-tree router to determine whether the mutation should be regenerated by a larger model~\citep{ray2026adaptevolve}. AdaEvolve allocates computation across concurrent evolutionary processes according to their recent progress and introduces meta-level reasoning to adjust search behavior under stagnation~\citep{cemri2026adaevolve}. RelayEvolve shifts allocation from individual mutations to entire search phases: a cheap model develops multiple trajectories in short blocks, Relay Gain measures the marginal quality and coverage of their candidate bank, and saturation of this signal triggers a one-time handoff of a curated seed population to strong-model evolution~\citep{luo2026relay}. LEVI combines three complementary mechanisms: bootstrapped CVT-MAP-Elites for diversity preservation, role-aware routing that assigns local refinements to smaller models and paradigm shifts to a stronger model, and rank-preserving proxy benchmarks that reduce evaluation overhead~\citep{tanveer2026levi}.

These approaches optimize different components of resource allocation but retain important structural restrictions. ShinkaEvolve and AdaptEvolve select among models at the mutation level, leaving them functionally interchangeable; AdaptEvolve additionally requires token-level probabilities that are unavailable from many proprietary APIs. AdaEvolve adapts the amount and direction of search effort but does not organize heterogeneous models into persistent functional roles. RelayEvolve captures population-level state more explicitly, but information flows through a single cheap-to-strong transition: once handoff occurs, lightweight exploration ends, and the strong phase inherits selected candidates but no consolidated knowledge about discarded attempts, recurring failures, or exhausted trajectories. LEVI provides a clearer division between local and structural mutations, but its stronger model is assigned primarily to a predefined paradigm-shift operation, without a dedicated component that continually aggregates trajectory-wide evidence for subsequent small-model search. \textsc{SpecEvo} instead treats model heterogeneity as a persistent division of labor throughout evolution. Its Speculators perform high-throughput variation, its Advisor repeatedly converts successes, failures, and niche-level progress into reusable guidance, and its Navigator examines the population at sparse checkpoints to choose among synthesis, surgical refinement, and reframing. Where prior methods optimize model selection, phase transition, or mutation type, \textsc{SpecEvo} coordinates candidate generation, experience consolidation, and strategic intervention within the same evolutionary process.

\section{Baseline Details}
\label{app:baseline-details}

\paragraph{Common evaluation protocol.}
We evaluate all baselines using their native search logic and publicly released implementations. We introduce only the task adapters required to expose the same problem specification, editable program scaffold, execution environment, and evaluator to every method. 
All three-hour wall-clock experiments use the same hardware and evaluation environment. For baseline implementations supporting native parallel candidate search, we configure four concurrent workers,
matching \textsc{SpecEvo}. The wall-clock results therefore characterize end-to-end performance under this protocol, complemented by comparisons under
fixed API-cost and evaluation budgets.

All initialization, mutation, repair, reflection, meta-guidance, and paradigm-shift calls are included in API-cost and wall-clock accounting. API expenditure is computed from the actual input and output tokens of each call using the prices in Table~\ref{tab:openrouter_model_pricing}. Under an evaluation budget, every candidate submitted to the evaluator consumes one evaluation, including programs that fail to parse, violate constraints, or time out; auxiliary calls that do not produce a candidate do not increment this counter but still contribute to API cost and elapsed time. Under the 500-call RQ2 budget, every LLM invocation counts toward the common limit. Each method returns the highest-scoring valid program discovered before its stopping condition is reached.

\paragraph{OpenEvolve.}
OpenEvolve~\citep{openevolve} implements asynchronous island-based program evolution with a MAP-Elites-style database. Candidate parents are selected from a mixture of high-performing and behaviorally diverse programs, mutated using evaluator feedback, and periodically exchanged across islands. We retain its native archive, migration, and prompt construction mechanisms, while assigning the selected frontier backbone to all LLM calls.

\paragraph{GEPA.}
We use the inference-time code-optimization configuration of GEPA~\citep{agrawal2025gepa}, treating the candidate program as the textual artifact being optimized. GEPA converts execution traces and evaluator diagnostics into natural-language reflection, uses the resulting analysis to propose targeted mutations, and maintains complementary candidates through Pareto-aware selection and merging. The same frontier backbone is used for both reflection and candidate generation.

\paragraph{AdaEvolve.}
AdaEvolve~\citep{cemri2026adaevolve} derives an accumulated-improvement signal from the search trajectory and uses it at three levels: adjusting exploration intensity within each island, allocating evaluations across islands through a bandit scheduler, and generating new high-level tactics when global progress stalls. We use the authors' released implementation and default adaptive schedules, with the designated frontier backbone serving both mutation and meta-guidance calls.

\paragraph{EvoX.}
EvoX~\citep{liu2026evox} jointly evolves candidate programs and the strategy used to generate them. Its inner loop evaluates programs under the current search strategy, while its outer loop periodically revises parent selection and variation behavior according to the progress achieved during the preceding search window. We preserve this two-level evolution and use the same frontier backbone for solution generation and search-strategy evolution.

\paragraph{LEVI.}
LEVI~\citep{tanveer2026levi} first generates diverse seeds to calibrate a CVT-MAP-Elites archive and then applies role-aware model routing. Qwen3-30B performs ordinary local refinements, whereas the frontier model is reserved for periodically triggered paradigm shifts conditioned on representatives from distinct archive regions. We retain its native archive descriptors, trigger policy, lightweight follow-up variants, and proxy-benchmark construction where applicable. All reported final programs are re-evaluated using the complete task evaluator.

\paragraph{RelayEvolve.}
RelayEvolve~\citep{luo2026relay} begins with multiple lightweight-model trajectories whose search blocks are allocated by a Grow--Deepen bandit. Relay Gain measures the marginal quality and coverage contributed to an online candidate bank and triggers handoff when further cheap-model exploration saturates. The terminal candidate pool is then curated using the method's quality--diversity objective, and the selected population initializes a shared frontier-model search using the remaining budget. We retain the default strong-model budget reserve and all handoff hyperparameters. Once handoff occurs, lightweight-model exploration does not resume.

\paragraph{Controlled model-allocation strategies.}
To isolate the effect of model coordination from the mere availability of two differently priced models, we additionally implement four controls on a shared asynchronous evolutionary backend. They use identical parent selection, population updates, mutation prompts, evaluator feedback, and context limits, differing only in how the lightweight and frontier models are assigned. For a total call budget \(B\), these strategies are defined as follows:
\begin{itemize}
    \item \textbf{All-cheap} assigns every call to the lightweight model.
    
    \item \textbf{Fixed-switch} uses the lightweight model for the first \(B/2\) calls and then performs a one-time switch to the frontier model for the remaining calls, carrying the current population forward without additional curation.
    
    \item \textbf{Random} independently selects the lightweight or frontier model with probability \(0.5\) for every call.
    
    \item \textbf{Bandit} treats the two models as arms of a UCB1 scheduler. After initializing both arms, model selection balances their observed mean rewards with an exploration bonus. The reward is the non-negative improvement in the global best score produced by the selected model; invalid or non-improving proposals receive zero reward.
\end{itemize}

\section{Algorithm Details}
\label{app:algorithm-details}
We set the temperatures to \(0.5\), \(0.5\), and \(0.8\) for the
\emph{synthesis}, \emph{surgical}, and \emph{reframe} modes, respectively,
and to \(0.7\) and \(0.4\) for the Speculator and Advisor. We set
\(H_g=100\), \(H_l=20\), \(\beta_{\min}=0.3\), \(\beta_{\max}=2.0\),
\(\gamma_1=0.4\), and \(\gamma_2=0.7\). We refit the archive descriptor space every \(\Delta_C=50\) completed evaluations, setting \(\Delta_C=\Delta_A\) for
consistency with the Advisor update interval. After each Navigator call generates a set of hypotheses, the Speculators produce \(J=5\) variants for each hypothesis.

\subsection{Behavioral Descriptor Construction}
\label{app:behavioral-descriptor}

We provide the implementation details of the structural--semantic descriptor
introduced in Eq.~\ref{eq:descriptor}. For a program \(p\), we first parse its source code into
a Python abstract syntax tree and extract the following eight statistics, in
fixed order:
\[
\begin{split}
a(p) = [\,
& a_{\mathrm{depth}},
  a_{\mathrm{cyc}},
  a_{\mathrm{loop}},
  a_{\mathrm{nest}},
  a_{\mathrm{branch}},
  a_{\mathrm{func}},
  a_{\mathrm{comp}},
  a_{\mathrm{call}}
\,].
\end{split}
\]
These denote the maximum AST depth; cyclomatic complexity; the number of
\texttt{For} and \texttt{While} nodes; the maximum nesting depth of such
loops; and the numbers of \texttt{If}, \texttt{FunctionDef}, comprehension,
and function-call nodes, respectively. Cyclomatic complexity is initialized
to one, incremented for each \texttt{If}, \texttt{While}, \texttt{For}, and
\texttt{ExceptHandler}, and additionally incremented by the Boolean fan-out
of each \texttt{BoolOp}. To reduce the influence of unusually large programs,
we apply an element-wise logarithmic transform,
\[
    \phi(p) = \log\!\bigl(1 + a(p)\bigr) \in \mathbb{R}^{8}.
\]
As a defensive fallback, an unparsable program is assigned an all-zero
structural vector, although such programs are normally rejected before
archive admission.

For the semantic block, we embed the model-generated strategy description
\(d(p)\) using \texttt{text-\allowbreak embedding-\allowbreak 3-\allowbreak small}. The resulting
\(1536\)-dimensional vector is \(L_2\)-normalized:
\[
    e(p)
    =
    \frac{f_{\mathrm{emb}}(d(p))}
         {\max\{\lVert f_{\mathrm{emb}}(d(p))\rVert_2,\varepsilon\}}
    \in \mathbb{R}^{1536}.
\]
The raw embedding is computed once for each distinct description and cached;
refits therefore do not require additional embedding calls.

At refit round \(r\), let \(A_r\) be the current archive and let
\(\bar e_r\) be the mean of its description embeddings. We fit PCA to the
centered embedding matrix and retain the leading eight principal directions,
\(U_r \in \mathbb{R}^{1536\times 8}\). The semantic representation is
\[
    q_r(p) = U_r^\top \bigl(e(p)-\bar e_r\bigr) \in \mathbb{R}^{8}.
\]
The structural and semantic blocks are standardized separately, coordinate
by coordinate, using their respective means and standard deviations over the
current archive. A coordinate with zero empirical variance is mapped to zero.
The final descriptor is the direct concatenation
\[
    z_r(p)
    =
    \left[
        \mathcal{Z}^{\mathrm{ast}}_r\!\left(\phi(p)\right)
        \;\middle\Vert\;
        \mathcal{Z}^{\mathrm{sem}}_r\!\left(q_r(p)\right)
    \right]
    \in \mathbb{R}^{16},
\]
where \(\Vert\) denotes concatenation. We apply no additional block weighting
or normalization after concatenation; separate standardization prevents
either block from dominating the Euclidean geometry used for niche
assignment.
The archive holds at most \(K=50\) incumbents. Rather than partitioning the population with a separate clustering step, we anchor each niche at its own incumbent: niche \(i\) is the Voronoi cell of the site
\(c_i = z_r(p_i) \in \mathbb{R}^{16}\), so the tessellation is determined
entirely by the archive contents under the current descriptor map \(z_r\).

Before the first refit, no descriptor map has been fitted and niche
assignment does not use one: each admitted program simply opens a new niche, in admission order. This is the regime in which the initialization seeds and their lightweight variants enter the archive, so initialization requires neither a PCA basis nor a set of sites. The first refit is triggered as soon as the archive first holds \(K\) incumbents, which, since \(M \cdot V \gg K\),
occurs during the lightweight diversification stage, and establishes
\(z_1\) together with the initial sites \(\{c_i\}_{i=1}^{K}\).

Between refit events, \(z_r\) and the sites \(\{c_i\}\) are held fixed. An incoming candidate \(q\) is projected using the most recent PCA basis and normalization statistics and assigned to \(i^\star = \arg\min_i \lVert z_r(q) - c_i \rVert_2\). It is admitted if and only if \(F(q) > F(p_{i^\star})\), in which case it replaces \(p_{i^\star}\)
but does not displace \(c_{i^\star}\); the site moves only at the next refit. The archive therefore remains saturated at \(K\) occupied niches for the remainder of the run.

After initialization, every \(\Delta_C=50\) completed evaluations we refit PCA on the current archive, refresh the per-coordinate normalization statistics, rebuild the descriptors \(z_r(p_i)\) of all incumbents, and re-anchor every site to its incumbent, \(c_i \leftarrow z_r(p_i)\). Because the sites are the incumbents, a refit changes only the geometry used to route future candidates: it never merges two incumbents into a single niche, except in the degenerate case of two identical descriptors, where the higher-scoring program is retained, and niche index \(i\) continues to denote the same archive slot across refits. This preserves the per-niche statistics the Advisor uses to identify working and saturated niches, and it removes any need to seed the new geometry with the previous round's sites, whose coordinates are in any case no longer comparable once the PCA basis and normalization statistics have been refit.

\subsection{The complete \textsc{SpecEvo} workflow}
\label{app:specevo-main}

Algorithm~\ref{alg:specevo} illustrates the overall workflow of \textsc{SpecEvo}.
Initialization first uses a small number of frontier-model calls to obtain a diverse set of
executable strategy families. Lightweight variants then expand these seeds into a sufficiently
broad population for constructing the behavioral archive. After initialization, \(W\) asynchronous
Speculators perform nearly all candidate generation, while the two consolidation routines are
triggered on independent evaluation schedules. Importantly, Navigator interventions are
non-blocking in our implementation: once a Navigator checkpoint is triggered, the intervention
is dispatched as a background task, while the Speculator workers continue generating and
evaluating candidates. Navigator-generated candidates and their lightweight follow-up variants
are incorporated into the shared search state as they complete. Thus, the \textsc{Navigator} and \textsc{Advisor} calls
in Algorithm~\ref{alg:specevo} are written as single operations for notational compactness and
should not be interpreted as synchronization barriers for the Speculator workers.

\begin{algorithm}[t]
\caption{SpecEvo: asynchronous speculate-then-consolidate search}
\label{alg:specevo}
\small
\begin{algorithmic}[1]
\Require
Task specification $\mathcal{T}$,
evaluator $F$,
budget $\mathcal{B}$,
number of Speculator workers $W$,
Speculator $\mathcal{M}_{\mathrm S}$,
Navigator $\mathcal{M}_{\mathrm N}$,
Advisor $\mathcal{M}_{\mathrm A}$,
number of initialization seeds $M$,
number of variants per seed $V$,
archive capacity $K$,
Navigator interval $\Delta_N$,
Advisor interval $\Delta_A$,
periodic refit interval $\Delta_C$,
Advisor-guidance probability $\rho_A$
\Ensure $\arg\max_{p\in\mathcal{A}_t} F(p)$

\State $t\gets0$ \Comment{completed evaluations}
\State $\mathcal{A}_t,\mathcal{H}_t,\mathcal{S}_0\gets\varnothing$;
       $\ell\gets\varnothing$
\Statex

\Comment{Frontier-seeded initialization}
\For{$i=1,\ldots,M$}
    \State $(p_i^0,d_i^0)\sim
      \mathcal{M}_{\mathrm N}\!
      \left(\cdot\mid\mathcal{T},
      \{(p_j^0,d_j^0,y_j^0)\}_{j<i}\right)$
    \State $(y_i^0,\mathcal{A}_t,\mathcal{H}_t,t)\gets
      \Call{EvaluateUpdate}{
      p_i^0,d_i^0,F,\mathcal{A}_t,\mathcal{H}_t,t}$
    \If{$p_i^0$ is executable}
        \State $\mathcal{S}_0\gets
          \mathcal{S}_0\cup\{(p_i^0,d_i^0,y_i^0)\}$
    \EndIf
\EndFor

\Statex
\Comment{Lightweight diversification}
\ForAll{$(p_i^0,d_i^0,y_i^0)\in\mathcal{S}_0$
        \textbf{ in parallel over up to $W$ workers}}
    \For{$j=1,\ldots,V$}
        \State $I_{i,j}\gets
          \Call{Inspirations}{\mathcal{S}_0,i}$
        \State $(p_{i,j}^0,d_{i,j}^0)\sim
          \mathcal{M}_{\mathrm S}
          (\cdot\mid\mathcal{T},p_i^0,I_{i,j})$
        \State $(\cdot,\mathcal{A}_t,\mathcal{H}_t,t)\gets
          \Call{EvaluateUpdate}{
          p_{i,j}^0,d_{i,j}^0,F,\mathcal{A}_t,\mathcal{H}_t,t}$
    \EndFor
\EndFor

\State $\mathcal{A}_t\gets
  \Call{RefitArchive}{\mathcal{A}_t,K}$
\State $t_N\gets t$; $t_A\gets t$; $t_C\gets t$

\Statex
\Comment{Asynchronous speculate-then-consolidate search}
\While{no stopping condition in $\mathcal{B}$ is met}

    \State $s_t\gets\Call{Stagnation}{\mathcal{H}_t}$

    \ForAll{idle Speculator workers}
        \State $(P_t,I_t,o_t)\gets
          \Call{SpeculationContext}{\mathcal{A}_t,s_t}$
        \State $\widetilde{\ell}\gets\ell$ with probability $\rho_A$;
          otherwise $\widetilde{\ell}\gets\varnothing$
        \State Dispatch $(p,d)\sim
          \mathcal{M}_{\mathrm S}
          (\cdot\mid\mathcal{T},P_t,I_t,o_t,\widetilde{\ell})$
    \EndFor

    \State $(p,d)\gets\Call{AwaitNext}{}$
    \State $(\cdot,\mathcal{A}_t,\mathcal{H}_t,t)\gets
      \Call{EvaluateUpdate}{
      p,d,F,\mathcal{A}_t,\mathcal{H}_t,t}$

    \If{$t-t_N\geq\Delta_N$}
        \State $(\mathcal{A}_t,\mathcal{H}_t,t)\gets
          \Call{Navigator}{
          \mathcal{T},F,\mathcal{A}_t,\mathcal{H}_t,t}$
        \State $t_N\gets t$
    \EndIf

    \If{$t-t_A\geq\Delta_A$}
        \State $\ell\gets
          \Call{Advisor}{
          \mathcal{T},\mathcal{A}_t,\mathcal{H}_t,\ell,t_A,t}$
        \State $t_A\gets t$
    \EndIf

    \If{$t-t_C\geq\Delta_C$}
        \State $\mathcal{A}_t\gets
          \Call{RefitArchive}{\mathcal{A}_t,K}$
        \State $t_C\gets t$
    \EndIf

\EndWhile

\State \Return $\arg\max_{p\in\mathcal{A}_t}F(p)$
\end{algorithmic}
\end{algorithm}

\paragraph{Speculator step.}
Let the archive programs be ordered by decreasing score, with $r_i$ denoting the rank of $p_i$.  At each worker step, the parent distribution and its stagnation-dependent exponent are:
\begin{equation}
 \Pr(p_i\mid\mathcal{A}_t)
 =\frac{r_i^{-\beta_t}}{\sum_{j=1}^{|\mathcal{A}_t|}r_j^{-\beta_t}},
 \qquad
 \beta_t=\beta_{\min}+(1-s_t)(\beta_{\max}-\beta_{\min}).
 \label{eq:app-rank-sampling}
\end{equation}
For crossover, the second parent is drawn from a different niche whenever possible.  Additional inspirations are sampled without replacement from the same rank distribution after excluding the parent(s), but only their descriptions and scores are shown.  This distinction is intentional: parent code supplies a concrete object to modify, whereas description-only inspirations broaden the idea set without encouraging verbatim copying or inflating the prompt.

The six variation prompts are general mutation, focused single-weakness repair, mechanism replacement, analysis-targeted mutation, structural crossover, and component-level crossover. They are sampled uniformly, so each prompt is used with equal probability during Speculator generation.

\paragraph{Asynchrony and checkpoint cadence.}
The evaluation index \(t\) counts completed candidate evaluations rather than optimizer ``generations.'' Because Speculators run asynchronously, multiple evaluations may complete between periodic checks, so a trigger can occur slightly after its nominal interval. After initialization, the Navigator, Advisor, and archive periodic refits maintain independent clocks \(t_N\), \(t_A\), and \(t_C\), all reset to the current \(t\). They are subsequently invoked whenever \(t-t_N\geq\Delta_N\), \(t-t_A\geq\Delta_A\), and \(t-t_C\geq\Delta_C\), respectively. Thus, initialization does not consume any of these intervals, and each periodic operation proceeds independently according to its own evaluation cadence.

\subsection[Navigator intervention]
{Navigator intervention}
\label{app:navigator-details}

\begin{algorithm}[t]
\caption{Navigator consolidation}
\label{alg:navigator}
\small
\begin{algorithmic}[1]
\Require
Task specification $\mathcal{T}$,
evaluator $F$,
archive $\mathcal{A}_t$,
history $\mathcal{H}_t$,
evaluation index $t$,
Navigator $\mathcal{M}_{\mathrm N}$,
Speculator $\mathcal{M}_{\mathrm S}$,
stagnation thresholds $\gamma_1,\gamma_2$,
number of fan-out variants $J$
\Ensure $\mathcal{A}_t,\mathcal{H}_t,t$

\State $(q_1,\ldots,q_{K_t})\gets
    \Call{RankedIncumbents}{\mathcal{A}_t}$
    \Comment{one elite per occupied niche}

\State $s_t\gets\Call{Stagnation}{\mathcal{H}_t}$

\If{$s_t\leq\gamma_1$}
    \State $m_t\gets\textsc{Synthesis}$; $a\gets3$
\ElsIf{$s_t\leq\gamma_2$}
    \State $m_t\gets\textsc{Surgical}$; $a\gets1$
\Else
    \State $m_t\gets\textsc{Reframe}$; $a\gets2$
\EndIf
\State $k\gets5$

\Statex
\Comment{Construct a two-resolution population view}
\State $A_t^{(m)}\gets
    \{(q_i,d(q_i),F(q_i)):
    1\leq i\leq\min(a,K_t)\}$
\State $I_t^{(m)}\gets
    \{(d(q_i),F(q_i)):
    a<i\leq\min(a+k,K_t)\}$
\State $V_t^{(m)}\gets
    \Call{PopulationView}{
    A_t^{(m)},I_t^{(m)},s_t,\mathcal{H}_t}$

\Statex
\Comment{Frontier-model intervention}
\State $(p_N,d_N)\sim
    \mathcal{M}_{\mathrm N}
    (\cdot\mid\mathcal{T},V_t^{(m)},m_t)$
\State $y_N\gets F(p_N)$; $t\gets t+1$

\If{$y_N$ is invalid}
    \State $\mathcal{H}_t\gets
        \Call{RecordFailure}{\mathcal{H}_t,p_N,d_N}$
    \State \Return $(\mathcal{A}_t,\mathcal{H}_t,t)$
\EndIf

\State $(\mathcal{A}_t,\mathcal{H}_t)\gets
        \Call{Update}{
        \mathcal{A}_t,\mathcal{H}_t,p_N,d_N,y_N}$

\Statex
\Comment{Lightweight refinement of the Navigator proposal}
\For{$j=1,\ldots,J$ \textbf{ in parallel}}
    \State $(p_{N,j},d_{N,j})\sim
        \mathcal{M}_{\mathrm S}
        (\cdot\mid\mathcal{T},p_N,d_N)$
    \State $(\cdot,\mathcal{A}_t,\mathcal{H}_t,t)\gets
        \Call{EvaluateUpdate}{
        p_{N,j},d_{N,j},F,\mathcal{A}_t,\mathcal{H}_t,t}$
\EndFor

\State \Return $(\mathcal{A}_t,\mathcal{H}_t,t)$
\end{algorithmic}
\end{algorithm}
Algorithm~\ref{alg:navigator} illustrates the overall workflow of a Navigator invocation.

\paragraph{Selecting anchors and inspirations.}
The Navigator receives a two-resolution view of the behavioral archive.
We first collect the incumbent from each occupied niche and rank these niche
elites by score. The highest-ranked elites are selected as
\emph{anchors}, for which the Navigator observes the complete program,
strategy description, and score. The subsequent elites are used as
\emph{inspirations}, exposing only their descriptions and scores.
Anchors and inspirations are disjoint within each invocation.

Full-code anchors support precise reasoning and modification, whereas description-only inspirations expose a broader set of mechanisms without overwhelming the prompt; ranking niche elites rather than the entire population keeps this context behaviorally diverse.

Table~\ref{tab:navigator-context} summarizes the mode-specific context: \textsc{Synthesis} needs several anchors' implementations to combine complementary mechanisms, \textsc{Surgical} needs only the champion it repairs, and \textsc{Reframe} uses two anchors to identify shared assumptions while relying on descriptions to seek a qualitatively different strategy family.
\newcolumntype{Y}{>{\raggedright\arraybackslash}X}
\begin{table}[t]
\centering
\small
\setlength{\tabcolsep}{6pt}
\renewcommand{\arraystretch}{1.15}
\rowcolors{2}{gray!8}{white}
\begin{tabularx}{\linewidth}{@{}lYcY@{}}
\toprule
\rowcolor{blue!12}
\textbf{Mode} &
\textbf{Full-code anchors} &
\textbf{Description-only inspirations} &
\textbf{Requested move} \\
\midrule
\textsc{Synthesis} &
top 3 niche elites &
next 5 niche elites &
combine mechanisms across anchors \\
\textsc{Surgical} &
global champion only &
next 5 niche elites &
one local structural correction \\
\textsc{Reframe} &
top 2 niche elites &
next 5 niche elites &
introduce an absent strategy family \\
\bottomrule
\end{tabularx}
\caption{Mode-specific context provided to the Navigator.}
\label{tab:navigator-context}
\end{table}

\subsection{Advisor intervention}
\label{app:advisor-details}

Algorithm~\ref{alg:advisor} summarizes the overall workflow of an Advisor intervention, from extracting niche-level search signals and recurrent failures to distilling guidance for subsequent Speculator generations.

\paragraph{Tracking niche activity and progress.}
At each Advisor checkpoint, we inspect the evaluations completed since the
previous checkpoint. For every occupied niche \(k\), we maintain two signals:
an \emph{activity count} \(a_k\), equal to the number of evaluated candidates
assigned to that niche during the interval, and a \emph{movement indicator}
\(m_k\), which records whether the niche incumbent was replaced during the
same interval. Importantly, the activity count includes unsuccessful
candidates. Hence, a niche can receive substantial search effort without
producing a better incumbent.

These signals distinguish productive search from saturation without requiring
a task-specific improvement threshold. A niche with recent incumbent movement
is treated as actively improving, whereas a niche with high activity but no
movement is considered saturated. This distinction is particularly useful
across our heterogeneous benchmarks, where the scale of score improvements
varies substantially between tasks.

At each checkpoint, the Advisor constructs two complementary population views:
\begin{itemize}
    \item \(D_t^{\mathrm{work}}\): up to four highest-scoring niches whose
    incumbents moved during the current interval;
    \item \(D_t^{\mathrm{sat}}\): up to three most active niches that received
    evaluations but showed no incumbent movement.
\end{itemize}
Together, these views separate current solution quality from recent search
progress and exploration coverage. In particular, a high-scoring niche need
not be improving, while a lower-scoring niche may reveal a productive direction
that has only recently emerged.








\begin{algorithm}[t]
\caption{Advisor checkpoint}
\label{alg:advisor}
\small
\begin{algorithmic}[1]
\Require
Task specification \(\mathcal{T}\),
archive \(\mathcal{A}_t\),
history \(\mathcal{H}_t\),
evaluation index \(t\),
previous Advisor checkpoint \(t_A\),
previous advice \(\ell\),
Advisor \(\mathcal{M}_{\mathrm A}\)
\Ensure Updated advice \(\ell\)

\State \((a_k,m_k)_k\gets
    \Call{RegionSignals}{\mathcal{A}_t,\mathcal{H}_t,t_A,t}\)
    \Comment{activity and movement}

\State \(D_t^{\mathrm{work}}\gets
    \Call{TopScore}{\{\mathcal{A}_t(k):m_k=1\},4}\)
\State \(D_t^{\mathrm{sat}}\gets
    \Call{TopActivity}{\{k:m_k=0\land a_k>0\},3}\)

\Statex
\State \(\mathcal{R}_t\gets
    \Call{AggregateFailures}{\mathcal{H}_t}\)
    \Comment{cumulative counts; no eviction}
\State \(E_t\gets
    \Call{MostFrequent}{\mathcal{R}_t,8}\)
    \Comment{break ties by recency}

\Statex
\State \(X_t\gets
    (D_t^{\mathrm{work}},
     D_t^{\mathrm{sat}},
     E_t,\ell,y_t^{\mathrm{best}},s_t)\)

\State \(\ell\sim
    \mathcal{M}_{\mathrm A}(\cdot\mid\mathcal{T},X_t)\)
    \Comment{distill structured guidance}

\State \Return \(\ell\)
\end{algorithmic}
\end{algorithm}

\textbf{Failure memory.}
We aggregate evaluator failures from the complete search history $\mathcal{H}_t$, including initialization, using normalized error signatures. For each signature, $\mathcal{R}_t$ records its cumulative
occurrence count, most recent occurrence, and a representative message. No signatures are evicted during a run. At each Advisor checkpoint, we select the eight most frequent signatures, breaking ties by recency. This bounds the failure evidence included in the prompt while retaining earlier failures in the underlying history. The Advisor combines this evidence with recent niche activity and its previous guidance to revise the next note.

\paragraph{Advisor output and guidance injection.}
The Advisor converts the population views, recurrent failures, and previous
advice into a compact note with three fields. \textsc{Working} summarizes
mechanisms associated with leading or recently improving niches;
\textsc{Saturated} identifies directions that continue to consume evaluations
without moving their incumbents; and \textsc{Avoid} highlights recurrent
anti-patterns supported by the failure memory.

We use temperature \(0.4\) and a 1000-token generation limit, with the resulting
note constrained to fewer than 500 words. The note is not imposed on every
worker. Instead, each subsequent Speculator receives it independently with
probability \(\rho_A=0.35\). This stochastic injection allows useful
population-level lessons to propagate through the search while preserving a
substantial fraction of independent exploration.

\section{Metric Details}
\label{app:metric-details}

\subsection{Aggregate Metrics Used in Figure~\ref{fig:rq1}}
\label{app:fig3-metrics}

Figure~\ref{fig:rq1} summarizes performance and API expenditure across tasks whose raw objective scales differ substantially, so we normalize both quantities within each benchmark family $f$: the seven mathematical discovery tasks and the four systems optimization tasks (all evaluators are higher-is-better). Let $\mathcal{T}_f$ be the tasks in $f$, $\mathcal{C}_f$ the method--backbone configurations shown in the corresponding panel, and $\bar s_{c,t}$ and $\bar u_{c,t}$ the final best score and the API cost (USD) of configuration $c$ on task $t$, each averaged over the three runs. We report
\[
\operatorname{Perf}(c;f)=\frac{100}{|\mathcal{T}_f|}\sum_{t\in\mathcal{T}_f}\frac{\bar s_{c,t}}{\max_{c'\in\mathcal{C}_f}\bar s_{c',t}},
\qquad
\operatorname{Cost}(c;f)=100\,\frac{\bar u_{c;f}}{\max_{c'\in\mathcal{C}_f}\bar u_{c';f}},
\]
where $\bar u_{c;f}$ is the mean of $\bar u_{c,t}$ over $t\in\mathcal{T}_f$. Every task thus contributes equally regardless of its native scale, and $100\%$ performance corresponds to matching the strongest displayed mean on every task. Higher normalized performance and lower normalized cost are preferable; the \emph{Best} scores in the detailed tables are complementary statistics that are not used in this aggregate.

\subsection{Code Diversity Index (CDI)}
\label{app:cdi}

CDI~\citep{dat2025hsevo} measures how evenly the population's code representations are separated along a minimum spanning tree (MST).  At checkpoint $t$, let
$\mathcal{Q}_t=\{p_1,\ldots,p_n\}$ be the complete archive snapshot.  We first
canonicalize each source string.  For each Python program, we parse its
abstract syntax tree, remove comments and module/function/class docstrings,
and emit a deterministically formatted program.  If parsing fails, we fall
back to whitespace-normalized source.  Let
\begin{equation}
 x_i=f_{\mathrm{emb}}\!\left(\operatorname{canon}(\operatorname{code}(p_i))\right)
 \in\mathbb{R}^{d}
 \label{eq:app-cdi-embedding}
\end{equation}
be the embedding of the resulting code.  We use
\texttt{text-embedding-3-small} for all methods and checkpoints.

Construct the complete undirected graph on the $n$ embeddings with Euclidean
edge weight
\begin{equation}
 d_{ij}=\lVert x_i-x_j\rVert_2.
 \label{eq:app-cdi-distance}
\end{equation}
Let $\mathcal{E}_t^{\mathrm{MST}}$ be the edge set of a minimum spanning tree,
and let $\mathcal{E}_t^+=\{e\in\mathcal{E}_t^{\mathrm{MST}}:d_e>0\}$ remove
zero-length duplicate edges.  Normalize its positive edge lengths as
\begin{equation}
 \pi_e=\frac{d_e}{\sum_{e'\in\mathcal{E}_t^+}d_{e'}}.
 \label{eq:app-cdi-mass}
\end{equation}
The reported Code Diversity Index is the Shannon entropy
\begin{equation}
 \operatorname{CDI}(\mathcal{Q}_t)
 =-\sum_{e\in\mathcal{E}_t^+}\pi_e\log \pi_e,
 \label{eq:app-cdi}
\end{equation}
where $\log$ is the natural logarithm.  If there is no positive MST edge
(e.g., $n\leq1$ or all embeddings coincide), CDI is defined as zero.

CDI is high when separation is distributed across many MST edges instead of
being dominated by one bridge between otherwise collapsed groups.  Since the
edge lengths are normalized, uniform rescaling of every embedding distance
does not change CDI; the metric captures the distribution of population
separations, not an absolute diameter.  With $m=|\mathcal{E}_t^+|\geq1$
positive edges,
$0\leq\operatorname{CDI}\leq\log m\leq\log(n-1)$, with the upper bound
attained when the positive MST edges have equal length.  Because this upper
bound depends on population size, comparisons should use checkpoints with
comparable archive sizes.  Our computation keeps duplicate programs by
default and uses the archive exactly as stored at the checkpoint.

\subsection{Stagnation rate}
\label{app:stagnation-rate}

The reported \emph{stagnation rate} is a post-hoc, run-level statistic
that asks what fraction of the full evaluation budget was consumed by the
single longest global-best plateau.

For a run of $T$ evaluations, define the new-best index set
\begin{equation}
 \mathcal{I}_{\mathrm{best}}=\left\{t\in\{1,\ldots,T\}:y_t>
 \max_{1\leq j<t}y_j\right\}
 =\{b_1<\cdots<b_m\}.
 \label{eq:app-new-best-set}
\end{equation}
Set the sentinels $b_0=0$ and $b_{m+1}=T+1$.  The number of consecutive non-new-best evaluations between two adjacent events is
$g_i=b_{i+1}-b_i-1$.  This convention includes the prefix before the first new
best and the suffix after the last new best.  The run-level metric is
\begin{equation}
 \operatorname{StagRate}
 =100\times\frac{\max_{0\leq i\leq m}g_i}{T}\quad(\%).
 \label{eq:app-stagnation-rate}
\end{equation}

\subsection{New bests traced to the frontier model}
\label{app:frontier-new-bests}

This metric measures how often observed progress can be attributed to the frontier model.
In \textsc{LEVI}, the frontier model is invoked through \texttt{paradigm\_shift},
while in \textsc{SpecEvo} it serves as the Navigator.
For each new-best event \(b_i\) in Eq.~\ref{eq:app-new-best-set}, define
\begin{equation}
z(b_i)=
\begin{cases}
1, & \text{if \(b_i\) is generated by the frontier model, or by a small model} \\
   & \text{with at least one frontier-generated parent}, \\
0, & \text{otherwise.}
\end{cases}
\label{eq:app-frontier-indicator}
\end{equation}

The run-level rate is
\begin{equation}
\operatorname{FrontierTrace}
=100\times\frac{\sum_{i=1}^{m} z(b_i)}{m}\quad(\%),
\label{eq:app-frontier-trace}
\end{equation}
where \(m\) is the number of new-best events.
Thus, \(\operatorname{FrontierTrace}\) measures the fraction of improvements
that are either directly produced by the frontier model or arise from
lightweight refinement of its programs.
Initialization is excluded, and the metric is reported as missing when \(m=0\).
Since it counts improvement events rather than their magnitude, it should be
interpreted together with final performance and API cost.

\section{Benchmark Details}
\label{app:benchmark-details}

We provide additional details about the benchmarks used in our evaluation. Unless otherwise stated, a candidate solution is represented as an executable program and evaluated by a task-specific evaluator. The evolutionary methods only observe the resulting scalar score and diagnostic feedback, without access to hidden test instances or ground-truth solutions.

\subsection{Mathematical Optimization Benchmarks}
\label{app:math-benchmarks}

We consider seven executable mathematical optimization tasks covering geometric packing, extremal point configurations, and signal processing. The geometric tasks follow the formulations used in AlphaEvolve~\citep{novikov2025alphaevolve}. Each candidate program constructs or refines a feasible solution, and the evaluator returns a scalar
higher-is-better objective value. The exact configurations used in our experiments are summarized in Table~\ref{tab:math-benchmark-details}.

\subsection{Systems Optimization Benchmarks}
\label{app:system-benchmarks}

We use four real-world systems optimization tasks from the AI-Driven Research for Systems benchmark suite~\citep{cheng2025barbarians}. Although some underlying objectives are naturally formulated as
minimization problems, the task evaluators convert them into scalar scores for which a larger value indicates a better candidate in all reported experiments. Table~\ref{tab:system-benchmark-details} summarizes the four tasks included in our evaluation.

\subsection{CO-Bench}
\label{app:cobench}

CO-Bench~\citep{sun2026co} evaluates the ability of language-model agents to design complete algorithms for combinatorial optimization rather than directly predicting solutions to individual instances. It contains 36 real-world optimization problems spanning eight categories and a broad range of problem sizes and constraint structures. Most instances are curated from OR-Library and related operations-research collections.

For each problem, the benchmark provides the agent with a natural language problem specification, a starter \texttt{solve} interface, and a development set. A hidden evaluation function checks feasibility and measures solution quality. The test instances and evaluator implementation are unavailable during search, reducing direct overfitting to the final evaluation data.

Let \(h(x;p)\) denote the raw objective value of a feasible solution
\(x\) on instance \(p\), and let \(h_p^{*}\) denote the reference value
recorded in the benchmark. We use the task-specific normalization
implemented in CO-Bench, denoted by \(\mathcal{N}_p\), to obtain a
higher-is-better score:
\begin{equation}
s(x,p)
=
\begin{cases}
\mathcal{N}_p\bigl(h(x;p),h_p^{*}\bigr),
& \text{if execution succeeds and } x \text{ is feasible},\\
0,
& \text{if execution fails, times out, or } x \text{ is infeasible}.
\end{cases}
\label{eq:cobench-score}
\end{equation}
For standard positive-reference ratio normalization,
\(\mathcal{N}_p(h,h_p^{*})=h/h_p^{*}\) for maximization and
\(\mathcal{N}_p(h,h_p^{*})=h_p^{*}/h\) for minimization, with nonzero
denominators. We retain the benchmark's special-case handling,
including shifted ratios for zero-reference instances.

The normalized score is always interpreted as higher-is-better,
regardless of the original optimization direction. Failed evaluations
receive a final normalized score of \(0\); this value is not substituted
into the raw objective before normalization. A score of \(1\) corresponds
to matching the reference, while scores above \(1\) can occur when a
feasible solution improves upon a recorded best-known reference under
the benchmark's uncapped ratio normalization.

The score for each problem is averaged over its test instances, and
the overall CO-Bench score is the unweighted mean across all 36 problems:
\begin{equation}
S(A)
=
\frac{1}{36}
\sum_{c=1}^{36}
\frac{1}{|\mathcal{P}_c|}
\sum_{p\in\mathcal{P}_c}
s\bigl(A(p),p\bigr),
\label{eq:cobench-overall}
\end{equation}
where \(A\) is a candidate algorithm and \(\mathcal{P}_c\) is the test set of problem \(c\). Due to the three-hour wall-clock budget, we sample at most 50 instances for each problem and split them into training and test sets using a 7:3 ratio, enabling a comprehensive comparison between \textsc{SpecEvo} and the baselines under a consistent computational budget.

Table~\ref{tab:cobench-categories} lists all problem families included in our evaluation.

\begin{table}[t]
\centering
\small
\setlength{\tabcolsep}{5pt}
\renewcommand{\arraystretch}{1.10}
\caption{Mathematical optimization benchmarks used in our evaluation.}
\label{tab:math-benchmark-details}
\begin{tabularx}{\textwidth}{
    @{}
    >{\raggedright\arraybackslash}p{0.235\textwidth}
    >{\centering\arraybackslash}p{0.15\textwidth}
    >{\raggedright\arraybackslash}X
    @{}}
\toprule
\textbf{Problem}
& \textbf{Configuration}
& \textbf{Objective and task definition} \\
\midrule

Circle Packing (Square)
& $N=26$
& Pack \(N\) pairwise-disjoint circles inside a unit square and
maximize the sum of their radii. \\

\addlinespace

Circle Packing (Rectangle)
& $N=21$
& Pack \(N\) pairwise-disjoint circles inside a rectangle with
perimeter \(4\) and maximize the sum of their radii. \\

\addlinespace

Heilbronn Problem (Triangle)
& $N=11$
& Place \(N\) points on or inside a triangle of unit area so as to
maximize the minimum area of any triangle formed by three selected
points. \\

\addlinespace

Heilbronn Problem (Convex Region)
& $N=13$
& Place \(N\) points on or inside a convex region of unit area so as
to maximize the minimum area of any triangle formed by three selected
points. \\

\addlinespace

MinMax Distance
& $(N,d)=(16,2)$
& Find \(N\) points in \(\mathbb{R}^{d}\) that maximize the ratio
between the minimum and maximum pairwise Euclidean distances. \\

\addlinespace

MinMax Distance
& $(N,d)=(14,3)$
& Optimize the same MinMax-distance objective for \(14\) points in
three-dimensional Euclidean space. \\

\addlinespace

Signal Processing
& 
& Optimize a continuous signal processing objective under noisy evaluation, where candidate programs transform an input signal and are scored by an external evaluator. \\

\bottomrule
\end{tabularx}
\end{table}

\begin{table}[t]
\centering
\small
\setlength{\tabcolsep}{5pt}
\renewcommand{\arraystretch}{1.10}
\caption{Systems optimization benchmarks used in our evaluation.}
\label{tab:system-benchmark-details}
\begin{tabularx}{\textwidth}{
    @{}
    >{\raggedright\arraybackslash}p{0.235\textwidth}
    >{\raggedright\arraybackslash}p{0.245\textwidth}
    >{\raggedright\arraybackslash}X
    @{}}
\toprule
\textbf{Task}
& \textbf{Objective}
& \textbf{Problem description} \\
\midrule

Expert Parallelism Load Balancer (EPLB)
& Balance expert-parallel workloads across GPUs
& Given expert load statistics for a mixture-of-experts model,
determine the number and placement of expert replicas across physical
GPU slots. The objective is to reduce computational skew and
stragglers while satisfying placement constraints and preserving
inference throughput. \\

\addlinespace

Model Placement (PRISM)
& Optimize model-to-GPU placement cost
& Assign multiple models or model components to a collection of GPUs
under memory and capacity constraints. The placement should minimize
global execution cost or peak resource pressure while maintaining
feasibility and service-level requirements. \\

\addlinespace

LLM-SQL
& Maximize prefix-cache efficiency
& Reorder table rows, columns, or fields so that serialized SQL
inputs share longer prefixes and obtain higher prefix-cache reuse.
The transformation must preserve the original table semantics and
query results. \\

\addlinespace

Transaction Scheduling (TXN)
& Minimize transaction-execution makespan
& Determine an execution order for dependent or conflicting database
transactions. The scheduler seeks to reduce total completion time
while respecting dependencies and avoiding excessive conflicts,
aborts, or waiting time. \\

\bottomrule
\end{tabularx}
\end{table}

\begin{table*}[t]
\centering
\small
\setlength{\tabcolsep}{5pt}
\renewcommand{\arraystretch}{1.12}
\caption{The 36 combinatorial optimization problems in CO-Bench, grouped into eight categories.}
\label{tab:cobench-categories}
\begin{tabularx}{\textwidth}{
    @{}
    >{\raggedright\arraybackslash}p{0.145\textwidth}
    >{\centering\arraybackslash}p{0.045\textwidth}
    >{\raggedright\arraybackslash}X
    @{}}
\toprule
\textbf{Category}
& \textbf{\#}
& \textbf{Included problems} \\
\midrule

Packing 
& 9
& Bin packing; Multi-Demand Multidimensional Knapsack problem;
Multidimensional knapsack problem; Container loading; Container loading
with weight restrictions; Packing unequal circles; Packing unequal
rectangles and squares number / area. \\

\addlinespace

Cutting
& 4
& Assortment problem; Constrained / unconstrained guillotine cutting;
Constrained non-guillotine cutting. \\

\addlinespace

Facility location
& 4
& Capacitated / Uncapacitated warehouse location; Capacitated /
Uncapacitated p-median problem. \\

\addlinespace

Scheduling
& 7
& Aircraft landing; Crew scheduling; Common due date scheduling; Flow
shop scheduling; Hybrid Reentrant Shop Scheduling; Job shop scheduling;
Open shop scheduling. \\

\addlinespace

Routing
& 3
& Traveling salesman problem; Period vehicle routing problem; Resource
constrained shortest path. \\

\addlinespace

Assignment
& 2
& Constrained / unconstrained assignment. \\

\addlinespace

Tree
& 2
& Euclidean Steiner; Corporate structuring. \\

\addlinespace

Graph and set
& 5
& Maximal Independent Set; Graph coloring; Equitable partitioning; Set
partitioning; Set covering. \\

\midrule
\textbf{Total}
& \textbf{36}
& \\
\bottomrule
\end{tabularx}
\end{table*}

\subsection{LSR-Synth}
\label{app:llmsrbench}

We use LSR-Synth from LLM-SRBench~\citep{shojaee2025llm}, a benchmark designed to
evaluate LLM-based scientific equation discovery. Unlike traditional symbolic regression benchmarks that often contain well-established canonical equations~\citep{la2021contemporary, matsubara2022rethinking}, LSR-Synth combines known functional terms with novel synthetic terms. This construction reduces the likelihood that models can solve the tasks through memorization or direct recall of equations encountered during pre-training.

Each problem provides a scientific context, descriptions of the relevant variables, and an observational dataset. The objective is to discover a mathematical expression that accurately captures the underlying relationship between the input variables and the target quantity. LSR-Synth covers four scientific domains: chemistry, biology,
physics, and material science.

We use all 129 problems from LSR-Synth, comprising 36 chemistry problems, 24 biology problems, 44 physics problems, and 25 material science problems. 

During the search process, candidate equations are ranked using
\[
\mathrm{Score}
=
\log_{10}\left(\frac{1}{1+\mathrm{NMSE}_{\mathrm{train}}}\right),
\]
where \(\mathrm{NMSE}_{\mathrm{train}}\) is computed on the training data. For the final evaluation, we directly report NMSE on both the in-distribution (ID) test set, whose variable ranges follow those seen during training, and the out-of-distribution (OOD) test set, which
contains values outside the training ranges to assess extrapolation ability.

Detailed statistics are reported in
Table~\ref{tab:lsr-synth-statistics}.

\begin{table}[t]
\centering
\small
\setlength{\tabcolsep}{5pt}
\renewcommand{\arraystretch}{1.10}
\caption{Statistics of the LSR-Synth problems used in our evaluation.}
\label{tab:lsr-synth-statistics}
\resizebox{0.7\columnwidth}{!}{%
\begin{tabular}{@{}lccccc@{}}
\toprule
&
\textbf{Chemistry}
&
\textbf{Biology}
&
\textbf{Physics}
&
\textbf{Material}
&
\textbf{Overall}
\\
\midrule

\textbf{\# Problems}
& 36
& 24
& 44
& 25
& 129
\\

\textbf{Avg. Datapoints (k)}
& 5.0
& 5.0
& 5.0
& 5.0
& 5.0
\\

\textbf{Avg. Variants}
& 3.0
& 3.0
& 3.9
& 3.0
& 3.3
\\

\bottomrule
\end{tabular}%
}
\end{table}

\section{Additional Experimental Results}
\label{app:add_result}

\subsection{Breakdown of API Cost and Token Usage}
\label{app:breakdown_cost}

\begin{figure}
    \centering
\includegraphics[width=1\linewidth]{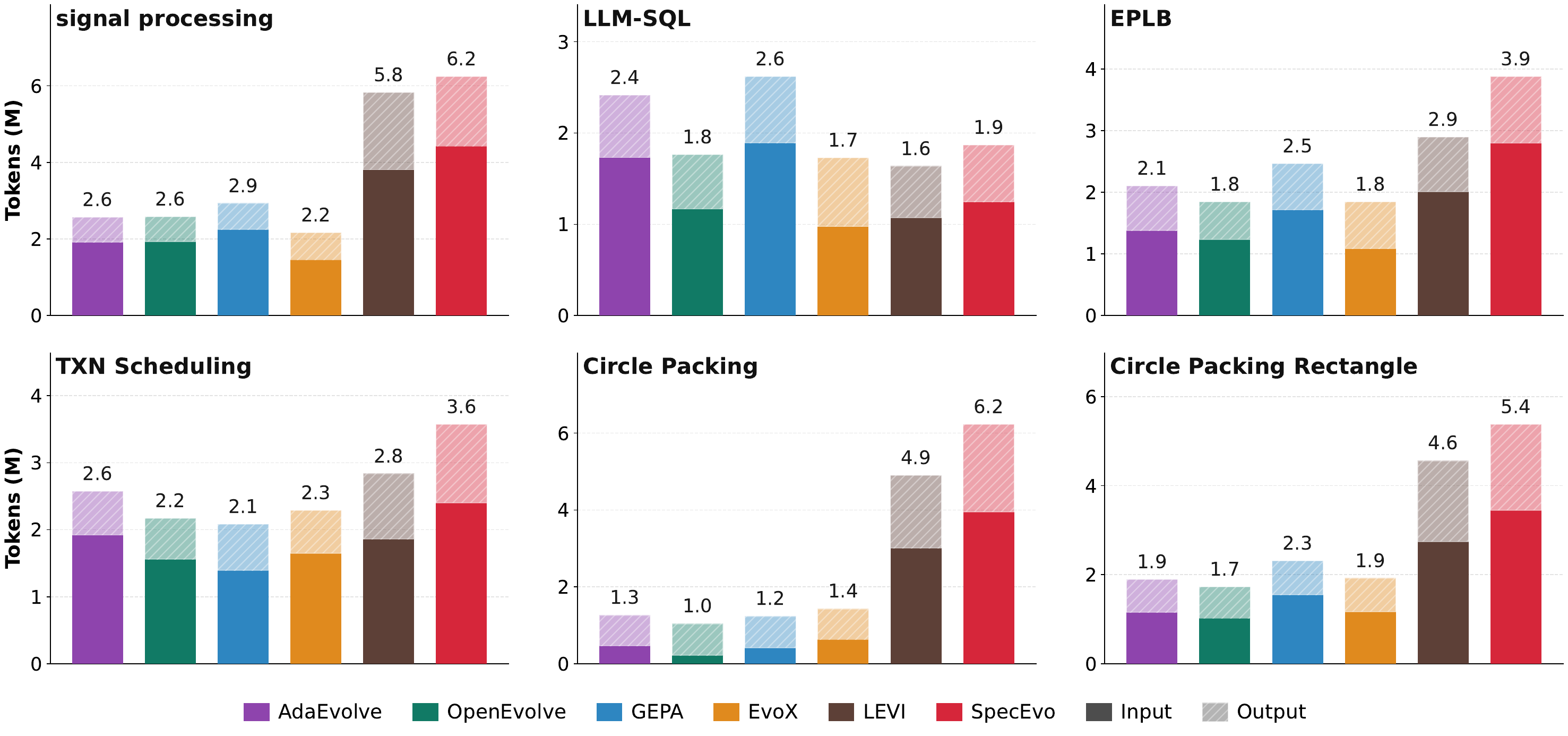}
\caption{Input and output token usage across six representative tasks using GPT-5 as the frontier-model backbone. Total token consumption is reported in millions (M) for each method.}
\label{fig:token-usage}
\end{figure}

Table~\ref{tab:openrouter_model_pricing} reports the OpenRouter list prices of all models used in our experiments, and Figure~\ref{fig:token-usage} shows the corresponding input and output token usage on six representative tasks with GPT-5 as the frontier-model backbone. \textsc{SpecEvo} often uses more tokens in total than several baselines, and in some cases more than \textsc{LEVI}, because it hands a large volume of routine search, mutation, and refinement to lightweight Speculators. Higher token usage does not mean higher cost: most of the additional tokens come from inexpensive small models, and costly frontier-model calls are reserved for infrequent Navigator interventions. With GPT-5 output tokens at \$10.00 per million versus \$0.30 for Qwen3-30B-A3B-Instruct-2507, moving most of the search workload to lightweight models yields substantial savings even when the total token count is larger.

Tables~\ref{tab:math_api_cost} and~\ref{tab:system_api_cost} provide the task-level API expenditure under the fixed three-hour wall-clock budget. Excluding \textsc{LEVI}, \textsc{SpecEvo} reduces API cost by approximately \(49.9\%\)--\(70.7\%\) across the four benchmark--backbone settings. We treat \textsc{LEVI} separately because it also mixes models: it uses a lightweight model for routine search and calls the frontier model only for occasional \texttt{paradigm\_shift} operations, so it already benefits from the same price gap between lightweight and frontier models. The comparison with \textsc{LEVI} therefore more directly reflects how each method allocates scarce frontier-model reasoning.

\begin{table*}[t]
\centering
\small
\setlength{\tabcolsep}{10pt}
\renewcommand{\arraystretch}{1.15}
\caption{
OpenRouter API pricing for the models used in our experiments,
reported in USD per one million tokens (accessed August 1, 2026).
Kimi-K2 in the main text, figures, and tables refers to Kimi-K2-Thinking.
}
\label{tab:openrouter_model_pricing}

\begin{tabularx}{0.76\textwidth}{
    @{}
    >{\raggedright\arraybackslash}X
    S[table-format=2.5]
    S[table-format=2.5]
    @{}
}
\toprule
\textbf{Model}
& \multicolumn{2}{c}{\textbf{Price per 1M Tokens (USD)}} \\
\cmidrule(lr){2-3}
& \multicolumn{1}{c}{\textbf{Input}}
& \multicolumn{1}{c}{\textbf{Output}} \\
\midrule

\multicolumn{3}{@{}l}{\textbf{Frontier Models}} \\
GPT-5
& 1.25
& 10.00 \\

Kimi-K2-Thinking
& 0.60 
& 2.50 \\
\midrule
\addlinespace[4pt]
\multicolumn{3}{@{}l}{\textbf{Small Models}} \\
Qwen3-30B-A3B-Instruct-2507
& 0.10
& 0.30 \\

Qwen3-14B
& 0.10
& 0.22 \\

Qwen3-8B
& 0.11
& 0.45 \\

Qwen2.5-7B-Instruct
& 0.10
& 0.20 \\
\midrule
\addlinespace[4pt]
\multicolumn{3}{@{}l}{\textbf{Embedding Model}} \\
Text Embedding 3 Small
& 0.02
&  \\
\bottomrule

\end{tabularx}
\end{table*}

\begin{table*}[t]
\centering
\footnotesize
\setlength{\tabcolsep}{3.5pt}
\renewcommand{\arraystretch}{1.15}
\caption{
API cost in USD under the fixed three-hour wall-clock budget on
mathematical discovery tasks. Each entry reports the mean and standard
deviation over three independent runs. Lower is better.
}
\label{tab:math_api_cost}
\begin{adjustbox}{max width=\textwidth}
\begin{tabular}{@{}lccccccc@{}}
\toprule
\textbf{Strategy}
& \textbf{Circle Packing}
& \textbf{Circle Rect.}
& \textbf{Heil. Conv.}
& \textbf{Heil. Tri.}
& \textbf{MinMax 16-2}
& \textbf{MinMax 14-3}
& \textbf{Signal} \\
\midrule

\multicolumn{8}{@{}l}{\textit{Backbone: GPT-5}} \\

OpenEvolve
& 7.65 $\pm$ 0.40
& 8.65 $\pm$ 1.01
& 8.76 $\pm$ 2.48
& 8.53 $\pm$ 0.02
& 8.31 $\pm$ 0.28
& 7.12 $\pm$ 0.86
& 8.53 $\pm$ 0.64 \\

GEPA
& 8.41 $\pm$ 0.10
& 5.20 $\pm$ 1.14
& 8.37 $\pm$ 1.04
& 9.36 $\pm$ 1.71
& 7.82 $\pm$ 1.41
& 8.09 $\pm$ 0.55
& 9.74 $\pm$ 1.75 \\

AdaEvolve
& 8.83 $\pm$ 0.31
& 6.68 $\pm$ 1.29
& 8.62 $\pm$ 0.06
& 8.93 $\pm$ 0.86
& 7.96 $\pm$ 0.26
& 8.58 $\pm$ 0.01
& 7.91 $\pm$ 0.91 \\

EvoX
& 7.66 $\pm$ 0.50
& 7.33 $\pm$ 0.68
& 9.18 $\pm$ 0.22
& 9.12 $\pm$ 1.58
& 8.15 $\pm$ 0.56
& 7.17 $\pm$ 0.67
& 8.72 $\pm$ 2.05 \\

LEVI
& 3.40 $\pm$ 1.23
& 2.62 $\pm$ 1.05
& \bestcell{1.60} $\pm$ 0.25
& 1.33 $\pm$ 0.19
& 1.80 $\pm$ 0.17
& 2.45 $\pm$ 0.75
& \bestcell{3.64} $\pm$ 0.12 \\

SpecEvo
& \bestcell{3.37} $\pm$ 1.61
& \bestcell{2.61} $\pm$ 1.29
& 1.62 $\pm$ 0.35
& \bestcell{1.32} $\pm$ 0.18
& \bestcell{1.65} $\pm$ 0.06
& \bestcell{2.33} $\pm$ 0.86
& 3.88 $\pm$ 0.01 \\

\midrule

\multicolumn{8}{@{}l}{\textit{Backbone: Kimi-K2}} \\

OpenEvolve
& 1.98 $\pm$ 0.20
& 1.63 $\pm$ 0.19
& 1.64 $\pm$ 0.16
& 1.63 $\pm$ 0.36
& 1.61 $\pm$ 0.21
& 1.52 $\pm$ 0.19
& 2.28 $\pm$ 0.28 \\

GEPA
& 1.95 $\pm$ 0.10
& 1.69 $\pm$ 0.05
& 1.45 $\pm$ 0.16
& 1.45 $\pm$ 0.11
& 1.62 $\pm$ 0.04
& 1.49 $\pm$ 0.23
& 2.18 $\pm$ 0.26 \\

AdaEvolve
& 1.67 $\pm$ 0.16
& 1.79 $\pm$ 0.31
& 1.59 $\pm$ 0.12
& 1.49 $\pm$ 0.24
& 1.63 $\pm$ 0.31
& 1.52 $\pm$ 0.09
& 2.23 $\pm$ 0.06 \\

EvoX
& 1.76 $\pm$ 0.15
& 1.67 $\pm$ 0.22
& 1.62 $\pm$ 0.16
& 1.75 $\pm$ 0.22
& 1.55 $\pm$ 0.14
& 1.66 $\pm$ 0.39
& 2.33 $\pm$ 0.05 \\

LEVI
& \bestcell{0.79} $\pm$ 0.12
& \bestcell{0.77} $\pm$ 0.21
& 0.45 $\pm$ 0.15
& 0.65 $\pm$ 0.30
& 1.12 $\pm$ 0.49
& 1.04 $\pm$ 0.18
& \bestcell{1.39} $\pm$ 0.20 \\

SpecEvo
& 0.83 $\pm$ 0.33
& 0.83 $\pm$ 0.22
& \bestcell{0.42} $\pm$ 0.21
& \bestcell{0.50} $\pm$ 0.27
& \bestcell{1.05} $\pm$ 0.27
& \bestcell{1.02} $\pm$ 0.26
& 1.41 $\pm$ 0.14 \\

\bottomrule
\end{tabular}
\end{adjustbox}
\end{table*}

\begin{table*}[t]
\centering
\scriptsize
\setlength{\tabcolsep}{7pt}
\renewcommand{\arraystretch}{1.12}
\caption{
API cost in USD under the fixed three-hour wall-clock budget on
systems optimization tasks. Each entry reports the mean and standard
deviation over three independent runs. Lower is better.
}
\label{tab:system_api_cost}
\begin{tabular}{@{}lcccc@{}}
\toprule
\textbf{Strategy}
& \textbf{EPLB $(\uparrow)$}
& \textbf{LLM-SQL $(\uparrow)$}
& \textbf{Transaction $(\uparrow)$}
& \textbf{PRISM $(\uparrow)$} \\
\midrule

\multicolumn{5}{@{}l}{\textit{Backbone: GPT-5}} \\
OpenEvolve
& 7.69 $\pm$ 0.36
& 5.16 $\pm$ 0.95
& 2.48 $\pm$ 0.25
& 6.38 $\pm$ 0.25 \\

GEPA
& 10.04 $\pm$ 1.26
& 5.59 $\pm$ 0.43
& 2.38 $\pm$ 0.52
& 4.03 $\pm$ 0.37 \\

AdaEvolve
& 8.26 $\pm$ 0.11
& 6.44 $\pm$ 0.40
& 2.46 $\pm$ 0.39
& 6.22 $\pm$ 0.19 \\

EvoX
& 9.52 $\pm$ 0.57
& 5.19 $\pm$ 0.23
& 2.47 $\pm$ 0.31
& 5.97 $\pm$ 0.22 \\

LEVI
& 2.90 $\pm$ 0.25
& 1.65 $\pm$ 0.12
& \bestcell{1.20} $\pm$ 0.21
& 2.42 $\pm$ 0.98 \\

SpecEvo
& \bestcell{2.87} $\pm$ 0.38
& \bestcell{1.62} $\pm$ 0.07
& 1.21 $\pm$ 0.27
& \bestcell{2.41} $\pm$ 1.00 \\

\midrule

\multicolumn{5}{@{}l}{\textit{Backbone: Kimi-K2}} \\
OpenEvolve
& 2.25 $\pm$ 0.32
& 1.64 $\pm$ 0.05
& 0.90 $\pm$ 0.15
& 1.97 $\pm$ 0.27 \\

GEPA
& 1.88 $\pm$ 0.29
& 1.55 $\pm$ 0.04
& 1.02 $\pm$ 0.28
& 1.69 $\pm$ 0.12 \\

AdaEvolve
& 2.04 $\pm$ 0.07
& 1.73 $\pm$ 0.27
& 1.15 $\pm$ 0.23
& 1.96 $\pm$ 0.14 \\

EvoX
& 2.24 $\pm$ 0.05
& 1.77 $\pm$ 0.13
& 1.14 $\pm$ 0.31
& 2.16 $\pm$ 0.10 \\

LEVI
& 1.01 $\pm$ 0.17
& \bestcell{0.51} $\pm$ 0.03
& 0.40 $\pm$ 0.18
& \bestcell{0.62} $\pm$ 0.00 \\

SpecEvo
& \bestcell{0.99} $\pm$ 0.20
& 0.53 $\pm$ 0.02
& \bestcell{0.39} $\pm$ 0.19
& 0.65 $\pm$ 0.00 \\

\bottomrule
\end{tabular}
\end{table*}

\subsection{Detailed Results for Mathematical Discovery and Systems Optimization Tasks}
\label{app:breakdown_math}

The results below break down the aggregate trends in Figure~\ref{fig:rq1} by task and backbone configuration. Beyond final performance, we check whether the gains persist when either the frontier model or the lightweight Speculator is changed, which separates the contribution of the search architecture from that of any particular model.

\paragraph{Mathematical discovery.}
Across both frontier backbones, \textsc{SpecEvo} attains the strongest or tied-best single-run result on six of the seven tasks and the leading or tied mean performance on five. Its advantage is most consistent on the Heilbronn and MinMax problems, whose landscapes contain multiple structurally distinct solution families and densely populated local optima. The adaptive archive preserves these alternatives, and the Navigator can synthesize promising mechanisms or reframe the search once lightweight refinement saturates. Even smaller Speculators therefore often match the strongest configurations on these tasks. Signal Processing is the main exception: its noisy evaluation makes short-term progress less reliable, so results vary more across lightweight backbones and conventional baselines remain competitive.

\paragraph{Systems optimization.}
The gains are more uniform on systems tasks, where at least one \textsc{SpecEvo} configuration achieves the best or tied-best mean performance on every problem under both GPT-5 and Kimi-K2. No single lightweight backbone dominates: Qwen3-30B performs strongest on EPLB and LLM-SQL with GPT-5, Qwen3-14B leads Transaction Scheduling, and Qwen3-8B becomes strongest on Transaction Scheduling with Kimi-K2. Even Qwen2.5-7B reaches the performance ceiling on PRISM. This non-monotonic relationship with model size suggests that the gains do not come simply from a more capable Speculator. In the closed-loop workflow, inexpensive models with different strengths carry out high-volume refinement, and population-level evidence determines when strategic redirection is needed.

\paragraph{Robustness and cost efficiency.}
This division of labor also reduces sensitivity to the frontier backbone. Several frontier-only baselines degrade substantially when GPT-5 is replaced by Kimi-K2, particularly on PRISM, whereas \textsc{SpecEvo} stays close to the task ceiling because the frontier model acts through sparse, high-impact interventions instead of shaping every mutation. The same organization reduces API expenditure by 34\%--85\% on mathematical discovery and 51\%--71\% on systems optimization relative to the four frontier-heavy baselines. \textsc{SpecEvo} also generally improves upon LEVI at comparable cost, which indicates that the remaining quality difference comes mainly from how scarce frontier reasoning and accumulated evidence are integrated into the search, not from the model price gap alone.

\begin{table}[tbp]
\centering
\footnotesize
\setlength{\tabcolsep}{2.4pt}
\renewcommand{\arraystretch}{1.15}
\caption{\small Performance comparison on mathematical discovery tasks. Red arrows indicate \textsc{SpecEvo}'s average API cost reduction relative to the four baselines, excluding LEVI.}
\label{tab:math_results_gpt5_kimi}
\begin{adjustbox}{max width=\textwidth}
\begin{tabular}{@{}lcccccccccccccc@{}}
\toprule
\multirow{2}{*}{\textbf{Strategy}}
& \multicolumn{2}{c}{\textbf{Circle Packing}}
& \multicolumn{2}{c}{\textbf{Circle Rect.}}
& \multicolumn{2}{c}{\textbf{Heil. Conv.}}
& \multicolumn{2}{c}{\textbf{Heil. Tri.}}
& \multicolumn{2}{c}{\textbf{MinMax 16-2}}
& \multicolumn{2}{c}{\textbf{MinMax 14-3}}
& \multicolumn{2}{c}{\textbf{Signal}} \\
\cmidrule(lr){2-3}
\cmidrule(lr){4-5}
\cmidrule(lr){6-7}
\cmidrule(lr){8-9}
\cmidrule(lr){10-11}
\cmidrule(lr){12-13}
\cmidrule(lr){14-15}
& \textbf{Avg} & \textbf{Best}
& \textbf{Avg} & \textbf{Best}
& \textbf{Avg} & \textbf{Best}
& \textbf{Avg} & \textbf{Best}
& \textbf{Avg} & \textbf{Best}
& \textbf{Avg} & \textbf{Best}
& \textbf{Avg} & \textbf{Best} \\
\midrule

\multicolumn{15}{@{}l}{\textit{Backbone: GPT-5}} \\
OpenEvolve
& 2.541 $\pm$ 0.000 & 2.541
& 2.341 $\pm$ 0.022 & 2.354
& 0.022 $\pm$ 0.000 & 0.022
& 0.033 $\pm$ 0.002 & \bestcell{0.036}
& \secondcell{0.077} $\pm$ 0.000 & \secondcell{0.077}
& 0.229 $\pm$ 0.008 & 0.238
& 0.595 $\pm$ 0.005  & 0.598 \\
GEPA
& 2.536 $\pm$ 0.010 & 2.541
& 2.313 $\pm$ 0.024 & 2.327
& 0.024 $\pm$ 0.004 & \secondcell{0.029}
& 0.032 $\pm$ 0.000 & 0.032
& \secondcell{0.077} $\pm$ 0.000 & \secondcell{0.077}
& 0.226 $\pm$ 0.002 & 0.227
& 0.593 $\pm$ 0.038 & 0.636  \\
AdaEvolve
& 2.563 $\pm$ 0.037 & 2.606
& 2.296 $\pm$ 0.042 & 2.345
& 0.022 $\pm$ 0.001 & 0.023
& 0.029 $\pm$ 0.000 & 0.029
& \secondcell{0.077} $\pm$ 0.000 & \secondcell{0.077}
& 0.235 $\pm$ 0.003 & 0.237
& \bestcell{0.733} $\pm$ 0.001 & \bestcell{0.734} \\
EvoX
& 2.521 $\pm$ 0.035 & 2.541
& \bestcell{2.352} $\pm$ 0.007 & 2.360
& 0.027 $\pm$ 0.000 & 0.027
& 0.033 $\pm$ 0.001 & 0.034
& \bestcell{\bestcell{0.078}} $\pm$ 0.000 & \bestcell{\bestcell{0.078}}
& \secondcell{0.239} $\pm$ 0.000 & \secondcell{0.239}
& 0.688 $\pm$ 0.013 & 0.703 \\
LEVI
& \secondcell{2.593} $\pm$ 0.001 & 2.595
& 2.301 $\pm$ 0.040 & 2.358
& 0.028 $\pm$ 0.001 & \secondcell{0.029}
& 0.034 $\pm$ 0.000 & \secondcell{0.035}
& \secondcell{0.077} $\pm$ 0.000 & \secondcell{0.077}
& 0.239 $\pm$ 0.000 & \secondcell{0.239}
& 0.593 $\pm$ 0.068 & 0.691 \\
\midrule
\textbf{\textsc{SpecEvo}}
& \multicolumn{2}{c}{\costsave{59}}
& \multicolumn{2}{c}{\costsave{63}}
& \multicolumn{2}{c}{\costsave{81}}
& \multicolumn{2}{c}{\costsave{85}}
& \multicolumn{2}{c}{\costsave{80}}
& \multicolumn{2}{c}{\costsave{70}}
& \multicolumn{2}{c}{\costsave{56}} \\
Qwen3-30b
& 2.580 $\pm$ 0.041 & 2.627
& 2.338 $\pm$ 0.021 & \bestcell{2.363}
& \bestcell{0.030} $\pm$ 0.000 & \bestcell{0.030}
& 0.034 $\pm$ 0.002 & \bestcell{0.036}
& \secondcell{0.077} $\pm$ 0.000 & \bestcell{\bestcell{0.078}}
& \bestcell{\bestcell{0.240}} $\pm$ 0.000 & \bestcell{\bestcell{0.240}}
& \secondcell{0.711} $\pm$ 0.012 & \secondcell{0.717} \\
Qwen3-14b
& \bestcell{2.629} $\pm$ 0.001 & \secondcell{2.630}
& 2.306 $\pm$ 0.012 & 2.342
& \secondcell{0.029} $\pm$ 0.001 & \bestcell{0.030}
& \bestcell{0.036} $\pm$ 0.000 & \bestcell{0.036}
& \secondcell{0.077} $\pm$ 0.000 & \bestcell{0.078}
& \bestcell{0.240} $\pm$ 0.000 & \bestcell{0.240}
& 0.545 $\pm$ 0.012 & 0.552  \\
Qwen3-8b
& 2.572 $\pm$ 0.055 & \bestcell{2.636}
& \secondcell{2.351} $\pm$ 0.008 & \secondcell{2.361}
& \bestcell{0.030} $\pm$ 0.000 & \bestcell{0.030}
& \secondcell{0.035} $\pm$ 0.001 & \bestcell{0.036}
& \secondcell{0.077} $\pm$ 0.000 & \bestcell{0.078}
& \bestcell{0.240} $\pm$ 0.000 & \bestcell{0.240}
& 0.550 $\pm$ 0.004 & 0.553
\\
Qwen2.5-7b
& 2.517 $\pm$ 0.054 & 2.548
& 2.333 $\pm$ 0.052 & \bestcell{2.363}
& 0.025 $\pm$ 0.001 & 0.026
& 0.034 $\pm$ 0.002 & \bestcell{0.036}
& \bestcell{0.078} $\pm$ 0.000 & \bestcell{0.078}
& \bestcell{0.240} $\pm$ 0.000 & \bestcell{0.240}
& 0.708 $\pm$ 0.009 & 0.713 \\
\midrule
\midrule

\multicolumn{15}{@{}l}{\textit{Backbone: Kimi-K2}} \\
OpenEvolve
& 2.379 $\pm$ 0.109 & 2.500
& 2.184 $\pm$ 0.171 & 2.296
& 0.018 $\pm$ 0.001 & 0.019
& 0.026 $\pm$ 0.001 & 0.027
& 0.068 $\pm$ 0.011 & 0.075
& 0.213 $\pm$ 0.004 & 0.218
& \bestcell{0.671} $\pm$ 0.045 & \bestcell{0.709} \\
GEPA
& 2.429 $\pm$ 0.062 & 2.481
& 2.167 $\pm$ 0.145 & 2.251
& 0.019 $\pm$ 0.002 & 0.022
& 0.021 $\pm$ 0.014 & \secondcell{0.035}
& 0.068 $\pm$ 0.011 & 0.075
& 0.211 $\pm$ 0.000 & 0.211
& 0.643 $\pm$ 0.020 & 0.658 \\
AdaEvolve
& \bestcell{2.595} $\pm$ 0.037 & 2.630
& 2.256 $\pm$ 0.005 & 2.261
& 0.025 $\pm$ 0.002 & 0.026
& 0.028 $\pm$ 0.005 & 0.031
& 0.074 $\pm$ 0.004 & \bestcell{\bestcell{0.078}}
& 0.220 $\pm$ 0.008 & 0.227
& 0.614 $\pm$ 0.090 & 0.693 \\
EvoX
& 2.561 $\pm$ 0.083 & 2.617
& 2.266 $\pm$ 0.063 & 2.336
& 0.024 $\pm$ 0.003 & 0.026
& 0.025 $\pm$ 0.003 & 0.027
& 0.073 $\pm$ 0.002 & 0.075
& 0.213 $\pm$ 0.004 & 0.218
& \secondcell{0.655} $\pm$ 0.031 & 0.677 \\
LEVI
& 2.553 $\pm$ 0.045 & 2.618
& 2.185 $\pm$ 0.121 & \secondcell{2.357}
& 0.023 $\pm$ 0.001 & 0.025
& \secondcell{0.033} $\pm$ 0.000 & 0.034
& \secondcell{0.076} $\pm$ 0.001 & \secondcell{0.077}
& 0.237 $\pm$ 0.000 & 0.238
& 0.597 $\pm$ 0.036 & 0.623 \\
\midrule
\textbf{\textsc{SpecEvo}}
& \multicolumn{2}{c}{\costsave{55}}
& \multicolumn{2}{c}{\costsave{51}}
& \multicolumn{2}{c}{\costsave{73}}
& \multicolumn{2}{c}{\costsave{68}}
& \multicolumn{2}{c}{\costsave{34}}
& \multicolumn{2}{c}{\costsave{34}}
& \multicolumn{2}{c}{\costsave{37}} \\
Qwen3-30b
& \secondcell{2.593} $\pm$ 0.037 & \bestcell{2.635}
& \bestcell{2.334} $\pm$ 0.029 & \bestcell{2.361}
& 0.026 $\pm$ 0.001 & \secondcell{0.027}
& 0.031 $\pm$ 0.002 & 0.033
& \bestcell{\bestcell{0.078}} $\pm$ 0.000 & \bestcell{0.078}
& 0.223 $\pm$ 0.012 & 0.236
& 0.629 $\pm$ 0.065 & \secondcell{0.703} \\
Qwen3-14b
& 2.539 $\pm$ 0.035 & 2.559
& 2.278 $\pm$ 0.007 & 2.282
& \bestcell{0.030} $\pm$ 0.001 & \bestcell{0.031}
& 0.030 $\pm$ 0.007 & 0.034
& \bestcell{0.078} $\pm$ 0.000 & \bestcell{0.078}
& \bestcell{0.239} $\pm$ 0.001 & \bestcell{0.240}
& 0.624 $\pm$ 0.045 & 0.650  \\
Qwen3-8b
& 2.584 $\pm$ 0.080 & \secondcell{2.631}
& 2.269 $\pm$ 0.146 & 2.353
& \secondcell{0.027} $\pm$ 0.000 & \secondcell{0.027}
& \bestcell{0.034} $\pm$ 0.003 & \bestcell{0.036}
& \bestcell{0.078} $\pm$ 0.000 & \bestcell{0.078}
& 0.229 $\pm$ 0.009 & \secondcell{0.239}
& 0.613 $\pm$ 0.065 & 0.650
\\
Qwen2.5-7b
& 2.489 $\pm$ 0.013 & 2.497
& \secondcell{2.315} $\pm$ 0.055 & 2.347
& 0.023 $\pm$ 0.003 & 0.025
& \secondcell{0.033} $\pm$ 0.003 & \secondcell{0.035}
& 0.073 $\pm$ 0.005 & 0.076
& \secondcell{0.238} $\pm$ 0.003 & \bestcell{0.240}
& 0.615 $\pm$ 0.020 & 0.645 \\
\bottomrule
\end{tabular}
\end{adjustbox}
\label{tab:math-time}
\end{table}

\begin{table}[tbp]
\centering
\scriptsize
\setlength{\tabcolsep}{4pt}
\renewcommand{\arraystretch}{1.12}
\caption{\small Performance comparison on systems optimization tasks.}
\label{tab:system_results_gpt5_kimi}
\resizebox{0.75\textwidth}{!}{%
\begin{tabular}{@{}lcccccccc@{}}
\toprule
\multirow{2}{*}{\textbf{Strategy}}
& \multicolumn{2}{c}{\textbf{EPLB $(\uparrow)$}}
& \multicolumn{2}{c}{\textbf{LLM-SQL $(\uparrow)$}}
& \multicolumn{2}{c}{\textbf{Transaction $(\uparrow)$}}
& \multicolumn{2}{c}{\textbf{PRISM $(\uparrow)$}} \\
\cmidrule(lr){2-3}
\cmidrule(lr){4-5}
\cmidrule(lr){6-7}
\cmidrule(lr){8-9}
& \textbf{Avg} & \textbf{Best}
& \textbf{Avg} & \textbf{Best}
& \textbf{Avg} & \textbf{Best}
& \textbf{Avg} & \textbf{Best} \\
\midrule

\multicolumn{9}{@{}l}{\textit{Backbone: GPT-5}} \\
OpenEvolve
& 0.1398 $\pm$ {0.011} & 0.1459
& 0.7152 $\pm$ {0.024} & 0.7289
& 4006.0 $\pm$ {65.47} & 4081.6
& \secondcell{26.242} $\pm$ { 0.008} & \secondcell{26.247} \\
GEPA
& 0.1394 $\pm$ {0.010} & 0.1453
& 0.7242 $\pm$ {0.006} & 0.7315
& 4078.2 $\pm$ {213.73} & 4201.6
& 26.235 $\pm$ {0.002} & 26.236 \\
AdaEvolve
& 0.1462 $\pm$ {0.000} & 0.1464
& \secondcell{0.7287} $\pm$ {0.004} & 0.7307
& 3958.7 $\pm$ {195.14} & 4184.1
& \bestcell{\bestcell{26.256}} $\pm$ {0.000} & \bestcell{\bestcell{26.256}} \\
EvoX
& 0.1448 $\pm$ {0.001} & 0.1454
& 0.6936 $\pm$ {0.031} & 0.7298
& 3827.8 $\pm$ {354.60} & 4237.3
& \bestcell{\bestcell{26.256}} $\pm$ {0.000} & \bestcell{\bestcell{26.256}} \\
LEVI
& 0.1446 $\pm$ 0.0001 & 0.1449
& 0.7228 $\pm$ 0.0141 & \secondcell{0.7328}
& 3777.1 $\pm$ 180.37 & 4032.2
& \bestcell{26.256} $\pm$ 0.000 & \bestcell{26.256} \\
\midrule
\textbf{\textsc{SpecEvo}}
& \multicolumn{2}{c}{\costsave{68}}
& \multicolumn{2}{c}{\costsave{71}}
& \multicolumn{2}{c}{\costsave{51}}
& \multicolumn{2}{c}{\costsave{57}} \\
Qwen3-30b
& \bestcell{0.1472} $\pm$ {0.001} & \bestcell{0.1485}
& \bestcell{0.7327} $\pm$ {0.003} & \bestcell{0.7347}
& \secondcell{4173.3} $\pm$ {71.01} & \secondcell{4255.3}
& \bestcell{\bestcell{26.256}} $\pm$ {0.000} & \bestcell{\bestcell{26.256}} \\
Qwen3-14b
& 0.1458 $\pm$ 0.0005 & 0.1461
& 0.7226 $\pm$ 0.0147 & 0.7311
& \bestcell{4366.8} $\pm$ 0.0000 & \bestcell{4366.8}
& \bestcell{\bestcell{26.256}} $\pm$ {0.000} & \bestcell{\bestcell{26.256}} \\
Qwen3-8b
& \secondcell{0.1468} $\pm$ 0.0008 & \secondcell{0.1472}
& 0.7268 $\pm$ 0.0051 & 0.7307
& 4132.2 $\pm$ 406.34 & \bestcell{4366.8}
& \bestcell{\bestcell{26.256}} $\pm$ {0.000} & \bestcell{\bestcell{26.256}} \\
Qwen2.5-7b
& 0.1427 $\pm$ 0.0021 & 0.1438
& 0.7210 $\pm$ 0.0130 & 0.7307
& 4080.4 $\pm$ 65.04 & 4184.1
& \bestcell{\bestcell{26.256}} $\pm$ {0.000} & \bestcell{\bestcell{26.256}} \\
\midrule
\midrule

\multicolumn{9}{@{}l}{\textit{Backbone: Kimi-K2}} \\
OpenEvolve
& 0.1339 $\pm$ 0.010 & 0.1453
& 0.6980 $\pm$ 0.005 & 0.7012
& 3966.1 $\pm$ 368.9 & 4237.2
& 23.443 $\pm$ 0.922 & 24.498 \\
GEPA
& 0.1287 $\pm$ 0.001 & 0.1297
& \secondcell{0.7043} $\pm$ 0.003 & 0.7071
& 3748.1 $\pm$ 126.7 & 3891.0
& 23.049 $\pm$ 0.932 & 24.120 \\
AdaEvolve
& 0.1312 $\pm$ 0.005 & 0.1373
& 0.6996 $\pm$ 0.021 & 0.7139
& 3761.1 $\pm$ 193.5 & 3984.0
& \secondcell{26.242} $\pm$ 0.019 & \bestcell{\bestcell{26.256}} \\
EvoX
& 0.1337 $\pm$ 0.010 & 0.1451
& 0.7038 $\pm$ 0.015 & 0.7181
& 3613.9 $\pm$ 143.6 & 3759.3
& 25.436 $\pm$ 1.229 & \secondcell{26.233} \\
LEVI
& 0.1446 $\pm$ 0.000 & 0.1449
& 0.6654 $\pm$ 0.013 & 0.6843
& \secondcell{4201.6} $\pm$ 0.000 & 4201.6
& 24.342 $\pm$ 0.034 & 24.498 \\
\midrule
\textbf{\textsc{SpecEvo}}
& \multicolumn{2}{c}{\costsave{53}}
& \multicolumn{2}{c}{\costsave{68}}
& \multicolumn{2}{c}{\costsave{63}}
& \multicolumn{2}{c}{\costsave{67}} \\
Qwen3-30b
& \bestcell{0.1464} $\pm$ 0.001 & \bestcell{0.1470}
& 0.7034 $\pm$ 0.016 & \secondcell{0.7201}
& 4143.2 $\pm$ 178.8 & 4255.3
& 26.212 $\pm$ 0.076 & \bestcell{\bestcell{26.256}} \\
Qwen3-14b
& \secondcell{0.1457} $\pm$ 0.0006 & \secondcell{0.1464}
& \bestcell{0.7080} $\pm$ 0.0237 & \bestcell{0.7217}
& 4183.0 $\pm$ 175.7 & \secondcell{4385.9}
& \bestcell{26.256} $\pm$ 0.000 & \bestcell{26.256}  \\
Qwen3-8b
& 0.1438 $\pm$ 0.0002 & 0.1440
& 0.6909 $\pm$ 0.0282 & 0.7072
& \bestcell{4398.9} $\pm$ 22.40 & \bestcell{4424.8}
& 26.240 $\pm$ 0.013 & \bestcell{26.256}  \\
Qwen2.5-7b
& 0.1448 $\pm$ 0.0008 & 0.1453
& 0.6996 $\pm$ 0.0234 & 0.7130
& 4032.5 $\pm$ 354.4 & 4237.2
& \bestcell{26.256} $\pm$ 0.000 & \bestcell{26.256}  \\

\bottomrule
\end{tabular}%
}
\label{tab:system-time}
\end{table}

\subsection{Detailed Results for Each CO-Bench Problem Category}
\label{app:breakdown_cobench}

\begin{figure}
    \centering
    \includegraphics[width=1\linewidth]{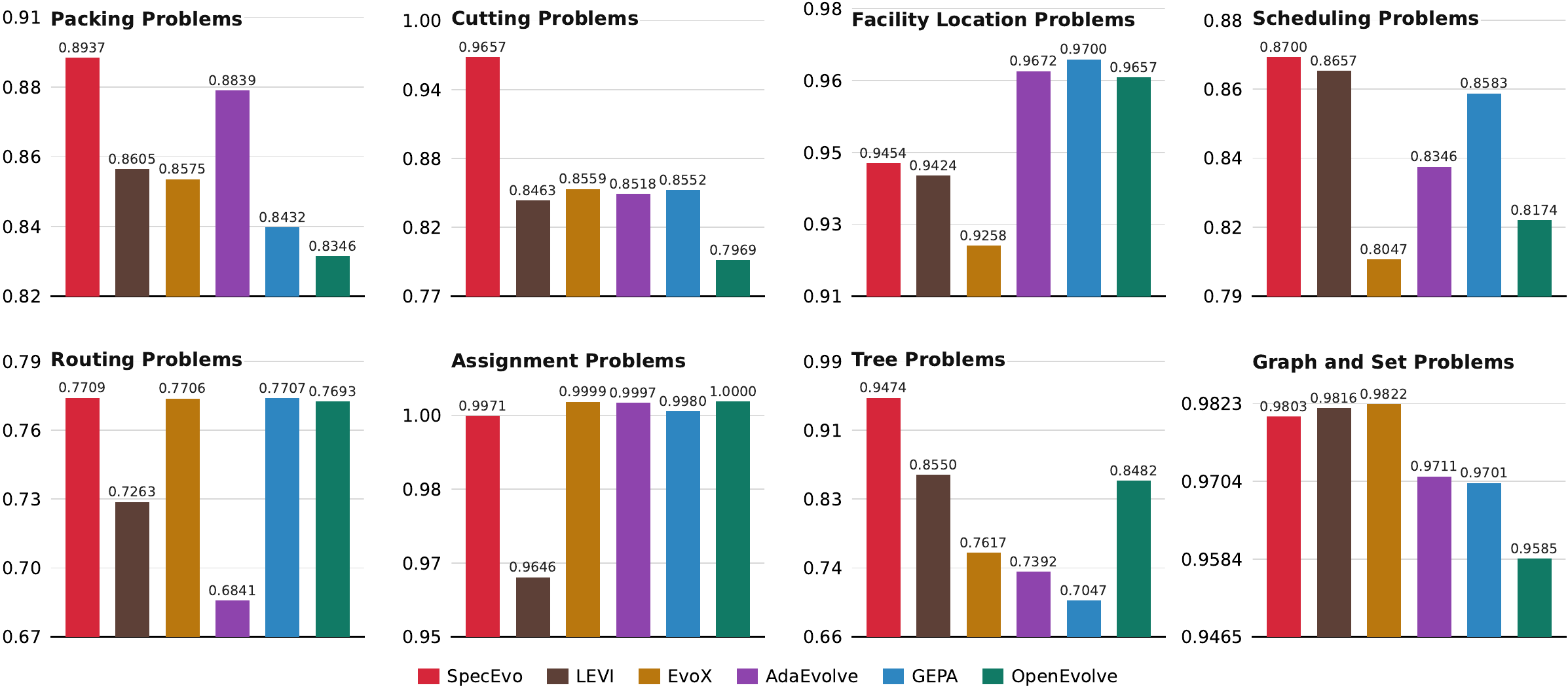}
\caption{
Category-level performance on CO-Bench. Each bar represents the
average normalized score over all problems within the corresponding
category. Higher is better.
}
\label{fig:cobench_category_breakdown}
\end{figure}

Figure~\ref{fig:cobench_category_breakdown} summarizes performance
across the eight CO-Bench categories, while Table~\ref{tab:cobench_per_task} reports the complete results for all 36 problems. \textsc{SpecEvo} achieves the highest category-level score in five categories---Packing, Cutting, Scheduling, Routing, and Tree---with the largest advantages observed on Cutting and Tree problems. At the problem
level, \textsc{SpecEvo} ranks first on 20/36 problems, with strong results on assortment, Euclidean Steiner, flow-shop scheduling, and periodic vehicle-routing problems. Assignment is nearly saturated across methods, and results on Facility Location are more mixed.

\begin{figure}[tbp]
    \centering
\includegraphics[width=0.8\linewidth]{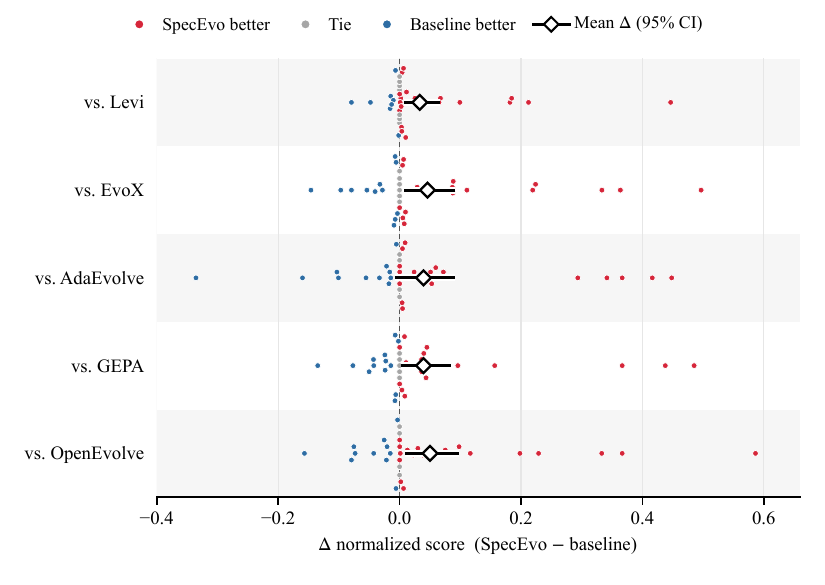}
\caption{Per-task score differences on the 36 CO-Bench tasks.}
    \label{significant_testing:co_bench}
\end{figure}

Figure~\ref{significant_testing:co_bench} and Table~\ref{tab:cobench_significance} further examine whether these gains are consistent across individual problems. The mean paired score difference is
positive against every baseline, ranging from $+0.0332$ against LEVI to $+0.0500$ against OpenEvolve, and \textsc{SpecEvo} wins more problems than it loses in all five pairwise comparisons. The bootstrap confidence intervals for the mean difference are also positive for four of the five baselines, with AdaEvolve being the exception. 

\begin{table*}[p]
\centering
\caption{Per-task normalized scores on the 36 CO-Bench tasks.
Higher scores indicate better performance.}
\label{tab:cobench_per_task}

\begin{adjustbox}{
    max width=\textwidth,
    max totalheight=0.88\textheight,
    keepaspectratio,
    center
}
\begin{minipage}{\textwidth}
\centering

\fontsize{8.7}{9.0}\selectfont
\setlength{\tabcolsep}{3.0pt}
\renewcommand{\arraystretch}{1.08}
\setlength{\aboverulesep}{0.4pt}
\setlength{\belowrulesep}{0.4pt}

\begin{tabularx}{\textwidth}{
    @{}
    >{\raggedright\arraybackslash}X
    cccccc
    @{}
}
\toprule
\textbf{Task}
& \textbf{OpenEvolve}
& \textbf{GEPA}
& \textbf{AdaEvolve}
& \textbf{EvoX}
& \textbf{LEVI}
& \textbf{\textsc{SpecEvo}} \\
\midrule

\multicolumn{7}{@{}l}{\textbf{Packing Problems}} \\

\quad Bin packing -- one-dimensional
& 0.8963
& 0.8903
& \secondcell{0.9213}
& 0.8979
& 0.8870
& \bestcell{0.9863} \\

\quad Multi-Demand Multidimensional Knapsack
& 0.4473
& 0.5196
& \secondcell{0.6234}
& 0.5893
& 0.6090
& \bestcell{0.6763} \\

\quad Multidimensional knapsack problem
& 0.9867
& 0.9846
& \secondcell{0.9890}
& 0.9866
& \secondcell{0.9890}
& \bestcell{0.9932} \\

\quad Container loading
& 0.8362
& \secondcell{0.8392}
& 0.8281
& 0.8156
& \bestcell{0.8396}
& 0.8329 \\

\quad Container loading with weight restrictions
& 0.3540
& 0.3577
& \bestcell{0.6162}
& \secondcell{0.3777}
& 0.2808
& 0.2808 \\

\quad Packing unequal circles
& 0.9666
& 0.9666
& 0.9666
& 1.0000
& \secondcell{1.1209}
& \bestcell{1.3333} \\

\quad Packing unequal circles area
& 1.0293
& \bestcell{1.0913}
& \secondcell{1.0847}
& 1.0574
& 1.0686
& 1.0671 \\

\quad Packing unequal rectangles and squares
& \secondcell{0.9616}
& 0.9061
& \bestcell{0.9855}
& \secondcell{0.9616}
& \secondcell{0.9616}
& 0.8823 \\

\quad Packing unequal rectangles and squares area
& \bestcell{1.0338}
& \secondcell{1.0337}
& 0.9402
& 1.0315
& 0.9880
& 0.9912 \\

\midrule
\multicolumn{7}{@{}l}{\textbf{Cutting Problems}} \\

\quad Assortment problem
& 0.4164
& \secondcell{0.5650}
& 0.5545
& 0.5060
& 0.5562
& \bestcell{1.0027} \\

\quad Constrained guillotine cutting
& 0.8709
& 0.9247
& \secondcell{0.9569}
& \bestcell{0.9892}
& 0.9176
& 0.9354 \\

\quad Unconstrained guillotine cutting
& 0.9460
& \bestcell{0.9829}
& 0.9636
& \secondcell{0.9777}
& 0.9571
& 0.9685 \\

\quad Constrained non-guillotine cutting
& \secondcell{0.9542}
& 0.9481
& 0.9323
& 0.9508
& \secondcell{0.9542}
& \bestcell{0.9562} \\

\midrule
\multicolumn{7}{@{}l}{\textbf{Facility Location Problems}} \\

\quad Capacitated warehouse location
& 0.8732
& 0.9122
& 0.8763
& 0.7245
& \secondcell{0.9232}
& \bestcell{0.9485} \\

\quad Uncapacitated warehouse location
& \bestcell{1.0000}
& \bestcell{1.0000}
& \bestcell{1.0000}
& \bestcell{1.0000}
& \bestcell{1.0000}
& \bestcell{1.0000} \\

\quad Capacitated $p$-median
& \secondcell{0.9896}
& 0.9677
& \bestcell{0.9928}
& 0.9789
& 0.8475
& 0.8329 \\

\quad Uncapacitated $p$-median
& \secondcell{0.9998}
& \bestcell{1.0000}
& \secondcell{0.9998}
& \bestcell{1.0000}
& 0.9990
& \bestcell{1.0000} \\

\midrule
\multicolumn{7}{@{}l}{\textbf{Scheduling Problems}} \\

\quad Aircraft landing
& 0.9026
& 0.9047
& \bestcell{0.9827}
& \secondcell{0.9144}
& 0.8976
& 0.8820 \\

\quad Crew scheduling
& \secondcell{0.5285}
& 0.4682
& \bestcell{0.5294}
& 0.2939
& 0.5132
& 0.5132 \\

\quad Common due date scheduling
& 0.8887
& 0.9468
& 0.9271
& \bestcell{0.9902}
& 0.9802
& \secondcell{0.9867} \\

\quad Flow shop scheduling
& 0.6364
& \secondcell{0.9356}
& 0.6279
& 0.8585
& 0.9300
& \bestcell{0.9694} \\

\quad Hybrid Reentrant Shop Scheduling
& \secondcell{0.9917}
& \secondcell{0.9917}
& \secondcell{0.9917}
& \secondcell{0.9917}
& \secondcell{0.9917}
& \bestcell{0.9918} \\

\quad Job shop scheduling
& 0.8116
& 0.8238
& 0.8416
& 0.7849
& \secondcell{0.8501}
& \bestcell{0.8602} \\

\quad Open shop scheduling
& \bestcell{0.9621}
& 0.9370
& \secondcell{0.9421}
& 0.7991
& 0.8971
& 0.8868 \\

\midrule
\multicolumn{7}{@{}l}{\textbf{Routing Problems}} \\

\quad Traveling salesman problem
& 0.9465
& \secondcell{0.9641}
& 0.9543
& 0.9494
& \bestcell{0.9691}
& 0.9211 \\

\quad Period vehicle routing problem
& 0.7527
& 0.7392
& 0.4892
& \secondcell{0.7538}
& 0.6010
& \bestcell{0.7828} \\

\quad Resource-constrained shortest path
& \bestcell{0.6087}
& \bestcell{0.6087}
& \bestcell{0.6087}
& \bestcell{0.6087}
& \bestcell{0.6087}
& \bestcell{0.6087} \\

\midrule
\multicolumn{7}{@{}l}{\textbf{Assignment Problems}} \\

\quad Assignment problem
& \bestcell{1.0000}
& \bestcell{1.0000}
& \bestcell{1.0000}
& \bestcell{1.0000}
& \bestcell{1.0000}
& \bestcell{1.0000} \\

\quad Generalized assignment problem
& \bestcell{1.0000}
& 0.9960
& 0.9994
& \secondcell{0.9998}
& 0.9292
& 0.9941 \\

\midrule
\multicolumn{7}{@{}l}{\textbf{Tree Problems}} \\

\quad Euclidean Steiner problem
& \secondcell{0.6965}
& 0.4095
& 0.4784
& 0.5310
& 0.7101
& \bestcell{0.8947} \\

\quad Corporate structuring
& \secondcell{0.9999}
& \secondcell{0.9999}
& \secondcell{0.9999}
& 0.9923
& \secondcell{0.9999}
& \bestcell{1.0000} \\

\midrule
\multicolumn{7}{@{}l}{\textbf{Graph and Set Problems}} \\

\quad Maximal Independent Set
& \bestcell{0.9347}
& 0.9202
& \secondcell{0.9275}
& 0.9202
& 0.9261
& 0.9130 \\

\quad Graph coloring
& 0.8835
& 0.9382
& \secondcell{0.9500}
& \bestcell{1.0000}
& \bestcell{1.0000}
& \bestcell{1.0000} \\

\quad Equitable partitioning problem
& \bestcell{1.0000}
& \bestcell{1.0000}
& \bestcell{1.0000}
& \bestcell{1.0000}
& \bestcell{1.0000}
& \bestcell{1.0000} \\

\quad Set partitioning
& 0.9912
& \bestcell{1.0000}
& 0.9912
& \secondcell{0.9997}
& 0.9898
& 0.9924 \\

\quad Set covering
& 0.9831
& 0.9921
& 0.9869
& 0.9911
& \secondcell{0.9923}
& \bestcell{0.9960} \\

\bottomrule
\end{tabularx}
\end{minipage}
\end{adjustbox}
\end{table*}

\begin{table}[tbp]
\centering
\caption{
Paired statistical comparison of \textsc{SpecEvo} against baselines on the 36 CO-Bench tasks.
}
\label{tab:cobench_significance}
\small
\setlength{\tabcolsep}{6pt}
\begin{tabular}{lccc}
\toprule
\textbf{Baseline}
& \textbf{W / T / L}
& \textbf{Mean $\Delta$}
& \textbf{95\% CI} \\
\midrule
LEVI
& 21 / 7 / 8
& $+0.0332$
& $[+0.0070,\,+0.0668]$ \\

EvoX
& 18 / 6 / 12
& $+0.0460$
& $[+0.0073,\,+0.0906]$ \\

AdaEvolve
& 21 / 4 / 11
& $+0.0395$
& $[-0.0076,\,+0.0908]$ \\

GEPA
& 18 / 5 / 13
& $+0.0393$
& $[+0.0020,\,+0.0841]$ \\

OpenEvolve
& 21 / 4 / 11
& $+0.0500$
& $[+0.0091,\,+0.0985]$ \\
\bottomrule
\end{tabular}
\end{table}

\subsection{Detailed Results on LSR-Synth}
\label{app:breakdown_lsrsynth}

We extend the comparison on LSR-Synth to three additional LLM-based equation discovery methods: LLM-SR~\citep{shojaee2025llma}, LaSR~\citep{grayeli2024symbolic}, and SGA~\citep{ma2024llm}.
In addition to normalized mean squared error (NMSE), we report numeric precision ($\mathrm{Acc}_{0.1}$) and symbolic accuracy (SA), following the evaluation protocol of LLM-SRBench~\citep{shojaee2025llm}.

For a discovered equation, accuracy to tolerance $\tau$ is defined as
\begin{equation}
\mathrm{Acc}_{\tau}
=
\mathbb{I}
\left[
\max_{1 \leq i \leq N_{\mathrm{test}}}
\left|
\frac{\hat{y}_i-y_i}{y_i}
\right|
\leq \tau
\right],
\label{eq:acc_tau}
\end{equation}
where $\hat{y}_i$ and $y_i$ denote the predicted and ground-truth
values, respectively. We use $\tau=0.1$ and report the percentage
of problems satisfying this criterion. While NMSE measures
aggregate prediction error, $\mathrm{Acc}_{0.1}$ requires relative
error to remain within the specified tolerance at every test point.
We compute both metrics separately on the in-distribution (ID)
and out-of-distribution (OOD) test sets to assess interpolation
and extrapolation performance.

Symbolic accuracy complements these numerical metrics by assessing
whether the discovered hypothesis recovers the ground-truth
symbolic form. Following LLM-SRBench~\citep{shojaee2025llm}, we use
GPT-4o to judge mathematical equivalence between the predicted and
ground-truth expressions after removing parameters and constants.
This model-based assessment accommodates different representations
of equivalent expressions without requiring an exact syntactic
match. We report SA as the percentage of problems judged
symbolically equivalent. Unlike the numerical metrics, SA compares
the equations themselves and is therefore reported once per domain,
rather than separately for the ID and OOD splits.

Table~\ref{tab:lsr_synth_results} summarizes the domain-level results
under a common budget of 500 candidate evaluations. We find that
\textsc{SpecEvo} achieves the lowest NMSE in six of the eight
domain--split settings, including three of the four OOD settings.
Biology is the exception, where LLM-SR obtains lower NMSE on both
splits. For numeric precision, \textsc{SpecEvo} achieves the highest values on both splits in chemistry, biology, and physics. In material science, it ties for the highest ID precision, while several baselines achieve higher OOD precision.
The table further shows that
\textsc{SpecEvo} achieves the highest SA in biology ($41.66\%$)
and physics ($11.36\%$), while tying for the highest SA in
chemistry ($13.88\%$) and material science ($24.0\%$).
These results point to improvements in both numerical prediction and symbolic-form recovery, and the consistently low OOD NMSE is further evidence of generalization beyond the training ranges.

\begin{table*}[t]
\centering
\small
\setlength{\tabcolsep}{5.5pt}
\renewcommand{\arraystretch}{1.16}
\caption{
Performance comparison on LSR-Synth under a fixed budget of 500 candidate evaluations.
Metrics include normalized mean squared error (NMSE), numeric precision
($\mathrm{Acc}_{0.1}$), and symbolic accuracy (SA).
}
\label{tab:lsr_synth_results}
\begin{adjustbox}{max width=\textwidth}
\begin{tabular}{@{}llccccc@{}}
\toprule
\multirow{2}{*}{\textbf{Domain}}
& \multirow{2}{*}{\textbf{Method}}
& \multicolumn{2}{c}{\textbf{NMSE} $(\downarrow)$}
& \multicolumn{2}{c}{\textbf{Acc.}$_{0.1}$ \textbf{(\%)} $(\uparrow)$}
& \multirow{2}{*}{\textbf{SA (\%)} $(\uparrow)$} \\
\cmidrule(lr){3-4}
\cmidrule(lr){5-6}
& & \textbf{ID} & \textbf{OOD} & \textbf{ID} & \textbf{OOD} & \\
\midrule

\multirow{10}{*}{Chemistry}
& LLM-SR     & 5.69e-07       & 1.98e-02       & 36.1  & 44.3   & \bestcell{13.88} \\
& LaSR       & 6.43e-05       & 5.46e-01       & 28.9  & 30.5   & 8.33 \\
& SGA        & 8.12e-06       & 5.25e-02       & 31.5  & 41.4   & 0.0 \\
& OpenEvolve & 5.27e-06 & 4.99e-02 & 33.3 & 41.7 & 5.55 \\
& GEPA       & 1.72e-05 & 2.69e-01 & 19.4 & 38.9 & 11.11 \\
& AdaEvolve  & 2.48e-06 & 6.05e-02 & 33.3 & 41.7 & 13.88 \\
& EvoX       & 6.87e-06 & 8.68e-02 & 27.8 & 44.4 & 11.11 \\
& RelayEvolve & 2.59e-06 & 1.30e-01 & 36.1 & 38.9 & 5.55 \\
& LEVI       & 5.80e-06 & 1.17e-01 & 27.8 & 42.5 & 8.33 \\
& \textbf{\textsc{SpecEvo}}
& \bestcell{4.57e-07} & \bestcell{1.68e-02}
& \bestcell{41.7} & \bestcell{55.6} & \bestcell{13.88} \\
\midrule

\multirow{10}{*}{Biology}
& LLM-SR     & \bestcell{3.30e-07}       & 1.08e-03       & 19.2   & 23.8   & 33.33 \\
& LaSR       &    2.98e-06    & 3.14e-02       & 18.5   & 21.7   & 4.16 \\
& SGA        & 3.47e-05       & 1.08e-02       & 8.3   & 12.5   & 4.16 \\
& OpenEvolve & 6.33e-05 & 7.48e-01 & 8.3  & 12.5 & 29.16 \\
& GEPA       & 8.79e-05 & 1.92e+00 & 8.3  & 12.5 & 12.5 \\
& AdaEvolve  & 9.92e-06 & 1.28e-01 & 12.5 & 16.7 & 29.16 \\
& EvoX       & 3.79e-05 & 5.83e-01 & 8.3  & 16.7 & 29.16 \\
& RelayEvolve & 1.01e-05 & 1.56e-01 & 8.3 & 12.5 & 33.33 \\
& LEVI       & 3.87e-06 & 1.96e-02 & 8.3  & 12.5 & 12.5 \\
& \textbf{\textsc{SpecEvo}}
& \bestcell{3.34e-07} & \bestcell{3.74e-03}
& \bestcell{20.8} & \bestcell{29.2} & \bestcell{41.66} \\
\midrule

\multirow{10}{*}{Physics}
& LLM-SR     & 1.75e-05       & 2.63e-03       & 10.0  & 12.3   & 9.09 \\
& LaSR       & 2.46e-05       & 2.33e-03       & 6.8  & 11.4   & 0.00 \\
& SGA        & 5.21e-06       & 1.35e-02       & 11.5  & 16.4   & 0.00 \\
& OpenEvolve & 3.29e-04 & 6.93e-03 & 4.5 & 11.4 & 0.00 \\
& GEPA       & 1.53e-03 & 2.82e-02 & 0.0 & 6.8  & 6.81 \\
& AdaEvolve  & 2.79e-04 & 1.32e-02 & 6.8 & 11.4 & 9.09 \\
& EvoX       & 6.66e-04 & 5.32e-02 & 6.8 & 6.8  & 6.81 \\
& RelayEvolve & 1.81e-04 & 3.30e-03 & 2.3 & 9.1 & 0.00 \\
& LEVI       & 1.38e-05 & 2.23e-03 & 5.6 & 12.8 & 9.09 \\
& \textbf{\textsc{SpecEvo}}
& \bestcell{2.92e-06} & \bestcell{4.75e-04}
& \bestcell{20.5} & \bestcell{20.5} & \bestcell{11.36} \\
\midrule

\multirow{10}{*}{Material Science}
& LLM-SR     & 5.24e-08       & 1.92e-06       & \bestcell{52.0}   & 88.0              & 12 \\
& LaSR       & 6.25e-07       & 4.75e-5       & 36.0   & 72.0              & 8 \\
& SGA        & 4.26e-07       & 1.88e-06       & 48.0   & \bestcell{96.0}              & 8 \\
& OpenEvolve & 5.92e-08 & 2.44e-06 & 48.0 & \bestcell{96.0} & \bestcell{24} \\
& GEPA       & 1.86e-07 & 7.57e-05 & 28.0 & \bestcell{96.0} & \bestcell{24} \\
& AdaEvolve  & 6.32e-07 & 1.18e-04 & 44.0 & 72.0            & 12 \\
& EvoX       & 9.05e-08 & 1.63e-05 & 36.0 & \bestcell{96.0} & 16 \\
& RelayEvolve & 5.79e-08 & 6.72e-06 & 48.0 & \bestcell{96.0} & 20 \\
& LEVI       & 5.73e-07 & 8.36e-06 & 36.0 & 68.0            & 20 \\
& \textbf{\textsc{SpecEvo}}
& \bestcell{4.84e-08} & \bestcell{1.68e-06}
& \bestcell{52.0} & 88.0 & \bestcell{24} \\
\bottomrule
\end{tabular}
\end{adjustbox}
\end{table*}

\begin{figure}[t]
    \centering
\includegraphics[width=0.8\linewidth]{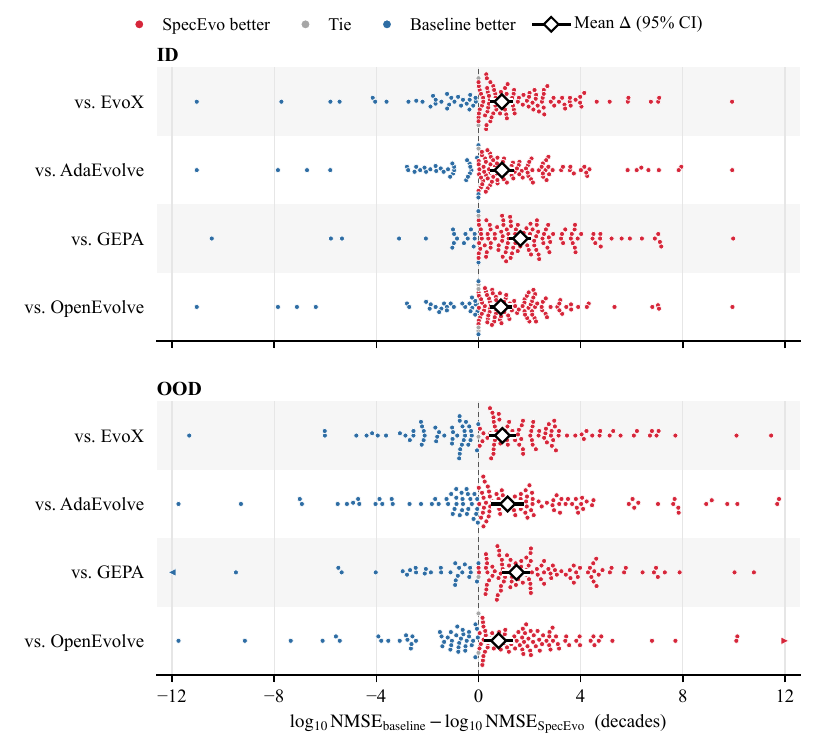}
\caption{Per-task score differences on the 129 symbolic regression tasks.}
    \label{significant_testing:lsr_bench}
\end{figure}

Figure~\ref{significant_testing:lsr_bench} and Table~\ref{tab:lsrsynth_significance} provide a task-level comparison on the 129 LSR-Synth problems under both ID and OOD evaluation. For each problem, we measure the paired difference as
$\Delta=\log_{10}\mathrm{NMSE}_{\mathrm{baseline}}
-\log_{10}\mathrm{NMSE}_{\textsc{SpecEvo}}$,
such that positive values indicate lower NMSE for \textsc{SpecEvo}.
Across all four baselines, \textsc{SpecEvo} wins substantially more problems
than it loses under both evaluation settings. On ID data, it wins
95--109 of the 129 problems, compared with 19--33 losses, while on OOD data
it wins 81--100 problems compared with 28--47 losses.
Mean paired differences are also positive for every baseline,
ranging from $+0.87$ to $+1.64$ decades on ID data and from $+0.79$ to
$+1.48$ decades on OOD data, with all bootstrap 95\% confidence intervals
remaining above zero.
The strongest separation is observed against GEPA, for which
\textsc{SpecEvo} wins 109/129 ID problems and 100/129 OOD problems, with
mean differences of $+1.64$ and $+1.48$ decades, respectively.
The aggregate LSR-Synth improvements are thus reflected at the task level: \textsc{SpecEvo} wins a majority of problems against every baseline under both ID and OOD evaluation.

\begin{table}[t]
\centering
\caption{
Paired comparison of \textsc{SpecEvo} against baselines on
LSR-Synth under the ID and OOD evaluation settings.
}
\label{tab:lsrsynth_significance}
\small
\setlength{\tabcolsep}{5pt}
\begin{tabular}{llccc}
\toprule
\textbf{Split}
& \textbf{Baseline}
& \textbf{W / T / L}
& \textbf{Mean $\Delta$}
& \textbf{95\% CI} \\
\midrule
ID
& EvoX
& 97 / 2 / 30
& $+0.91$
& $[+0.45,\,+1.37]$ \\
ID
& AdaEvolve
& 95 / 1 / 33
& $+0.92$
& $[+0.45,\,+1.38]$ \\
ID
& GEPA
& 109 / 1 / 19
& $+1.64$
& $[+1.20,\,+2.06]$ \\
ID
& OpenEvolve
& 96 / 4 / 29
& $+0.87$
& $[+0.44,\,+1.30]$ \\
\midrule
OOD
& EvoX
& 81 / 1 / 47
& $+0.93$
& $[+0.40,\,+1.47]$ \\
OOD
& AdaEvolve
& 85 / 0 / 44
& $+1.14$
& $[+0.49,\,+1.79]$ \\
OOD
& GEPA
& 100 / 1 / 28
& $+1.48$
& $[+0.91,\,+2.02]$ \\
OOD
& OpenEvolve
& 83 / 2 / 44
& $+0.79$
& $[+0.22,\,+1.37]$ \\
\bottomrule
\end{tabular}
\end{table}

%
%

\definecolor{mygreen}{rgb}{0,0.5,0}
\definecolor{myblue}{rgb}{0.1,0.1,0.5}
\definecolor{myred}{rgb}{0.6,0.1,0.1}
\definecolor{codebg}{rgb}{0.97,0.97,0.97}

\newcommand{\promptfield}[2]{%
  \par\smallskip\noindent\textcolor{#1}{\texttt{\{#2\}}}\par\smallskip}

\newcommand{\speculatoroutput}{%
  \promptfield{myblue}{Output Format}
  Return exactly three sections in this order:
  \texttt{\#\# Analysis}, \texttt{\#\# Description}, and
  \texttt{\#\# Code}.  The analysis must follow the operator-specific
  structure below.  The description must be one plain-prose paragraph of
  2--4 sentences (at most 80 words) naming the algorithmic paradigm, its key
  data structures, and its distinguishing mechanism.  The code section must
  contain one fenced \texttt{python} block with the complete, runnable,
  self-contained program.  The required function signature must match
  exactly; include every import and do not use placeholders, ellipses, or
  incomplete branches.}

\newcommand{\speculatorbasicoutput}{%
  \promptfield{myblue}{Output Format}
  Return exactly two sections in this order: \texttt{\#\# Description} and
  \texttt{\#\# Code}.  The description must be one plain-prose paragraph of
  2--4 sentences (at most 80 words) naming the algorithmic paradigm, its key
  data structures, and its distinguishing mechanism.  The code section must
  contain one fenced \texttt{python} block with the complete, runnable,
  self-contained program.  Match the required function signature exactly and
  include every necessary import.}

\newcommand{\navigatoroutput}[1]{%
  \promptfield{mygreen}{Output Format}
  Return exactly three sections in this order: \texttt{\#\# Analysis},
  \texttt{\#\# Description}, and \texttt{\#\# Code}.  The first line of the
  analysis must be \texttt{MOVE: #1}.  The description must be one plain-prose
  paragraph of 2--4 sentences (at most 80 words) naming the paradigm, key data
  structures, and distinguishing mechanism.  The code must be complete and
  runnable, reproduce the required signature exactly, define every helper, and
  include every import.}

\newtcolorbox{speculatorprompt}[1]{%
  enhanced,
  breakable,
  colframe=myblue!85,
  colback=white,
  sharp corners=southwest,
  colbacktitle=myblue!85,
  title=\textbf{Speculator --- #1},
  fonttitle=\bfseries\color{white},
  boxrule=0.6pt,
  arc=6pt,
  left=1mm, right=1mm,
  top=0.5mm, bottom=0.5mm,
  before upper={\small\raggedright},
}

\newtcolorbox{navigatorprompt}[1]{%
  enhanced,
  breakable,
  colframe=mygreen!85!black,
  colback=white,
  sharp corners=southwest,
  colbacktitle=mygreen!85!black,
  title=\textbf{Navigator --- #1},
  fonttitle=\bfseries\color{white},
  boxrule=0.6pt,
  arc=6pt,
  left=1mm, right=1mm,
  top=0.5mm, bottom=0.5mm,
  before upper={\small\raggedright},
}

\newtcolorbox{advisorprompt}[1]{%
  enhanced,
  breakable,
  colframe=myred!90,
  colback=white,
  sharp corners=southwest,
  colbacktitle=myred!90,
  title=\textbf{Advisor --- #1},
  fonttitle=\bfseries\color{white},
  boxrule=0.6pt,
  arc=6pt,
  left=1mm, right=1mm,
  top=0.5mm, bottom=0.5mm,
  before upper={\small\raggedright},
}

\clearpage
\section{Prompt Templates}
\label{app:prompt-templates}

This appendix reports the prompt templates used by the three functional roles in \textsc{SpecEvo}.  

\subsection{Speculator}
\label{app:prompt-speculator}

The Speculator uses the lightweight model for high-throughput exploration: it creates initial variants, reviews archived parents, and applies several mutation and crossover operators.  Operator diversity prevents repeated generic rewrites, while the shared output contract ensures that every accepted candidate has both executable code and an embeddable strategy description. The six variation operators share the template below, and Table~\ref{tab:speculator-operators} lists their operator-specific instructions together with the auxiliary Speculator prompts.

\begin{speculatorprompt}{Shared template for the six variation operators}
You are an expert in optimization heuristics [operator-specific role; see Table~\ref{tab:speculator-operators}].

\promptfield{myblue}{Task Description}

\promptfield{myblue}{Function Signature}

\promptfield{myblue}{Parent Program(s) and Score(s): one parent, or two for crossover}

\promptfield{myblue}{Archive Inspirations: descriptions and scores only}

\promptfield{myblue}{Optional Lessons Learned from the Advisor}

\promptfield{myblue}{Operator-Specific Instruction: see Table~\ref{tab:speculator-operators}}

\speculatoroutput
\end{speculatorprompt}

\begin{table}[H]
\centering
\small
\caption{Operator-specific instructions inserted into the shared Speculator template, and the auxiliary Speculator prompts. Unless noted otherwise, outputs follow the three-section format (\texttt{\#\# Analysis}, \texttt{\#\# Description}, \texttt{\#\# Code}); the required \texttt{\#\# Analysis} parts are listed in italics.}
\label{tab:speculator-operators}
\begin{tabularx}{\linewidth}{@{}>{\raggedright\arraybackslash}p{2.7cm}>{\raggedright\arraybackslash}X@{}}
\toprule
\textbf{Prompt} & \textbf{Role and instruction} \\
\midrule
\multicolumn{2}{@{}l}{\textit{Variation operators (sampled uniformly; shared template)}} \\
General mutation & Improve a candidate algorithm: preserve the parent's useful structure but make one concrete algorithmic change likely to raise the score; no renaming, reformatting, comment edits, random-seed changes, or isolated constant retuning; treat inspirations as high-level ideas. \textit{Components (3--5 main phases); Strengths; Weaknesses (most likely bottleneck, citing a code element); Plan (one sentence).} \\
Focused single-weakness repair & Make a surgical improvement to an already functional candidate: fix exactly one weakness while preserving the parent's algorithmic identity and every unrelated component; no unrelated improvements, constant retuning, seed-only changes, or unnecessary helpers. \textit{Tightest constraint; Fix (minimum meaningful change); Preserved.} \\
Mechanism replacement & Replace one interchangeable local mechanism without changing the overall algorithm (e.g., uniform with stratified initialization, a fixed step with an adaptive step or line search, greedy with Metropolis acceptance); preserve the signature, I/O contract, and parent structure. \textit{Mechanism identified; Old versus new; Risk (main failure mode and guard).} \\
Analysis-targeted mutation & Additional input: a cached analysis of the parent (parent-analysis prompt below). Implement exactly the one suggested change most likely to improve the score, without inventing another direction. \textit{Chosen bottleneck; Implementation plan; What stays unchanged.} \\
Structural crossover & Two parents (A and B): build a coherent hybrid from the strongest compatible mechanism of each parent within a single data flow, objective, and constraint-handling logic, rather than pasting blocks together. \textit{Component table (initialization, optimization core, constraint handling; each from A, B, or a hybrid); Compatibility note; Expected improvement.} \\
Component-level crossover & Base and donor parents: keep the base's control flow, data flow, termination logic, and I/O contract, and transplant exactly one clearly stronger donor component with only the necessary glue. \textit{Donor component chosen; Glue work; What is not changed.} \\
\midrule
\multicolumn{2}{@{}l}{\textit{Auxiliary Speculator prompts}} \\
Initialization variant & Input: code and scores of one or two seeds. Write a complete, feasible, and meaningfully different variant rather than copying a seed. Output: \texttt{\#\# Description} and \texttt{\#\# Code} only. \\
Parent analysis & Review why a parent likely does not score higher, without writing code: an algorithm summary (1--2 sentences), the top three bottlenecks ranked by expected impact (each tied to a specific code element and labeled High, Medium, or Low), and one implementable change per bottleneck; under 250 words. \\
Navigator fan-out variant & Input: an accepted Navigator program and its score. Keep its core approach intact while modifying constants, thresholds, secondary heuristics, edge-case handling, or implementation details. Output: \texttt{\#\# Description} and \texttt{\#\# Code} only. \\
Missing-description recovery & Input: a program whose description is missing or too short. Write one paragraph (2--4 sentences, under 80 words) naming the paradigm, key data structures, and distinguishing heuristic, without headings or boilerplate. \\
\bottomrule
\end{tabularx}
\end{table}

\subsection{Navigator}
\label{app:prompt-navigator}

The Navigator uses the frontier model for the small number of decisions that
require population-level reasoning.  It first creates distinct seed families
and later selects Synthesis, Surgical, or Reframe according to the stagnation signal.

\begin{navigatorprompt}{Prompt for population initialization}
You are an expert in the domain of optimization heuristics helping to design
algorithms that can effectively solve optimization problems.

\promptfield{mygreen}{Task Description}

\promptfield{mygreen}{Function Signature}

\promptfield{mygreen}{Previously Accepted Seeds: complete programs and scores}

Design a solution using a fundamentally different algorithm from every
existing seed.  First analyze which paradigm each seed uses, what information
it exploits, and what relevant information it ignores.  Then design from first
principles using a different modeling, decomposition, construction, or search
strategy.  Do not make a minor variation, tune parameters, reorder existing
logic, or refactor the same core algorithm.

\promptfield{mygreen}{Output Format}
Return \texttt{\#\# Description} followed by \texttt{\#\# Code}.  The
description is a 2--4 sentence paragraph (at most 80 words) stating the
paradigm, key data structures, and distinguishing mechanism.  The code must be
complete, runnable, self-contained, and expose the exact required signature.
\end{navigatorprompt}

\begin{navigatorprompt}{Prompt for Synthesis}
You are an expert research lead synthesizing complementary mechanisms from
competitive solutions.

\promptfield{mygreen}{Task Description}

\promptfield{mygreen}{Function Signature}

\promptfield{mygreen}{Archive State: evaluation count, occupied niches, and stagnation}

\promptfield{mygreen}{Top Three Anchors: description, score, and complete program}

\promptfield{mygreen}{Five Additional Inspirations: descriptions and scores only}

The search is only mildly stalled.  Synthesize, rather than invent, one program
that coherently combines 2--3 concrete mechanisms from different anchors and
plausibly beats each anchor individually.  After the line
\texttt{MOVE: SYNTHESIS}, include exactly three analysis parts:
\textbf{Component table} (initialization, optimization core, and constraint
repair, with anchor provenance), \textbf{Coherence note} (shared data layout,
units, and call order), and \textbf{Why this should beat all anchors} (one
sentence per anchor). 

\navigatoroutput{SYNTHESIS}
\end{navigatorprompt}

\begin{navigatorprompt}{Prompt for Surgical improvement}
You are an expert research lead applying one precise structural improvement to
the current champion.

\promptfield{mygreen}{Task Description}

\promptfield{mygreen}{Function Signature}

\promptfield{mygreen}{Archive State: evaluation count, occupied niches, and stagnation}

\promptfield{mygreen}{Current Champion: description, score, and complete program}

\promptfield{mygreen}{Five Additional Inspirations: descriptions and scores only}

The search is moderately stalled, so improve the champion without changing its
algorithmic family.  After the line \texttt{MOVE: SURGICAL}, include exactly
four analysis parts: \textbf{Tightest constraint} (cite the exact limiting
mechanism), \textbf{Structural fix} (one local behavioral change),
\textbf{Preservation list}, and \textbf{Expected delta} (benefit, main risk,
and guard).  Implement exactly one local structural change; prefer using
existing score, feasibility, slack, loss, or quality signals over a blind
heuristic, and do not add unrelated improvements.

\navigatoroutput{SURGICAL}
\end{navigatorprompt}

\begin{navigatorprompt}{Prompt for Reframe}
You are an expert research lead proposing a new algorithmic family when the current population has plateaued.

\promptfield{mygreen}{Task Description}

\promptfield{mygreen}{Function Signature}

\promptfield{mygreen}{Archive State: evaluation count, occupied niches, and stagnation}

\promptfield{mygreen}{Top Two Anchors: description, score, and complete program}

\promptfield{mygreen}{Five Additional Inspirations: descriptions and scores only}

The search is deeply stalled.  Design a fundamentally different paradigm whose
data structures, control flow, and termination rule do not appear in the
anchors or inspirations.  After the line \texttt{MOVE: REFRAME}, include exactly four analysis parts: \textbf{Paradigm name} (use the textbook name), \textbf{Why it fits the problem} (two problem-specific sentences), \textbf{Why the current anchors miss it} (the shared assumption being dropped), and \textbf{Risk} (the main implementation pitfall and guard). Forbidden moves are constant retuning, subroutine grafting, and variable renaming.  

\navigatoroutput{REFRAME}
\end{navigatorprompt}

\subsection{Advisor}
\label{app:prompt-advisor}

The Advisor converts deterministic, niche-level search statistics and accumulated failures into short natural-language guidance. 

\begin{advisorprompt}{Prompt for population-level reflection}
You are a lessons-learned advisor reviewing the search trajectory of an
evolutionary code-optimization system.  Write a short prescriptive note that a
mutation model will read verbatim before some future attempts.  Do not restate
the task or generic constraints already enforced by the evaluator.

\promptfield{myred}{Task Description}

\promptfield{myred}{Function Signature}

\promptfield{myred}{Current State: best score, evaluations, accept rate, and stagnation}

\promptfield{myred}{Improving Niches: frontiers advanced in the current window}

\promptfield{myred}{Saturated Niches: recent attempts, unchanged frontiers, and scores}

\promptfield{myred}{Accumulated Failure Knowledge: recurrence count and example}

\promptfield{myred}{Previous Advice}

Write exactly the following sections, in this order, using the literal headers and no other Markdown:

\texttt{WORKING:} In 1--2 sentences, name the concrete algorithmic structure
or mechanism shared by improving niches and leaders.  Describe the algorithm,
not the search process; if nothing is clearly working, say so.

\texttt{SATURATED:} In 1--2 sentences, name the busy strategy families whose
frontiers no longer improve and should be de-emphasized.  If there is no clear
saturation, write ``none.''

\texttt{AVOID:} Provide up to 12 concrete error-prevention instructions as a numbered checklist. Ground each instruction in the supplied failure evidence or still-applicable previous guidance. 

Keep the complete answer under 500 words. Return advice only; do not output code.

\end{advisorprompt}

\newtcolorbox{advisorbox}[1]{%
  breakable, colback=black!3, colframe=black!45, boxrule=0.4pt, arc=2pt,
  left=6pt, right=6pt, top=5pt, bottom=5pt,
  fonttitle=\bfseries\small, title={#1}}

\section{Qualitative Analysis}
\label{app:qualitative}

This appendix examines one \textsc{SpecEvo} search run on each of two tasks: Circle Packing Rectangle, a continuous geometric problem, and EPLB, a discrete resource-allocation problem for mixture-of-experts inference. The evaluation log of each run records every candidate program, the role and model that produced it, its execution status, and the evaluator message. For each task, we describe why candidates were rejected, which algorithmic mechanisms the population used, and the guidance the Advisor derived from the failures. Failure classes are listed from most to least common. We label how widely each mechanism appears in the population as \emph{ubiquitous}, \emph{common}, \emph{occasional}, or \emph{rare}; a program usually combines several mechanisms.

\subsection{Circle Packing Rectangle}
\label{app:qualitative:cpr}

The task is to place 21 non-overlapping circles inside a rectangle of fixed perimeter so as to maximize the sum of their radii. A configuration that violates feasibility receives no score, even if the violation is only one part in a thousand.

\paragraph{Recurring errors.}
Table~\ref{tab:cpr-errors} lists the failure classes. The most common failure, by a wide margin, is a pair of overlapping circles. These are not near misses: the overlaps appear when a program inflates the radii or rescales a finished layout to fill the container and returns it without re-checking pairwise distances, so relaxing the tolerance would accept almost none of them. The second most common class is a wrong output shape. The evaluator expects an array with 21 rows, each holding a center and a radius, but helper functions, early returns, and especially fallback branches, which run only when the main path fails, often return a different shape. The remaining classes are ordinary programming faults; a few programs cannot be parsed because they contain typographic Unicode symbols, such as a minus sign or a less-than-or-equal sign copied from prose.

\begin{table}[t]
\centering
\footnotesize
\setlength{\tabcolsep}{3pt}
\renewcommand{\arraystretch}{1.08}
\caption{Failure classes among rejected candidates on Circle Packing Rectangle,
ordered from most to least common. Each rejected candidate belongs to exactly one
class.}
\label{tab:cpr-errors}
\begin{tabularx}{\linewidth}{@{}>{\raggedright\arraybackslash}p{1.65cm}>{\raggedright\arraybackslash}X@{}}
\toprule
Class & What goes wrong \\
\midrule
Feasibility  & Two circles intersect after the radii are inflated or the layout is rescaled, but the program returns without rechecking pairwise feasibility \\
Output shape & The returned array does not contain 21 rows with three values each, usually because a helper or fallback path uses a different representation \\
Type         & Lists and arrays are combined in arithmetic, or a coordinate is missing or infinite \\
Name         & A variable or import is used before being defined, often within a nested helper or early return branch \\
Runtime      & The program exhausts the evaluation time limit instead of stopping after convergence \\
Numeric      & An array is used directly in a truth test, or a reduction is applied to an empty array \\
Indexing     & An index exceeds the array bounds while circle pairs are enumerated \\
Syntax       & Typographic Unicode symbols or malformed code prevent parsing \\
\bottomrule
\end{tabularx}
\end{table}

The same failure classes appear under all model-allocation strategies in Appendix~\ref{app:baseline-details}, and overlapping circles are the most common failure under every one of them. In those runs, the number of rejections per quarter of the run stays roughly constant, because no component carries information about earlier failures forward. In the \textsc{SpecEvo} run, the rejection rate drops substantially from the initialization phase, which has no Advisor guidance, to the evolutionary phase, in which Advisor notes are included in Speculator prompts.

The RelayEvolve logs also show how failures differ between the two model tiers. When both models receive comparable numbers of calls, their rejection rates and overlap rates are similar. Timeouts, however, come almost entirely from the strong model, whose programs are longer and rely on expensive global optimizers, whereas indexing and shape errors come mostly from the lightweight model. \textsc{SpecEvo} uses the lightweight model for high-volume exploration and reserves the strong model for sparse interventions.

\paragraph{Algorithmic mechanisms.}
Table~\ref{tab:cpr-ideas} groups the mechanisms by role. Nearly every program starts from a structured layout, typically hexagonal rows with alternating offsets, and most then search over the aspect ratio of the rectangle, because its shape determines how many circles fit in each row. Refinement uses penalty descent on pairwise overlaps, simulated annealing, or damped pairwise separation. Repair rescales the layout to the perimeter constraint, grows the radii greedily up to the nearest-neighbor limit, and ends with a shrink pass that removes the remaining overlaps. A wall-clock guard that returns the best feasible configuration found so far is present in almost every program.

\begin{table}[t]
\centering
\footnotesize
\setlength{\tabcolsep}{3pt}
\renewcommand{\arraystretch}{1.08}
\caption{Algorithmic mechanisms observed on Circle Packing Rectangle, grouped by
their role. A program typically combines mechanisms from every group.}
\label{tab:cpr-ideas}
\begin{tabularx}{\linewidth}{@{}>{\raggedright\arraybackslash}p{1.8cm}>{\raggedright\arraybackslash}p{0.32\linewidth}>{\raggedright\arraybackslash}X@{}}
\toprule
Role & Mechanism & Prevalence and remarks \\
\midrule
Construction & Structured lattice or patterned row initialization & Ubiquitous; hexagonal rows with an explicit pattern rather than random points \\
             & Search over the container aspect ratio & Common; the objective couples container shape with row occupancy \\
\addlinespace
Refinement   & Simulated annealing with stochastic acceptance & Common; uses a cooling schedule over perturbed configurations \\
             & Continuous penalty descent & Common; penalizes intersections smoothly rather than forbidding them \\
             & Pairwise separation relaxation & Common; applies damped opposing displacements to overlapping pairs \\
             & Enlargement with backtracking for each circle & Occasional; halves the step until an individual radius fits \\
\addlinespace
Repair       & Global rescaling to satisfy the perimeter constraint & Common; applied after every stage rather than only at the end \\
             & Final feasibility shrink pass & Common; uses the smallest global factor that removes all intersections \\
             & Greedy radius inflation to the nearest neighbor limit & Occasional; grows each radius to the largest value permitted by its neighbors \\
\addlinespace
Engineering  & Wall-clock guard with a best feasible fallback & Ubiquitous; the only defense against the runtime failure class \\
             & Priority queue processing & Occasional; handles the largest radius or violation first \\
             & Multiple restart portfolio & Occasional; retains the best of several independent attempts \\
             & Neighbor queries using a spatial index & Rare; replaces pairwise distance calculations with a spatial tree or triangulation \\
\bottomrule
\end{tabularx}
\end{table}

The initialization phase and the Speculators introduce individual mechanisms and refine them one at a time, so the record improves in small steps. In this run, the Navigator's record-setting interventions combined existing mechanisms instead of introducing new ones. The final best program initializes a hexagonal layout, relaxes it with damped pairwise separation, grows the radii toward their nearest-neighbor limits, adds a backtracking enlargement pass for each circle, rescales to the perimeter constraint after every stage, and ends with a shrink pass that guarantees feasibility. Each of these components already existed in the population, but no program using only one of them reached the score of their combination.

\paragraph{Advisor feedback.}
The box below shows the guidance the Advisor had consolidated by the end of the run. Each item is an instruction for one failure class, not a diagnosis of a single program, and the items are ordered by the number of rejected candidates they address. The note is rewritten at every checkpoint instead of extended, which keeps it short enough to include verbatim, and it is added to only a fraction of Speculator prompts so that it does not push the whole population toward one defensive coding style.

\begin{advisorbox}{Consolidated Advisor guidance: Circle Packing Rectangle}
\small
\begin{enumerate}\itemsep2pt
\item Ensure that no two circles overlap by keeping the distance between their
      centers at least equal to the sum of their radii minus the tolerance, as the
      validator requires.
\item Return exactly 21 circles as an array with 21 rows and three values per row,
      and preserve this shape through every operation.
\item Convert all coordinates and radii to double precision before arithmetic,
      and ensure that no value is missing or infinite.
\item Define and import every name before use, and call only documented public
      interfaces.
\item Keep the total runtime below the evaluation limit by bounding iteration
      counts and stopping early after convergence.
\item Confirm that an array is nonempty before using it in a truth test or a
      reduction.
\item Clip indices to the actual array size, especially when enumerating circle
      pairs.
\item Use only plain ASCII characters, replacing typographic minus and less than
      or equal symbols with their ASCII equivalents.
\item Clip radii to a small positive lower bound before any geometric computation
      to avoid division by zero and invalid square roots.
\end{enumerate}
\end{advisorbox}

\subsection{EPLB}
\label{app:qualitative:eplb}

The task is to place logical experts onto physical GPU slots for mixture-of-experts inference. Given load statistics for each layer, a program decides how many replicas each expert receives and which slot each replica occupies, subject to a fixed replica budget, a fixed hierarchy of groups, nodes, and GPUs, and a capacity limit for each GPU. The problem is discrete, and a program must keep three indices consistent as replica counts change: the logical expert, the replica, and the physical slot.

\paragraph{Recurring errors.}
Table~\ref{tab:eplb-errors} lists the failure classes. On Circle Packing Rectangle, most rejected programs violate a geometric constraint; on EPLB, indexing errors dominate: a slot or replica index computed from capacity arithmetic exceeds the bounds of the tensor it indexes, even when the allocation policy itself is sound. Two classes have a single main cause. Most name errors come from programs that schedule each replica on the least-loaded GPU with a priority queue, implement the queue operations correctly, and never import the priority-queue module; some of these programs also call heap routines of the tensor library that do not exist. Most type errors come from a deterministic hash initialization, which multiplies a large hexadecimal constant by a slot index to spread the replicas of an expert across devices without a random number generator. Unless the product is reduced modulo the number of GPUs before it is used as an index, it overflows the integer range and yields an index far outside the tensor. Nearly every overflow rejection in the run came from a program containing such a constant, and a single guidance item can remove this defect while keeping the mechanism.

\begin{table}[t]
\centering
\footnotesize
\setlength{\tabcolsep}{3pt}
\renewcommand{\arraystretch}{1.08}
\caption{Failure classes among rejected candidates on EPLB, ordered from most to
least common. Each rejected candidate belongs to exactly one class.}
\label{tab:eplb-errors}
\begin{tabularx}{\linewidth}{@{}>{\raggedright\arraybackslash}p{1.65cm}>{\raggedright\arraybackslash}X@{}}
\toprule
Class & What goes wrong \\
\midrule
Indexing     & A physical or logical index derived from capacity arithmetic exceeds a tensor bound \\
Name         & A module or attribute is used without being imported or defined, most often the priority queue module \\
Type         & A tensor has the wrong element type, or an integer computation exceeds its range \\
Feasibility  & Replica counts do not sum to the budget for every layer, or the placement exhausts the available capacity \\
Runtime      & The placement loop exhausts the evaluation time limit \\
Shape        & A tensor dimension does not match the extent expected by the evaluator \\
Numeric      & A nonscalar tensor is used directly in a truth test \\
Output shape & The function does not return the three required tensors in the required order \\
Syntax       & A typographic Unicode symbol makes the program unparseable \\
\bottomrule
\end{tabularx}
\end{table}

The rejection rate decreases throughout the run, with the largest drop between the initialization phase and the evolutionary phase. The mechanical faults that are frequent early on, namely the missing import, the overflowing constant, and the invalid slot index, are among the failure signatures accumulated by the Advisor. They are largely absent from the later population, although the same algorithmic ideas remain in use.

\paragraph{Algorithmic mechanisms.}
Table~\ref{tab:eplb-ideas} lists the mechanisms. The interface imposes a per-layer solution and the group, node, and GPU hierarchy, so programs differ in three decisions. To choose the number of replicas per expert, most programs repeatedly give the next replica to the expert whose load per replica it reduces most, which is the classical apportionment rule (some programs implement it explicitly as Huntington--Hill); a substantial minority instead binary-search a load threshold and give each expert enough replicas to fall below it, and the strongest programs use the threshold search to approach the budget and the marginal-gain rule to correct the remainder. To assign slots, programs use a modular or rotated round-robin walk over the GPUs, which spreads the replicas of each expert across devices, or a priority queue that places each replica on the least-loaded GPU. To enforce feasibility, most programs clamp to capacity, fill the remaining slots in a repair pass, and assert that the replica counts sum to the budget; the Advisor reinforces these mechanisms.

\begin{table}[t]
\centering
\footnotesize
\setlength{\tabcolsep}{3pt}
\renewcommand{\arraystretch}{1.08}
\caption{Algorithmic mechanisms observed on EPLB, grouped by the design decision
they resolve. A program typically combines mechanisms from every group.}
\label{tab:eplb-ideas}
\begin{tabularx}{\linewidth}{@{}>{\raggedright\arraybackslash}p{1.8cm}>{\raggedright\arraybackslash}p{0.32\linewidth}>{\raggedright\arraybackslash}X@{}}
\toprule
Decision & Mechanism & Prevalence and remarks \\
\midrule
Structure     & Independent solution for each layer & Ubiquitous; imposed by the interface \\
              & Explicit treatment of the group, node, and GPU hierarchy & Ubiquitous; implements placement as a nested traversal of the hierarchy \\
\addlinespace
Replicas      & Allocation by marginal gain or apportionment & Ubiquitous; assigns the next replica where it most reduces the load per replica \\
              & Water-filling threshold found by bisection & Common; gives each expert enough replicas to fall below a searched load threshold \\
              & Huntington--Hill apportionment by name & Occasional; implements the classical rule explicitly \\
\addlinespace
Placement     & Modular or rotated round-robin assignment & Common; disperses the replicas of each expert across devices \\
              & Deterministic hash initialization of the starting slot & Common; provides dispersion without a random number generator but causes the overflow failures \\
              & Priority queue scheduling by load & Occasional; places each replica on the least loaded GPU \\
              & Local swap or exchange refinement & Rare; exchanges two placed replicas after the initial assignment \\
\addlinespace
Feasibility   & Capacity clamping with a repair pass & Ubiquitous; fills remaining slots after the main assignment \\
              & Explicit assertion on the replica budget & Common; checks the contract rather than assuming it \\
              & Vectorized tensor operations instead of Python loops & Common; the main defense against the runtime failure class \\
\bottomrule
\end{tabularx}
\end{table}

As on Circle Packing Rectangle, the Navigator's \textsc{Synthesis} and \textsc{Surgical} interventions produce strong, valid programs that reinforce the incumbent design. Its later \textsc{Reframe} interventions, which introduce a strategy family absent from the population, score below the incumbent. This is expected: \textsc{Reframe} is triggered by stagnation to increase diversity, and the archive keeps these programs as parents even when they do not improve the record. The final record was set by a lightweight structural crossover that combines threshold-based replica allocation with the marginal-gain correction and a placement that respects the hierarchical capacity constraints; its structure came from an earlier Navigator proposal that lightweight variants then refined.

\paragraph{Advisor feedback.}
The EPLB guidance below is more mechanical than the Circle Packing Rectangle guidance, because most rejected EPLB programs use sound allocation policies and fail on indexing, imports, or element types.

\begin{advisorbox}{Consolidated Advisor guidance: EPLB}
\small
\begin{enumerate}\itemsep2pt
\item Clip every index to the valid range of the dimension it addresses, keeping
      physical and logical indices within bounds and consistent with the expected
      shapes.
\item Import all required modules, including the priority queue module, at the top
      of the program, and call only documented public interfaces rather than
      private tensor library routines.
\item Convert tensor values to the intended element type before use, and maintain
      that type consistently across operations.
\item Ensure that replica counts sum to the replica budget for every layer, and
      clamp them to both the budget and the available GPU capacity.
\item Bound the number of placement iterations so that the program remains within
      the evaluation time limit.
\item Give the mapping from logical to physical indices the required three
      dimensions, with the final extent equal to the largest replica count.
\item Pass tensor shapes as tuples of integers rather than as a single integer.
\item Reduce a tensor to a scalar before using it in a truth test.
\item Return exactly the three required tensors in the order specified by the
      function signature.
\item Use only plain ASCII characters because a typographic hyphen is not valid
      syntax.
\item Use floor rounding rather than ceiling rounding in integer division so that
      the result matches the expected element type.
\end{enumerate}
\end{advisorbox}

In both tasks, a few instructions cover most of the observed failures, and the first instructions concern output and feasibility requirements, not the algorithm. A stronger model alone would not be expected to remove these failures: in the RelayEvolve runs, the strong and lightweight models fail at similar overall rates and differ mainly in which requirements they violate. The difference in \textsc{SpecEvo} is that this evidence is carried forward: the note is rebuilt at every checkpoint from all accumulated failure signatures, so a failure observed early in the run still affects later proposals.

\end{document}